\documentclass{article}

\usepackage[preprint]{neurips_2026}
\usepackage[utf8]{inputenc}
\usepackage[T1]{fontenc}
\usepackage{hyperref}
\hypersetup{hidelinks,pdftitle={The Window Lever: Longer Observation Windows Outweigh the Observed Effects of Data, Capacity, and Architecture in Customer Behavior Forecasting},pdfauthor={Tara Sowrirajan, Yanqiao Wang}}
\usepackage{url}

\usepackage{booktabs}
\usepackage{array}
\newcolumntype{L}[1]{>{\raggedright\arraybackslash}p{#1}}
\usepackage{amsfonts,amsmath,amssymb,mathtools}
\usepackage{microtype}
\usepackage{xcolor}
\usepackage{graphicx}
\usepackage{enumitem}
\usepackage{natbib}

\usepackage{tikz}
\usetikzlibrary{arrows.meta,positioning,calc,fit}
\usepackage{pgfplots}
\pgfplotsset{compat=1.16}

\newcommand{\E}{\mathbb{E}}
\DeclareMathOperator{\softmax}{softmax}

\newcommand{\abstractAssertive}{%
In our experiments on long-horizon customer-behavior forecasting, a longer observation window
outweighed the observed effects of more data, more capacity and newer architectures. The window is the weeks of history a model sees
before the week it forecasts from. Going from an 8- to a 16-week window adds a raw
$+0.070$--$0.107$ log-space $R^2$ in every family at 1M. Matched test cohorts, which score
the same customer-weeks at every window, show this raw comparison \emph{understates} the gain in
every family they cover. The direction holds on five public datasets, four spanning over a year
(ridge and our transformer on all five).
On our internal data, longer
windows admit up to $10\%$ more and colder training customers, which matching does not fix. Our design
does not separate window length from calendar coverage, because every test anchor (the week a
forecast is made from) falls in one eight-week band. On training-period anchors, the weekly ridge probe's
gain persists outside the holiday season, keeping about $72\%$ of it. We therefore report a robust positive
association with longer windows and do not identify the magnitude of a calendar-independent
history effect. Of four internally registered predictions, the only fully blind one failed. An
activity-mixture model predicted $+0.014$ from 24 to 32 weeks; the
weekly-aggregate probe returned $-0.0004$. For the sequence families, the other
three levers are about an order of magnitude smaller, bounded and unresolved. Their observed
effects are at most $0.006$ between the nominal 1M and 10M tiers and $0.0052$ among five sequence
families at 16 weeks with clocks as built, both single-seed on the full test split, and $0.006$ for capacity on three-seed means. These
are observed intervals, not equivalence bounds. Our claim is the order-of-magnitude contrast
between the window and those bounds. A team building on this class of data should start by
lengthening the window.%
}

\title{The Window Lever: Longer Observation Windows\\Outweigh the Observed Effects of Data, Capacity,\\and Architecture in Customer Behavior Forecasting}

\author{%
  Tara Sowrirajan\thanks{Corresponding author: \texttt{tarastar@amazon.com}} \quad Yanqiao Wang \\
  Amazon \\
}

\begin{document}
\maketitle

\begin{abstract}
\abstractAssertive
\end{abstract}

\section{Introduction}
\label{sec:intro}

The usual levers for a sequence model are more data, more parameters and a newer
architecture~\citep{kaplan2020scaling,hoffmann2022chinchilla}, while history length is usually fixed
or treated as a capacity setting (\S\ref{sec:related}). We compare them on an
industrial task, forecasting a customer's activity from their event
history~\citep{fader2005counting,wang2019deepltv,li2022billion}, and of the levers we tested, history
length, the one usually fixed, moves accuracy most. Two lessons carry over. A longer window changes who
is scored, biasing raw comparisons either way; scoring every window on one matched cohort removes that
bias on the test side (training cohorts still grow). A cheap ridge fit on activity aggregates shows the
history gain on every dataset we tried (toy model in \S\ref{app:toy}).

Whether any of these levers binds is not a new question. How much event history a churn model needs has been asked
directly~\citep{ballings2012churn}, and observation-window
sufficiency is under active study (\S\ref{sec:related}). However, the question is rarely
answered under controlled conditions. Interventions ship sequentially, on moving populations, and
improvements can be hard to attribute when several parts of the modeling pipeline change at
once. We run a controlled comparison of these commonly modified axes. We hold fixed the
export, the test anchor weeks, the target
(364-day forward activity, $\log(1{+}y)$-transformed, scored in log-space $R^2$), and
the evaluation. We then sweep, one factor at a time, the axes the field usually credits: model family
(eight, from a single-feature linear model to a decay-gated recurrent model~\citep{mei2017neural}),
population scale (the nominal 1M/5M/10M sampling tiers, \S\ref{sec:data}), capacity, and training
objective. The training recipe is held fixed within each family but not across families (Table~\ref{tab:recipe}).
We also sweep the length of the observation window, and it is the largest lever we measured. We
internally registered four falsifiable commitments on whether the window gain survives matching
and how it bends as the window grows, each with its falsifier fixed before the evaluation that decided it.
The one fully blind prediction failed on its own falsifier, and the other
three held (Table~\ref{tab:prereg}).

Doubling the observation window from 8 to 16 weeks moves
log-space $R^2$ by
$+0.070$--$0.107$ at the 1M tier in \emph{every} family we train. The families are the transformer,
GRU~\citep{cho2014gru}, LSTM~\citep{hochreiter1997lstm},
time-biased attention, a decay-gated recurrent model~\citep{mei2017neural}, boosted trees~\citep{ke2017lightgbm},
a 6-feature ridge, and a single-feature linear model.
Time-biased attention replaces the ODE-evolved keys and values of ContiFormer~\citep{chen2024contiformer} with a learned time-decay bias on the attention logits.
The two continuous-time baselines, labeled ContiFormer and Neural Hawkes in our tables, are simplified stand-ins for the cited
models. In the main runs the
inter-event interval they consume is identically zero, so their time mechanisms
contributed nothing (\S\ref{app:baselines}).
On this end-anchored export, where every input window ends at the week its forecast is made from, doubling the window also shifts which calendar months the input
covers (\S\ref{sec:limitations}).

\begin{figure}[t]
  \centering
  \includegraphics[width=0.9\linewidth]{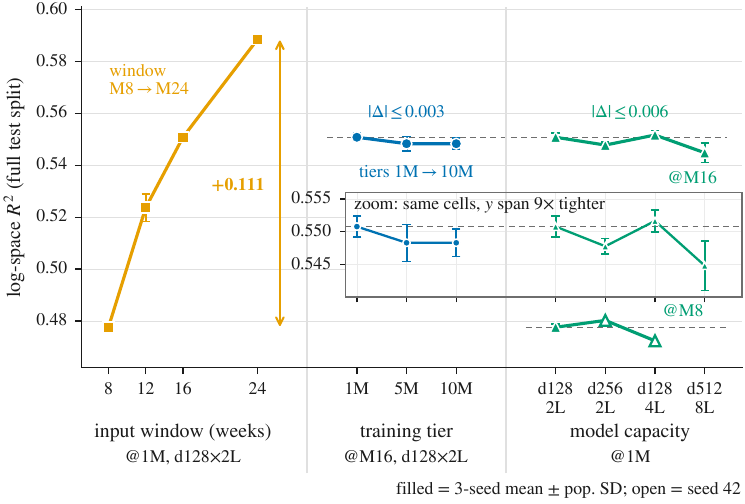}
  \caption{%
    \textbf{The observation window outweighs the observed effects of population scale and model capacity, our counterparts of the data and parameter axes swept by a compute-calibration
    scaling study that does not vary the window in training~\citep{gabrielsson2026scaling}.}
    Extending the input window M8$\to$M24 moves $R^2$ by $+0.111$ (left). On our counterparts of those axes, it moves by $0.0025$ across training-population tiers 1M$\to$10M at fixed window M16 (center)
    and by $0.006$ across width and depth at fixed window M16 and fixed data (right), both as 3-seed means.
    Both are observed intervals rather than resolved equivalence bounds. The scale interval sits within
    seed dispersion. The capacity interval does not, but it is unresolved against its own
    $0.016$ equivalence margin (\S\ref{sec:limitations}).
    The raw M24 endpoint carries the composition
    caveat of Table~\ref{tab:window-grid}; the matched-cohort window climb is larger
    still ($+0.131$, 3-seed, \S\ref{sec:matched}), so the raw delta is the conservative estimate.
    Every ladder shares one setup (Huber objective,
    log-space $R^2$, full test split). Apart from the factor each ladder varies, the only differences are a $100{\to}500$-step warmup at d512 with 8 layers (\S\ref{sec:ceiling}) and, on the tier ladder, the realized learning-rate schedule (\S\ref{sec:limitations}). Window length also shifts calendar coverage (\S\ref{sec:limitations}) and training eligibility (\S\ref{sec:data}).
    Filled markers: 3-seed means $\pm$ population SD; open markers: weaker evidence, here single-seed (seed-42) runs;
    dashed lines: levels of the shared base configurations. The inset redraws the two M16 ladders on a
    $y$ span about nine times tighter, column-aligned to the main axis. The M8 capacity replicate sits
    at its own level and stays on the main axis.
  }
  \label{fig:missing-axis}
\end{figure}

The other levers move the metric substantially less (Figure~\ref{fig:missing-axis}). Among them, the training objective and time representation (the transformer's positional encoding) give the largest gains. Both are smaller than the window gain, and we quantify them below. Moving from the nominal 1M to the 10M sampling tier changes sequence-model levels by
$\leq 0.006$, seed 42. The exception is the 6-feature ridge, whose level dips
$0.016$--$0.028$ at 10M while its window gain stays intact (\S\ref{sec:universal}). The recurrent models train on a fixed-size sample at every tier,
so their differences across tiers measure population composition at a fixed training-set size rather than a response to more data (\S\ref{sec:ceiling}). The measured window gain is similar across the population tiers we evaluate:
$+0.0737$ at 10M vs.~$+0.0731$ at 1M (both 3-seed, Huber). Doubling width or depth moves the metric by
$\leq 0.005$ at 8 weeks, seed 42.
Against the window's $+0.070$--$0.107$ gain, the observed customer-scale and capacity changes are approximately an order of magnitude smaller.
These are one-factor-at-a-time bounds inside the fixed
2-layer, width-128 (d128) recipe, except that the realized learning-rate schedule is not matched across tiers (\S\ref{sec:limitations}). We scale customers at fixed capacity and capacity at
fixed data (1M tier), and never grow capacity and data together, so we
measure no compute-optimal configuration. Two of our own measurements are also about as large as these
bounds. The three-seed detection floor for this class, $0.0059$, sits
above the architecture band reported below and within a factor of $1.2$ of the customer-scale and capacity bounds. Re-selecting
checkpoints in log space, the space we report in, moves validation scores by $+0.002$ to $+0.009$. The top of that range exceeds every one of these bounds, and the shift does not cancel out of the capacity contrast. These
three axes are therefore bounded but unresolved, and we read none of them as a null (\S\ref{sec:limitations}).

In these experiments, observation-window length moved the metric by roughly an
order of magnitude more than sequence-architecture choice did, on observed effect sizes rather than
interval bounds.
With the Huber objective at M16 (a 16-week window), five sequence architectures land within
${\sim}0.005$ of each other with their clocks as built ($0.5525$/$0.5522$/$0.5508$/$0.5478$/$0.5474$, seed 42, full test split).
A separate test that gives each continuous-time model a working clock detects no gain for ContiFormer and helps Neural Hawkes, by $+0.0116$ (95\% $t$ interval over three seeds $+0.0102$ to $+0.0130$), about twice that band. These paired runs are not bridged to the main ones, so where Neural Hawkes would land among the five with a working clock is not measured (\S\ref{app:baselines}).
Because those baselines ran at their own training budgets, we read that band
within families rather than as a cross-family ranking (\S\ref{sec:universal}).
With the MSE objective and 3-seed means, the three families we
retrained land within ${\sim}0.0025$ (transformer $0.5875$ / GRU $0.5863$ / LSTM $0.5850$,
\S\ref{sec:lossnull}). The architecture spread observed at shorter windows narrows substantially by 16 weeks: the decay-gated recurrent model starts
$0.015$ below the transformer at 8 weeks and closes most of that gap by 16 weeks (seed 42).

Through 16 weeks, matching the test customers across windows makes the window gain \emph{larger}
(\S\ref{sec:matched}). We re-evaluate the already-trained checkpoints on \emph{matched cohorts}, which hold
the same customer-weeks at every window, with identical targets and a constant $R^2$
denominator. The matched
$\Delta$(M16$-$M8)
is $+0.096$, against $+0.075$ on the full set (Huber, seed 42 for both; the 3-seed full
delta is $+0.073$). The raw 8-to-16-week comparison understates the gain in every objective and family
measured, and the same pattern holds at 10M
($+0.094$ matched vs.\ $+0.075$ full); both matched ladders replicate at three seeds.

\emph{A confound we do not resolve.} A longer window changes more than composition. The calendar months the input covers move with it. Every test forecast is made from one of three anchor weeks, four weeks apart
(\mbox{2025-02-24}, \mbox{2025-03-24}, \mbox{2025-04-21}; \S\ref{sec:data}). Each input window ends
at one of these three dates, all inside a single ${\approx}8$-week band, so lengthening the window
can only extend it \emph{backward} into particular calendar months. Only at the
earliest anchor does an 8-week window cross the turn of the year, and then by two days.
At that anchor, 16 weeks is the shortest window we test that reaches the prior November, which the
latest anchor reaches only at 24 weeks. The 24- and 32-week windows reach into the preceding
autumn and summer. Window length and calendar coverage are therefore
not separated in this design, against a target that is a 364-day forward
aggregate. The matched cohorts, the activity decomposition, and the
objective and family sweeps all hold those three anchor weeks fixed, and the Taobao
dataset (\S\ref{sec:external}) spans nine days in total. We therefore ran an anchor-varying control.

That control is \emph{void} under its own registered criterion.
When we vary which calendar months the
added block covers, the 8-to-16-week gain is $+0.18731$ at the 2025-02-24
anchor, $+0.14910$ at 2025-03-24 and $+0.04793$ at~\mbox{2025-04-21} for MSE seed 123, and
$+0.21159/{+}0.16369/{+}0.05818$ for MSE seed 456. At both seeds the gain is monotone in calendar position, the ordering a seasonal mechanism predicts. The earliest-minus-latest
contrast is $+0.13938$ (customer-clustered bootstrap 95\% CI $[+0.12459, +0.15432]$) and
$+0.15341$ (95\% CI $[+0.13851, +0.16930]$), both excluding zero, and a 200-draw same-cohort
placebo that shuffles the anchor labels never attains it. We nonetheless do not interpret it, because the criterion we
committed before running the control fails in every setting. That criterion voids the contrast if the
per-anchor $R^2$ spread at M8, \emph{before} the added block enters, exceeds
$0.010$. At both objectives and three seeds each ($6$ settings, $18$
per-anchor values) the spread is $0.0562 / 0.0562 / 0.0623$ on MSE and $0.0367 / 0.0378 / 0.0420$ on
Huber. That is $3.7$--$6.2\times$ the bound, so the criterion voids all six. The numbers show why the criterion exists. In all six settings the
anchors are ordered identically by baseline difficulty and by calendar position
(M8 $R^2 = 0.224 / 0.248 / 0.280$ at MSE seed 123).
The anchor with the most
headroom is therefore also the one whose added block covers the holiday period, and we cannot exclude headroom as an
explanation of the ordering. Algebraically, $37$--$46\%$ of each contrast is carried by the M8
model, which contains no added block at all. On Huber, the latest anchor's added eight weeks
are worth $-0.00451$ to $+0.01407$, small or slightly negative.
We report every per-anchor figure in
this paragraph only as a record of what failed and by how much. We do not interpret any of them as a calendar measurement.
These figures come from the uniquely keyed
matched cohort, the $n = 233{,}038$ rows with a unique target value in each window's test split. Its selection drops the zero-target rows, so its
\emph{levels} are not comparable to any full-set $R^2$.
In a post-hoc check we renormalize each anchor's gain by its own remaining variance,
$(R^2(16)-R^2(8))/(1-R^2(8))$. The two MSE seeds' earliest/latest ratios then move only
from $3.91$ to $3.63$ and from $3.64$ to $3.36$, so a proportional headroom effect accounts for roughly
$7$--$8\%$ of the ordering. That check does not change the void verdict.

The window gain's direction holds in every pooled analysis we report and at every anchor on MSE. Its magnitude, however, remains confounded with calendar coverage and is not identified as a calendar-independent history effect
(\S\ref{sec:limitations}). We conclude from the control that, on this end-anchored export, the test anchors in one
eight-week band cannot separate window length from calendar coverage at either objective. For the transformer runs, a calendar-balanced test set cannot be recovered from this export. For the weekly-aggregate ridge probe of \S\ref{app:probe}, and only for it, the export's training period partly separates window length from calendar coverage. We refit the probe on training-period anchors and score it on held-out training customers. Its 8-to-16-week gain on holiday-free anchors is then $0.72$ (95\% interval $[0.69, 0.75]$, covering score-side sampling only) of its gain on anchors in the test anchors' weeks of the year (the \emph{test calendar}), so by the probe's pre-stated rule the gain persists outside the holiday season (\S\ref{app:calendar}). This is a validation-style estimate from a linear probe, not from the transformer or the test set, so the share of the window gain due to history remains unidentified.

The gain from a longer window bends, and where it bends tracks how active the customer is.
At 24 weeks the measurement is closer to the saturating curve's prediction than to the log-linear curve's, missing the first by $0.005$ and the second by $0.066$.
We committed both falsifiable predictions before any 24-week evaluation.
The two forms
were near-tied on the data they were fit to, and the curvature inside the fit range had already strained the
log-linear form. The 24-week point therefore tests whether the better in-sample
form extrapolates, rather than testing two live hypotheses (\S\ref{sec:prereg}). A
weekly information curve built from a single cohort extract
(\S\ref{sec:results}) puts the pooled plateau at 22--24 weeks on the test calendar
(the marginal week is worth $+0.0001$ $R^2$ by $k{=}22$--$24$ weeks). When the probe is refit on training-period anchors, its returns diminish on both holiday-free and test-calendar anchors. The sharp bend persists on test-calendar anchors, scored on all their held-out customers, but is not shown to persist on holiday-free ones (\S\ref{app:calendar}) or on any public dataset of \S\ref{app:public-ladder} on which we assess the bend. The pooled law, however, is consistent with a
mixture, and decomposing it by customer activity yields a result on how to allocate history. The bend moves with customer
activity, sharply and monotonically across the top five deciles. The top activity decile saturates with a
half-saturation point of
$3.9$ weeks, while the bottom half of the cohort shows no detectable bend within 16 weeks and is
still climbing at the window edge. Higher-activity customers realize their predictive gains earlier, and lower-activity customers keep benefiting from added history over more of the measured range. Consistent with the
mixture reading inside the window, the pooled curve is visibly not a single exponential
(the fitted exponential underfits in the direction the per-decile mixture predicts). An
internally registered 32-week extension bounds the allocation result. Past 24 weeks, the per-decile refits find small remaining gains everywhere ($+0.012$--$+0.021$, with the lightest deciles at a third of the
mixture's predicted $+0.045$--$+0.059$), and the pooled curve is flat. The advice to allocate history by activity therefore applies only within the window. By ${\sim}6$ months, on the test calendar, the
weekly-aggregate probe finds little left in longer history at any activity level. On the nine-day public dataset of \S\ref{sec:external}, a
far sparser one, the saturating \emph{form} is favored on fit error
while the bend sits at or beyond the edge of the observable window.
An internally registered 32-week transformer test corroborates that the deep
model decelerates too. Its 24-to-32-week gain lands on the saturating prediction and misses the log-linear one.
That test ran at one seed and compared models trained on different cohort extracts (\S\ref{sec:miss}).
No single observation horizon is budget-efficient across
customer activity levels. Within the first few months, ``how much
history a customer needs'' is a per-decile quantity, predictable from activity and converging on a
saturating law. A uniform 24 weeks is still accuracy-optimal on this grid, so allocating history by activity pays off
in budget alone (\S\ref{sec:allocation}).

Against our rotary encoding as implemented, time representation produces the largest model-side improvement we measured. It clears
the customer-scale, capacity, and architecture bounds, and it composes with the window lever. MS-RoPE, a multi-scale variant of rotary positional encoding (RoPE)~\citep{su2021roformer}, constrains the rotary frequencies
to bands aligned with physical time (hours, days, weeks, years). Across three seeds, MS-RoPE gained $0.021$ in log-space $R^2$ over our integer-position
RoPE implementation. In all three rotary variants we compare, including the learnable control below, the coordinates of each rotary pair use different frequencies (\S\ref{app:pe}), so the gain over correctly paired RoPE remains unmeasured. MS-RoPE also sits above an
\emph{unconstrained} learnable-frequency control on 3-seed means, by a margin within the
control's own seed spread.
The MS-RoPE and objective gains are nearly additive:
stacking MS-RoPE with the best loss reaches
$0.6060{\pm}0.0006$ log-space $R^2$ (3-seed; $96\%$ of the additive gain). Against the stronger tabular baseline we ran (a fixed-hyperparameter
LightGBM gradient-boosted ensemble on six aggregate features, hereafter GBM), that stack widens the transformer's margin over it
by ${\approx}2.3\times$ on the matched cohort (${\approx}4\times$ full-split;
\S\ref{sec:baselines}). Explicit physical-time representation lifts the level at every window; it is not how the model exploits a longer window (\S\ref{sec:time}).

Our contributions:

\begin{itemize}[leftmargin=*,itemsep=2pt,topsep=2pt]
  \item \textbf{The observation window is the largest lever we measured.} At the 1M tier,
  all eight model families gain $+0.070$--$0.107$ log-space $R^2$ from a longer window (two of
  the eight ran with an inert time mechanism, \S\ref{app:baselines}). The gain replicates
  across all three population tiers in every family evaluated at all three. These
  figures pool three test anchors inside a single eight-week band, so window length and
  calendar coverage are not separated here (\S\ref{sec:limitations}). The direction
  replicates on a nine-day public dataset for linear and deep families, where longer windows also
  enlarge the eligible training set (\S\ref{sec:external}), and holds for a weekly-aggregate ridge
  probe on four public datasets that each span more than a year, and for the paper's transformer on all four,
  each read on its own (\S\ref{app:public-ladder}). On those four datasets, rows are held fixed across windows, so their gains do not come from more training rows.

  \item \textbf{Observed effects of scale, capacity, and architecture are much smaller than the
  window's; scale and capacity are unresolved at the precision of these experiments, and architecture effects are small but not zero.} On 3-seed means these three
  axes move the sequence families by $\leq 0.007$ (architecture at matched objective). We use $\leq 0.007$
  throughout as one band for these three axes, with the architecture bound taken at 16 weeks and the clocks
  as built. Each
  axis was varied alone, with the training recipe fixed within each family apart from the realized learning-rate schedule of the tier contrast, and measured
  in log space only. The architecture bound of $\leq 0.0052$ holds only at 16 weeks, on seed 42, on the full test split, with the clocks as built, and as a spread across five families
  that include the transformer. At 8
  weeks the decay-gated recurrent family starts $0.015$ below the transformer (seed 42). The three-seed detection floor
  for this class is $0.0059$, above the architecture band and within a factor of $1.2$ of the
  other two bounds. The capacity contrast's $0.006$ is unresolved at three seeds against its own
  $0.016$ equivalence margin, which is wider than the floor. We therefore claim only a large
  contrast in observed effect size relative to the window intervention. We do not claim
  equivalence among settings on these axes (linear-floor exception in \S\ref{sec:universal};
  \S\ref{sec:ceiling}, \S\ref{sec:history}, \S\ref{sec:lossnull}, \S\ref{sec:limitations}).

  \item \textbf{A matched-cohort framework for window comparisons.} Longer windows admit
  different test customers. Scoring every window on the same customer-weeks shows the
  raw 8-to-16-week deltas \emph{understate} the gain (matched $+0.096$ vs.\ full $+0.075$ at 1M;
  $+0.094$ vs.\ $+0.075$ at 10M; seed 42; \S\ref{sec:matched}).

  \item \textbf{An internally registered control that is void, and why.}
  In this end-anchored design, the magnitude attributable specifically to additional history
  cannot be separated from changes in calendar coverage. We committed a criterion before running
  the anchor-varying control (for two of the six settings, the shorter-window $R^2$ levels were already in hand)
  and ran it at two objectives and three seeds. It is \emph{void} in all six settings. Before
  the added block enters, the anchors differ in $R^2$ by $3.7$--$6.2\times$ the permitted spread.
  In every setting they are ordered identically by baseline difficulty and by calendar position,
  so the two explanations cannot be separated there. For the transformer runs, a calendar-balanced
  test set that would separate them cannot be recovered from this export. On training-period anchors,
  the weekly-aggregate ridge probe separates window length from holiday coverage in part: its
  8-to-16-week gain on holiday-free anchors is $0.72$ of its gain on the test calendar (\S\ref{app:calendar}).

  \item \textbf{An internally registered saturation analysis and a bounded allocation result.}
  The pooled weekly information curve saturates at 22--24 weeks on the test calendar. When the probe is refit on
  training-period anchors, its returns diminish on both holiday-free and test-calendar
  anchors. The sharp bend persists on test-calendar anchors, scored on all their held-out customers, but is not shown to persist on
  holiday-free ones (\S\ref{app:calendar}) or on any public dataset of \S\ref{app:public-ladder} on which we
  assess the bend. Within 24 weeks the most
  active customers saturate first (top-decile half-saturation point $3.9$ weeks), while the bottom half shows
  no detectable bend within 16 weeks, so the half-saturation point moves with activity. The pooled curve
  suggests a mixture, and inside the window, history is worth allocating inversely to
  activity. An internally registered 32-week extension bounds the law. The pooled
  weekly-aggregate probe is flat past 24 weeks (mixture prediction $+0.014$ falsified at
  $-0.0004$), and residual per-decile gains in the lightest deciles fall to a third of the
  mixture's extrapolation. For this probe and the test calendar, the pooled fit saturates at
  ${\sim}5$--$6$ months. The saturating form is favored on fit error on the nine-day public dataset
  (\S\ref{sec:saturation}, \S\ref{sec:miss}, \S\ref{sec:external}).

  \item \textbf{Against our rotary encoding as implemented, time representation is the largest model-side
  effect we measured, and its gain is recipe-specific.} Structured multi-scale rotary bands gain $+0.021$ over it (3-seed).
  They sit above an unconstrained learnable control on 3-seed means, by a margin within the
  control's own seed spread, and stack near-additively with the objective
  ($0.6060{\pm}0.0006$ stacked, 3-seed; \S\ref{sec:time}). Multi-scale physical time is prior
  art in several independent lines, and we claim no novelty for it (\S\ref{sec:related}). The
  gain is also specific to our training recipe and to log space. Retrained under a raw-target
  recipe and scored in that recipe's own statistic, it reverses sign (\S\ref{sec:external}).

  \item \textbf{A reproducibility protocol for multi-build experiment campaigns.}
  Before any result from a new build of the training environment is cited, that build must
  retrain a fixed reference run and reproduce it bit-exactly on each code path a reference
  run exercises. $R^2$ is reproduced to 15--17 digits, and three cited builds fall outside this check
  (\S\ref{sec:repro}). Primary-dataset transformer, recurrent and linear runs are deterministic
  end-to-end. The saturation predictions were committed, on internal and self-attested
  records, before their deciding evaluations ran (\S\ref{sec:results}).

  \item \textbf{A scored record of internally registered commitments, miss included.} We made
  four quantitative commitments with falsifiers before their deciding evaluations, and we
  disclose how blind each was. Three are hits under the qualifications
  stated above. The one fully blind prediction, a bet against the flat $k{=}22{\to}24$
  marginal already in hand, missed (Table~\ref{tab:prereg}).
\end{itemize}

\begin{table}[t]
\caption{Of the four internally registered commitments, only the fully blind one, the mixture tail (iii), missed. The other three are hits under the qualifications in the notes below. Each prediction and its
falsifier were committed before the deciding evaluation ran. The last column states what
was already known at commit time, since the commitments are not equally blind.
Two of the predictions, (ii) and (iv), are mechanical curve fits to
already-measured points with no free parameter choices. Of the other two, (i) is a threshold and (iii) is a
per-decile extrapolation with a calibration window pinned in advance. What each
registration adds is the discriminating criterion, fixed in advance (\S\ref{sec:prereg}).
$^\dagger$Commitment (ii) is a hit on levels, scored across a cohort shift. Scored on the
gain instead, the observation sits between the two predictions, nearer the saturating one; we report both scorings
(observed gain $+0.015$ vs.\ saturating $-0.008$ / log-linear $+0.053$; each predicted gain is that
form's $R^2(24)$ less the fit cohort's $k{=}16$ level, $0.4426$, so the negative saturating value is the
fit sitting below that point, not a predicted decline;
\S\ref{sec:prereg}).
The two forms were also nearly tied on the data they were fit to, and the losing one was
already strained by the curvature within the fit range, so this row tests whether the better
in-sample form extrapolates rather than adjudicating two live hypotheses
(\S\ref{sec:prereg}).
$^\ddagger$Commitment (iv) is scored as committed; the deciding measurement is
single-seed and carries a training-extract confound: an unmeasured
transformer-side offset like the $-0.033$ observed on the ridge sub-ladder would
put the observation near the log-linear prediction (\S\ref{sec:miss}).
\textsuperscript{\S}The observed intervals in this table are row-sampling uncertainty at the three
test anchors, not intervals over calendar anchors: the anchors sit in one
eight-week band with almost entirely coinciding forward label windows, so the calendar
direction carries close to a single effective draw and the bootstrap does not resample it.
Measured anchor by anchor on the Huber objective, the same eight-week extension is large at
the earliest anchor and worth approximately nothing at the latest. That spread is wider than
this row's interval, which does not cover it. Those per-anchor figures come from
a control that is void under its own criterion and are reported only as diagnostics
(\S\ref{sec:intro}).}
\label{tab:prereg}
\centering\footnotesize
\setlength{\tabcolsep}{4pt}
\begin{tabular}{@{}L{2.5cm}L{3.4cm}L{2.35cm}L{0.9cm}L{3.35cm}@{}}
\toprule
\textbf{Commitment} & \textbf{Prediction (falsifier)} & \textbf{Observed} & \textbf{Verdict} & \textbf{Known at commit time} \\
\midrule
(i) Composition survival &
matched $\Delta$(M16$-$M8) $\geq +0.05$ &
$+0.096$ {\scriptsize$[0.094, 0.098]$} &
Hit (${\sim}2\times$)\textsuperscript{\S} &
Bar set before the deciding transformer matched re-evaluation; full-set deltas (\mbox{${\approx}+0.075$}) and a matched ridge $+0.104$ that already cleared it were in hand. \\
\addlinespace[2pt]
(ii) Saturation form, 24 wk &
matched weekly-aggregate ridge (\S\ref{sec:saturation}), log-space $R^2$: saturating $R^2(24){=}0.4345$ vs.\ log-linear
$0.4952$ &
$0.4296$ {\scriptsize(off by 0.005 vs.\ 0.066)} &
Hit$^\dagger$ &
Forms fit before any 24-week evaluation; scored across a cohort shift, both scorings reported. \\
\addlinespace[2pt]
(iii) Mixture tail, 32 wk &
$\Delta(24{\to}32){=}{+}0.0142$ $[0.0126, 0.0231]$; falsifier $<0.007$ &
$-0.0004$ {\scriptsize$[-0.0014, +0.0004]$} &
\textbf{Miss} &
Fully blind: no 32-week job had ever run at commit. \\
\addlinespace[2pt]
(iv) Deep-model bend, 32 wk &
saturating $+0.012$ vs.\ log-linear $+0.041$; midpoint $0.026$ &
$+0.011$ {\scriptsize$[0.010, 0.013]$} &
Hit$^\ddagger$ &
Committed before any matched-cohort number existed for it; the confounded full-set value was already in hand. \\
\bottomrule
\end{tabular}
\end{table}

The primary evidence comes from one proprietary population, but its
1M/5M/10M tiers are customer-disjoint cohorts drawn from non-overlapping sampling-bucket
ranges (\S\ref{sec:data}). Cross-tier agreement on the window gain (all three tiers), its matched
form and the sequence-family scale bound (1M and 10M) is therefore replication across disjoint
customer cohorts of that population rather than resampling of one.
The nine-day external check speaks only to
the direction of the gain in linear and deep models and to its shape in ridge.
The deep model (MSE) replicates the direction under an internally registered three-seed test
on end-anchored 3/5/7-day extracts, with $\Delta$(D7$-$D3) $= +0.047$ (customer-clustered bootstrap interval
$[+0.035,+0.059]$) and all seeds positive. With seeds as the unit, the magnitude is unresolved at $n{=}3$ (\S\ref{sec:external}).
Absolute levels sit near zero and are never comparable across datasets. The M8/M12/M16 grid
points are 3-seed at all three scales, with per-seed ranges up to $0.0125$. The M24 points of the scale
ladder are 3-seed with wide per-seed spread (ranges $0.0094$/$0.0072$ at 5M/10M),
so the flatness there is only as precise as that spread. The 1M
transformer main results (including the
$0.6060$ stack) and the matched 1M Huber ladder (\S\ref{sec:matched}) are 3-seed, though the matched $+0.096$ and full-set $+0.075$ are seed-42.
The baseline-family rows (GRU/LSTM Huber, ContiFormer, Neural
Hawkes) are seed-42.

\begin{figure}[t]
  \centering
  \includegraphics[width=\linewidth]{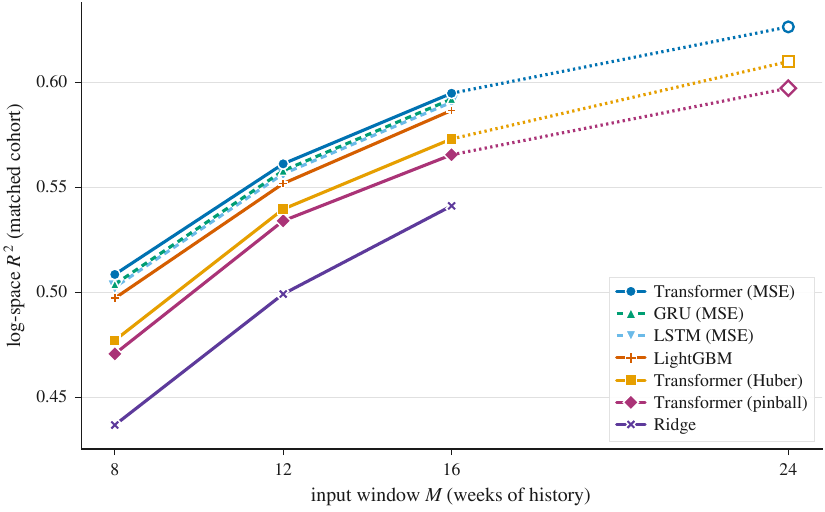}
  \caption{%
    \textbf{The input window is the lever, across families.}
    Every family scored on the matched cohort climbs $+0.086$ to $+0.104$ in log-space
    $R^2$ from M8 to M16, the plotted GRU and LSTM included at $+0.088$.
    At any fixed window, the spread across the plotted sequence architectures
    under a matched aggregate loss is an order of magnitude smaller than that climb.
    Among the plotted families trained with squared error, only ridge is meaningfully below the others.
    Scores use the matched cohort at the 1M tier, the customer-weeks present in each of
    the M8, M12 and M16 test sets ($n{=}407{,}338$). The three windows share these
    customer-weeks, so the $R^2$ denominator stays constant.
    Open markers again flag weaker evidence. Here they extend the
    transformer curves to M24 on the four-window matched cohort
    (M8$\cap$M12$\cap$M16$\cap$M24, $n{=}372{,}778$). The cohort changes at that step,
    so the dotted connectors to M24 do not trace a within-cohort trajectory.
    GRU and LSTM (dashed) are the recurrent families, retrained under the MSE transformer's objective and scored
    on the \emph{same} matched cohort. At M8/M12/M16 their 3-seed means are GRU
    $0.5039/0.5580/0.5921$ and LSTM $0.5025/0.5565/0.5906$, at the fixed training exposure
    of \S\ref{sec:repro}. Their full-test-split curves are the conservative counterpart
    (cf.\ Figure~\ref{fig:composition-defense}).
    Transformer matched points are seed-42 evaluations, so they
    sit slightly off the 3-seed matched means printed in
    Table~\ref{tab:window-grid}. At M8--M16, the matched level of each of the 3 seeds lies within
    $0.0025$ of the plotted value on the MSE curve and within $0.0111$ on the Huber ladder, whose seed spread
    is widest at M12 (\S\ref{sec:matched}). GBM seed SD is $\le 1.4{\times}10^{-4}$.
    Curves differ in training objective, so \emph{level} orderings across families
    are meaningful only among curves that share an objective. GBM optimizes squared error, and its lead
    over the Huber and pinball transformer curves reflects the objective rather than the model
    family. The MSE transformer shares that objective and exceeds GBM by
    $+0.008$ at M16, a residual family margin
    (\S\ref{sec:baselines}, Table~\ref{tab:objective-matched}).
  }
  \label{fig:window-lever}
\end{figure}

\begin{figure}[!tp]
  \centering
  \includegraphics[width=0.67\linewidth]{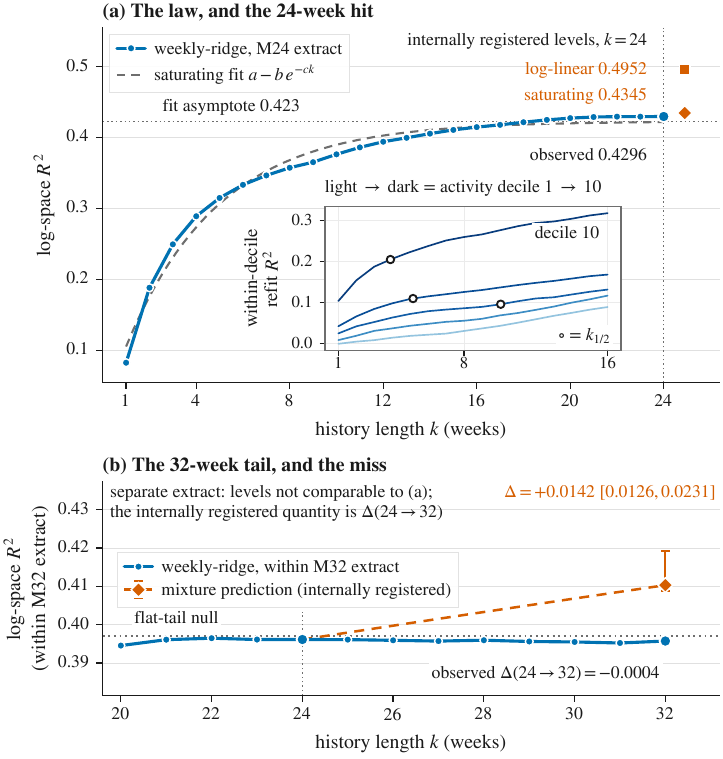}
  \caption{%
    \textbf{The saturating law's 24-week prediction hits, and the activity-mixture
    model's 32-week prediction misses. Both were registered internally.}
    \textbf{(a)} At $k{=}24$ the observed $0.4296$ lies $0.005$ below the saturating prediction and $0.066$ below the log-linear one, so both internally registered levels sit \emph{above} the outcome.
    The two forms were fit on the three-window matched ladder before any 24-week evaluation and are scored across a
    cohort shift (both scorings in \S\ref{sec:prereg}).
    The curve plots the log-space $R^2$ of the weekly-aggregate ridge probe (closed-form, no training seed)
    against history length $k{=}1..24$ weeks. Every point is scored on the single four-window matched cohort
    ($n{=}372{,}778$) and comes from the M24 extract alone, so no training-extract mismatch enters the curve.
    On the test calendar it flattens ($+0.00009$/week at $k{=}22{\to}24$), but a single
    exponential underfits it, with fit RMSE $0.0103$ and an asymptote of $0.423$, below
    the observed $0.4296$. A mixture of per-decile saturation rates predicts this underfit
    inside the window, and panel (b) rejects the mixture's extrapolation past
    $k{=}24$.
    Inset: the more active the customers, the sooner the curve bends.
    The top three deciles, 10, 9 and 8, reach the half-saturation point $k_{1/2}$ of their saturating fits (circles) at $3.9$, $5.2$ and $10.0$ weeks.
    In the fits, deciles 6--7 bend beyond the window ($k_{1/2}$ of $27.1$ and $24.7$ weeks), and the bottom five are still
    climbing at the window's edge. This is a claim about curve shape, not level, and it rests on the fitted
    parameters in \S\ref{sec:allocation}.
    Curves show per-decile \emph{refit} $R^2(k)$, $k{=}1..16$, from the M16 extract on the same matched cohort, shaded by
    prior-activity decile. Each (decile, $k$) pair has one ridge fit, evaluated on that decile's test rows.
    All ten deciles are fitted, but only deciles $1$, $5$, $8$, $9$
    and $10$ are drawn, because the lower eight lie within $0.051$ of one another
    at $k{=}16$ and their strokes would overlap.
    \textbf{(b)} The 32-week miss, drawn against the falsification region of the
    internally registered extension. This panel uses a separate M32 extract, scored on the matched-cohort rows
    that also fall in the M32 test split ($94.2\%$ coverage, $n{=}351{,}174$).
    Its level differs from panel (a) by $-0.033$, an offset between extracts in which the
    M32 training data and the scored rows change together. The internally registered quantity is
    therefore the gain within this extract, $\Delta(24{\to}32)$, anchored at the observed
    $k{=}24$ point.
    The activity-mixture model predicted $\Delta{=}{+}0.0142$
    $[0.0126,0.0231]$, the internally registered band and the
    only interval that panel (b) draws. Its lower endpoint is the calibrated and its upper endpoint the raw
    $90\%$ bootstrap bound, so it is not
    a single $90\%$ interval. The flat-tail null is $\le{+}0.001$.
    The observed gain is $\Delta{=}{-}0.0004$, with 95\% CI $[-0.0014, +0.0004]$ from the
    paired customer-clustered bootstrap on the same extract and rows.
    This observed interval reflects row-sampling uncertainty at the three
    test anchors, not variation across calendar anchors
    (\S\ref{sec:matched}). The
    curve enters the flat-tail null, and the observed interval excludes the entire internally registered
    range (discriminating threshold $0.007$). The mechanism check fails in the
    same direction. The three lowest deciles show a mean within-decile gain of
    $0.018$, against a predicted $0.045$--$0.059$.
  }
  \label{fig:prereg-arc}
\end{figure}

\begin{figure}[t]
  \centering
  \includegraphics[width=0.72\linewidth]{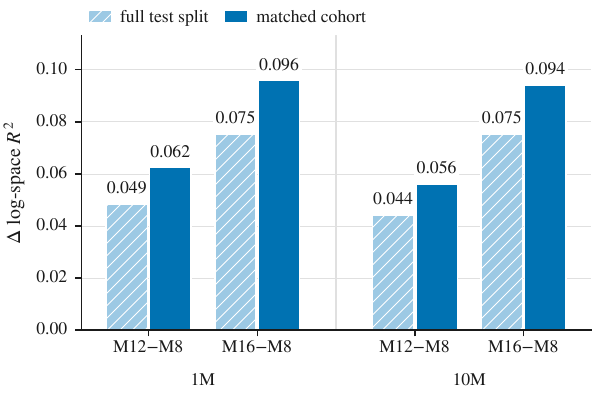}
  \caption{%
    \textbf{Through 16 weeks, cohort composition makes raw deltas understate the window effect.}
    Full-test-split vs.\ matched-cohort window deltas in log-space $R^2$
    (transformer, Huber, seed 42) at 1M and 10M.
    At both scales the matched delta exceeds the full-set delta for every
    window contrast (e.g.\ $\Delta$(M16$-$M8) $+0.094$ matched vs.\ $+0.075$ full at 10M).
    The gap arises because longer windows admit colder, less active customers, which changes both
    the error and the target variance in $R^2$. Here the net effect is that raw deltas between these windows
    \emph{under}state the window gain measured on a fixed population. The matched delta
    scores both windows on that fixed population, so it is unaffected by shifts in test composition. Bars are point estimates. Within seed 42, customer-clustered paired
    bootstrap intervals on the matched 10M deltas are tight.
  }
  \label{fig:composition-defense}
\end{figure}

\section{Results}
\label{sec:results}

\paragraph{Setup.}
All primary-dataset results except the 32-week extension (\S\ref{sec:prereg}) use the canonical cohorts. These are disjoint samples at three nominal population
tiers, with three test anchor weeks (the weeks a forecast is made from) and $M$ weeks of event history before
the anchor (M8/M12/M16/M24 $=$ 8/12/16/24 weeks). The 1M/5M/10M tier labels are sampling-bucket ranges, not customer counts
(\S\ref{sec:data}). The target is total activity over
the 364 days after the anchor, $\log(1{+}\max(0,y))$-transformed. Our primary metric is
log-space $R^2$ on the held-out test split
($n_{\mathrm{test}}{\approx}480$k per window at the 1M tier, ${\approx}2.4$M at 5M, ${\approx}4.8$M at
10M). We do not winsorize the log-space target.
The transformer base recipe is width d128, 2
layers, 20 epochs and rotary positional encoding, held fixed except where a swept lever
moves it. The rotary encoding is the implementation specified in \S\ref{app:pe}. The swept levers are input window,
training objective $\in\{\text{Huber},\text{MSE},\text{pinball}\}$, capacity, time representation,
training budget (\S\ref{sec:lossnull}), and scale. The time-representation sweep in \S\ref{sec:time} varies only the positional encoding. The training objective is the loss on the
aggregate prediction; every transformer run also optimizes a fixed auxiliary per-event term
alongside that loss (\S\ref{app:objective}).
Entries marked 3-seed report mean $\pm$ population SD over seeds $\{42,123,456\}$; single-seed
entries are flagged.
Every container build behind a cited result reproduces a fixed reference run bit-for-bit ($R^2$ identical to 15--17 digits), with three exceptions. One is the corrected-schedule build of \S\ref{sec:ceiling}, which was also used for the twelve-run scale retrains, the cross-evaluation and the four baseline families' three-seed runs of \S\ref{sec:limitations}. Another is the Taobao dataset's build. The third is the corrected-input build behind the working-clock baselines, the transformer runs on the other public datasets and the two capacity checks. Results from the last two are verified instead through the checks \S\ref{sec:repro} describes. The reference run exercises the transformer path and, on some builds, the recurrent path too. It does not exercise the continuous-time training path.
On the primary dataset, the transformer, recurrent and linear paths are bit-deterministic at fixed seed: independent
runs of the same configuration reproduced $R^2$ bit-identically in six groups of repeated runs (four transformer, one recurrent, one linear; \S\ref{sec:repro}).

\paragraph{Vocabulary and conventions.}
Recurring terms, defined once. An \emph{extract} is a fixed cohort sample cut from the
platform's raw event-stream export at a given window and
tier (\S\ref{sec:data}). A \emph{run} is one training or evaluation job at
one grid point (family, window, tier, objective) and one seed. A \emph{ladder} is a sweep
along one factor, and each of its settings (e.g.\ M12) is a point on it. An \emph{instrument} is the
measurement model from which a curve is estimated, e.g.\ the weekly-aggregate ridge probe of
\S\ref{sec:saturation}. A \emph{mismatch} is any difference in environment or data
between compared runs (container build, evaluation world size, training extract). A
factor is \emph{model-side} if it changes the model rather than how much history or which customers the model is shown, or
what it is asked to fit. Architecture, capacity, input content, and positional encoding
are model-side; the observation window, the cohort, and the objective are not.
Values listed as $x/y/z$ (or $x/y/z/w$) run M8/M12/M16 (M24) in that order
unless labeled otherwise.

\subsection{The ceiling at eight weeks}
\label{sec:ceiling}

With the input window fixed at 8 weeks, architecture, capacity and population tier move log-space $R^2$ very little. At M8 (1M),
four of five sequence families at the matched (Huber) objective land in a
narrow band ($0.4734$--$0.4776$): transformer
$0.4776{\pm}0.0013$ (3-seed), GRU
$0.4760$, LSTM
$0.4734$, time-biased attention (ContiFormer)
$0.4773$ (the three baselines at seed 42). The same transformer under pinball sits at
$0.4698{\pm}0.0007$, an objective effect rather than an architecture one
(\S\ref{sec:lossnull}). One family starts below that band. The decay-gated recurrent model (Neural Hawkes implementation) scores
$0.4619$ (seed 42), ${\sim}0.011$ below the band's low end, and converges as the window grows (\S\ref{sec:universal}).
Doubling width (d256) moves the transformer
$+0.003$, doubling depth (4 layers)
$-0.005$, adding product-category features
$+0.001$: all within or near single-seed noise.\footnote{The product-category
field, like every other categorical input to these transformer runs, reached the model
as one constant code on every event, so this ablation bounds nothing about content.
At 12 weeks the same single-seed ablation lands $0.013$ below the same-seed
transformer without the field, and at 16 weeks within $0.007$ of it.} Pushing capacity far past doubling
does not change that. At 16 weeks a $4\times$-wide, $4\times$-deep variant (d512, 8 layers,
effective batch held) scores
$0.5448{\pm}0.0038$ (3-seed), $0.006$ below the d128 3-seed mean
($0.5508{\pm}0.0016$), with the widest seed spread of the capacity variants. The gap's
interval is wider than this class's three-seed detection floor, so the gap is unresolved
and cannot be read as mildly negative.
That variant reuses the small model's learning rate, effective batch,
schedule, and epoch budget (the d128 recipe, Table~\ref{tab:recipe}). It reaches the same per-rank
effective batch through gradient accumulation and lengthens the warmup from $100$ to
$500$ steps. It therefore tests capacity \emph{at a
fixed recipe}, not compute-optimal scaling, which would require
re-tuning the larger model. A five-configuration retune of that variant did not
resolve the gap in either direction. The retune tried three learning rates bracketing the
inherited one, a longer warmup, and a joint increase in dropout and weight decay.
The sign of this gap does not reproduce in a corrected re-run. The reported capacity runs used the
constant-rate schedule and the original-scale checkpoint selection, both described in
\S\ref{sec:limitations}. With both corrected, the same capacity contrast comes out at
$+0.0004$, three seeds per model. That re-run used a different build, so it is not pooled with the results above.
The two estimates give the same bound on the gap's size but differ in sign, so the gap stays
unresolved.
A five-point learning-rate sweep of the larger model on corrected inputs leaves the capacity bound standing
(\S\ref{sec:repro}).

The population tier moves the metric very little, and not monotonically. The
transformer, which streams each tier's full training set, stays inside a narrow band
from 1M through 10M
($0.4776 \to 0.4731 \to 0.4746$, 3-seed at every tier; spread $\leq 0.005$). At M8 the recurrent baselines stay within $0.002$ across tiers: GRU
scores
$0.4760/0.4744/0.4751$ at 1M/5M/10M, LSTM
$0.4734/0.4717/0.4730$, both seed 42. Unlike the transformer, these baselines train from a fixed-size cached draw at every tier
(\S\ref{sec:repro}), so their rows measure population composition at fixed exposure,
not response to more data. At 1M, lifting that size cap (transformer budget, matched MSE
objective) moves them by at most $+0.002$
(\S\ref{sec:lossnull}). This scale bound is a \emph{sequence-model} result. The
6-feature linear floor is the one family that breaks it. At 10M its level dips $-0.028$
at M8, narrowing to $-0.016$ by M24 (\S\ref{sec:universal}), yet its window gain at that
tier is the steepest of all.

At eight weeks the training objective moves $R^2$ the most. Under MSE the transformer reaches
$0.5075{\pm}0.0010$, against $0.4698$ under pinball. Even that spread (${\sim}0.04$)
is smaller than the window gain that follows. Eight weeks looks like a ceiling for this task.
Neither more customers nor more parameters lifts $R^2$ past ${\sim}0.50$.

\subsection{The observation window is the largest lever we measured}
\label{sec:history}

That ceiling comes from the eight-week conditioning set. Extending the input window
moves log-space $R^2$ more than every architecture, scale and capacity
intervention we measured combined (Table~\ref{tab:window-grid}, Figure~\ref{fig:window-lever}; per-lever contrasts in Table~\ref{tab:bounds}).
It also moves $R^2$ more than either of the next two strongest levers, the objective (\S\ref{sec:lossnull})
and time representation (\S\ref{sec:time}), taken singly. Tiers, anchor weeks, target, and
recipe stay the same. Only the window grows, and with it the calendar months it covers
(\S\ref{sec:limitations}) and the eligible training and test populations
(Table~\ref{tab:window-grid}; \S\ref{sec:matched}). A validation-style check partly separates window length from holiday coverage. On training-period anchors with an 80/20 customer split (\S\ref{app:calendar}), the 8-to-16-week gain of the weekly-aggregate ridge probe (not the transformer) outside the holiday season is $0.72$ of its gain on anchors that share the test calendar (95\% interval $[0.69, 0.75]$, covering score-side sampling only). The remainder may still be calendar-related.

\begin{table}[t]
\caption{From M8 to M16, the transformer's log-space $R^2$ climbs under all three training objectives. The climb is larger on the matched cohort than on the full split. Input-window sweep at the 1M tier, scored on the full test split (raw
rows) and on the matched cohort (fixed population; \S\ref{sec:matched}). Transformer full rows are 3-seed mean $\pm$ population SD. The
MSE and Huber matched M8--M16 cells are 3-seed mean $\pm$ population SD; the pinball
matched row, the matched M24 entries (four-window cohort), and the matched
$\Delta$(M16$-$M8) column are the
seed-42 paired evaluations ($^{\dagger}$ on the MSE and Huber rows; clustered CIs in \S\ref{sec:matched}); the \emph{matched}
$\Delta$ column is that seed-42 pairing and need not equal the difference of the printed
3-seed cells, while the full-row $\Delta$ values are 3-seed paired means.
Raw M24
cells are denominator-inflated by cohort composition and are not comparable to the M16 column (see text); the
matched rows are the like-for-like comparison. Their M8--M16 cells sit on the three-window
cohort ($n{=}407{,}338$ customer-weeks), not the four-window cohort the matched M24 entries
use ($n{=}372{,}778$): no $\Delta$ in this table crosses that cohort change, and
re-scoring the seed-42 M8--M16 models on the four-window cohort moves those levels by
${\leq}0.0004$ and $\Delta$(M16$-$M8) by ${\leq}0.0002$.
\emph{Train-side eligibility.} The longer-window models also train on a
modestly larger and colder eligible population (up to $10\%$ more customers,
\S\ref{sec:data}). Matching fixes test composition but not this, so every window contrast here is ``window as extracted''
(\S\ref{sec:limitations}).}
\label{tab:window-grid}
\centering\footnotesize
\setlength{\tabcolsep}{4pt}
\begin{tabular}{llccccc}
\toprule
\textbf{Objective} & \textbf{Split} & \textbf{M8} & \textbf{M12} & \textbf{M16} & \textbf{M24} & $\Delta$(M16$-$M8) \\
\midrule
Huber & full (3-seed) & $0.4776{\pm}.0013$ & $0.5237{\pm}.0053$ & $0.5508{\pm}.0016$ & $0.5885{\pm}.0015$ & $+0.0731$ \\
Huber & matched (3-seed) & $0.4777{\pm}.0012$ & $0.5368{\pm}.0058$ & $0.5716{\pm}.0012$ & $0.6100^{\dagger}$ & $+0.0959^{\dagger}$ \\
\addlinespace
MSE & full (3-seed) & $0.5075{\pm}.0010$ & $0.5557{\pm}.0010$ & $0.5875{\pm}.0007$ & $0.6242{\pm}.0005$ & $+0.0800$ \\
MSE & matched (3-seed) & $0.5075{\pm}.0010$ & $0.5606{\pm}.0007$ & $0.5945{\pm}.0004$ & $0.6265^{\dagger}$ & $+0.0863^{\dagger}$ \\
\addlinespace
pinball & full (3-seed) & $0.4698{\pm}.0007$ & $0.5178{\pm}.0012$ & $0.5400{\pm}.0025$ & $0.5746{\pm}.0006$ & $+0.0701$ \\
pinball & matched (seed 42) & $0.4707$ & $0.5341$ & $0.5656$ & $0.5973$ & $+0.0948$ \\
\bottomrule
\end{tabular}
\end{table}

\begin{table}[t]
\caption{Observed change in log-space $R^2$ per lever, one factor at a time: point contrasts at the
recipes tested, not equivalence bounds (transformer rows: rotary encoding as implemented, \S\ref{app:pe}).
Window rows also move calendar coverage (\S\ref{sec:limitations}) and training eligibility (\S\ref{sec:data}). Population-tier
rows compare customer-disjoint tiers, so they bound cohort and size together, not size alone. The
${\approx}54\times$ capacity contrast's sign does not reproduce in a corrected re-run on a different build
($+0.0004$, three seeds per model; \S\ref{sec:ceiling}), so its sign is unresolved. The four baseline families
of the three-seed architecture row did not share a peak learning rate (\S\ref{sec:limitations}). The time-representation row and both architecture rows are scored on the full test split.}
\label{tab:bounds}
\centering\footnotesize
\setlength{\tabcolsep}{4pt}
\begin{tabular}{@{}lllll@{}}
\toprule
\textbf{Lever} & \textbf{Contrast} & \textbf{Seeds} & \textbf{Observed $|\Delta R^2|$} & \textbf{Source} \\
\midrule
Window & M8$\to$M16, matched, transformer (1M / 10M) & 3 (mean) & $+0.0939$ / $+0.0934$ & \S\ref{sec:matched} \\
 & M8$\to$M16, full split, eight families (1M) & mixed & $+0.070$ to $+0.107$ & \S\ref{sec:universal} \\
Objective & Huber / MSE / pinball, transformer, M8--M24 & 3 (mean) & $0.03$--$0.05$ & \S\ref{sec:lossnull} \\
Time representation & MS-RoPE vs.\ RoPE, transformer, M16 & 3 (mean) & $+0.0208$ & Table~\ref{tab:pe} \\
Population tier & 1M$\to$10M, transformer, M16 & 3 (mean) & $0.0025$ & Figure~\ref{fig:missing-axis} \\
 & 1M$\to$10M, transformer, each window M8--M24 & 1 (seed 42) & $0.006$ & Table~\ref{tab:scale-10m} \\
Capacity & $2\times$ width or $2\times$ depth, M8 & 1 (seed 42) & $0.005$ & \S\ref{sec:ceiling} \\
 & $4\times$ width and depth (${\approx}54\times$ parameters), M16 & 3 (mean) & $0.006$ & \S\ref{sec:ceiling} \\
Architecture & five sequence families, matched objective, M16 & 1 (seed 42) & $0.0052$ & \S\ref{sec:baselines} \\
 & four baseline families, M16 & 3 (mean) & $0.0059$ & \S\ref{sec:limitations} \\
\bottomrule
\end{tabular}
\end{table}

Doubling the observation window raises $R^2$ by $+0.070$--$0.107$ in every family at 1M, while going from the 1M to the 10M tier moves
sequence-family levels by ${\leq}0.006$ (seed 42; \S\ref{sec:universal}).
For the transformer at 1M, the three-seed M8-to-M16 climb is $+0.0701$--$+0.0800$ log-space $R^2$
by objective. It is monotone at every objective and seed, and an order of
magnitude above seed dispersion. M24 adds a further
$+0.035$--$+0.038$ raw, at a visibly decelerating per-week rate
(\S\ref{sec:saturation}). The raw M24 column is denominator-inflated by cohort
composition
(full-set zero-fraction $11.9\%$ vs.\ $4.8\%$ matched; target log-SD $2.80$ vs.\ $2.16$),
which is why Table~\ref{tab:window-grid} also reports matched rows.

\emph{What the window carries.} No elapsed time reaches the network in these runs. The two
declared per-event time channels (day of week and week index) are
identically zero on every event, and the base positional encoding is rotary over event
\emph{order}, so the per-event timestamp that the extract does contain is never read
(\S\ref{app:arch}). What grows with the window is therefore how many events the model sees
and in what order, not the gaps between them. Even though the network reads no inter-event timing, the window is the largest axis on
this task. \S\ref{sec:time} gives a consistent check from the
other direction: the physical-time encoding (MS-RoPE) raises the
log-space level at every window, yet contributes only a small fraction of the M8-to-M24
climb on the full test split (seed 42). In log space it shifts the level and is not the mechanism behind the
window gain.

\subsection{Window gains survive cohort matching}
\label{sec:matched}

From M8 to M16, holding the test population fixed enlarges the window gain in $R^2$ for every family we tested.
The raw climb has a composition confound, because longer windows admit different customers.
Eligibility widens with the window, and longer windows
reach lapsed customers whose only activity sits 57--112 days back. In the
$15\%$ of the M16 test set outside the shared population, $30.4\%$ of rows have a zero target, compared with $4.8\%$ in the shared population
(median 6 events vs.\ 60 at that window). Full-set target log-SD grows
$2.16 \to 2.39 \to 2.56$ across windows. Composition alone adds $+41\%$ target
variance at M16, inflating the raw $R^2$ denominator.
The composition shift runs in both directions: each window's split
also contains rows that another lacks, in nearly equal numbers each way
(\S\ref{sec:data}). The matched cohort is therefore the intersection of all three splits rather than the
smallest of them, and matched results describe that intersection
(\S\ref{sec:limitations}).

To control for composition, we re-evaluate on matched cohorts. The three-window cohort holds the
407{,}338 customer-weeks present in all three M8/M12/M16 test sets ($84.7$--$84.8\%$ of each),
keyed on (customer, target week). Adding M24 gives a four-window cohort of 372{,}778 rows. Targets are
numerically identical across the three windows
(max discrepancy $8{\times}10^{-8}$). Because the population is fixed, the denominator is constant
(target log-SD $2.1597$ in all three windows), so climbs are pure error reduction
on the same customers. Every row of both cohorts has at least one event in the 8 weeks before its target week, so the matched cohorts contain only recently active customers and a matched gain is a gain for them. Most rows that matching drops have no such event (\S\ref{app:cohort-keeps}). The full-split deltas describe the whole test split. From M8 to M16, for every family tested the matched climb exceeds the raw climb.
For the transformer, the matched $\Delta$(M16$-$M8) is $+0.0959$ vs.\ full-set $+0.0731$ under Huber,
$+0.0863$ vs.\ $+0.0800$ under MSE, and $+0.0948$
vs.\ $+0.0701$ under pinball (matched deltas from seed 42; full deltas are 3-seed means, Table~\ref{tab:window-grid}). The ridge
instrument agrees ($+0.1044$ vs.\ $+0.0992$), and its matched log-space test MSE drops monotonically,
$2.627 \to 2.336 \to 2.140$ (Table~\ref{tab:matched-cohort}). Each of the three per-objective matched deltas
carries a paired, customer-clustered bootstrap CI ($B{=}1000$).
For $\Delta$(M16$-$M8) these are
$[0.0940, 0.0978]$ (Huber), $[0.0846, 0.0879]$ (MSE), and $[0.0929, 0.0968]$ (pinball),
all bounded far from zero. The composition
confound is real, but at the $R^2$ level it \emph{deflates} the window gain, so the raw
delta understates it. The matched climb clears our internally registered survival threshold
($+0.05$) by ${\sim}2\times$.

The result replicates at 10M. We re-score the 10M checkpoints on the 10M matched
cohort
($n = 4{,}069{,}475$, $84.8\%$ of each window; matched target log-SD $2.16175$ throughout):
matched $R^2 = 0.4757 / 0.5318 / 0.5698$ at M8/M12/M16 (seed-42).
Every window contrast is composition-deflated, as at 1M (Figure~\ref{fig:composition-defense}):
$\Delta$(M16$-$M8) $+0.0941$ $[0.0935, 0.0948]$ matched vs.\ $+0.0753$ full,
$\Delta$(M12$-$M8) $+0.0561$ $[0.0555, 0.0566]$ vs.\ $+0.0442$, and $\Delta$(M16$-$M12)
$+0.0381$ $[0.0375, 0.0386]$ vs.\ $+0.0311$.
The 10M matched ladder also has 3 seeds: matched $\Delta$(M16$-$M8)
$= +0.0934 \pm 0.0005$ (3-seed, per-seed clustered CIs exclude zero), in line with the 1M
ladder's $+0.0939 \pm 0.0022$. With test composition held fixed, the window gain is similar across tiers at three seeds under Huber.
All matched-delta CIs in this paper are paired, customer-clustered bootstraps unless noted
($B{=}1000$).
On the primary population, every such interval measures the uncertainty from sampling test rows \emph{at the three test anchors}, with the trained models held fixed. It does not cover variation across training seeds or calendar anchors. Those anchors sit inside one
eight-week band, and their forward label
windows almost entirely coincide. The calendar dimension therefore contributes close to a single
effective draw, and the bootstrap does not resample it. Where a quantity varies by anchor,
as the window gain does (\S\ref{sec:intro}), that variation sits outside every
interval we report.

For the ridge model, the bottom three deciles of input-window activity carry $69.8\%$ of the matched
M16-vs-M8 squared-error reduction (the lowest decile alone $40.3\%$). An eight-week window starves
low-activity customers of signal, and extending the window helps those customers most.

The matched 1M Huber ladder holds across seeds. Re-scoring the seed-123 and seed-456 checkpoints
gives matched $\Delta$(M16$-$M8) $= +0.0939 \pm 0.0022$ (3-seed;
every seed's customer-clustered CI excludes zero), with each window's 3-seed mean within $0.004$ of its
seed-42 value. The composition deflation is therefore stable across seeds rather than a
seed-42 artifact. Matched $\Delta$(M24$-$M8) $= +0.1313 \pm 0.0022$ on the four-window
cohort (3-seed; every seed's clustered CI excludes zero).

\begin{table}[t]
\caption{Matched-cohort evaluation of the linear model (6-feature ridge, 1M): full-test-set vs.\
matched $R^2$ with the constant-denominator MSE column (log-target space). The matched
climb exceeds the raw
climb.}
\label{tab:matched-cohort}
\centering\small
\begin{tabular}{lccc}
\toprule
\textbf{Window} & \textbf{Full $R^2$} & \textbf{Matched $R^2$} & \textbf{Matched MSE (log)} \\
\midrule
M8  & $0.4373$ & $0.4369$ & $2.627$ \\
M12 & $0.4962$ & $0.4993$ & $2.336$ \\
M16 & $0.5365$ & $\mathbf{0.5412}$ & $2.140$ \\
\bottomrule
\end{tabular}
\end{table}

\subsection{An internally registered saturation analysis}
\label{sec:saturation}

To test where the window gain stops, we committed falsifiable predictions \emph{before} any 24-week evaluation existed, registered internally rather than in a public registry. We fit a saturating-exponential
and a log-linear form to the weekly information curve, measured by a nested ridge on $k=1..16$ weeks
of weekly aggregate features, with matched-cohort denominators (\S\ref{app:probe}).
The two forms were nearly tied on the $k{=}1..16$ data they were fit to, yet they predict matched
$R^2(24) = 0.4345$ and $0.4952$, a $0.06$ separation at $k{=}24$. One M24 measurement
therefore tests whether the better in-sample form extrapolates (\S\ref{sec:prereg}).

The M24 measurement sits closer to the saturating prediction than to the log-linear one (Figure~\ref{fig:prereg-arc}a). Matched
weekly-aggregate ridge
$R^2(24) = 0.4296$ is within $0.005$ of the saturating prediction and $0.066$ from the log-linear one.
The internally registered forms were fit on a different cohort, and the original curve
spliced weekly features from two extracts at $k=16$. For both reasons, we rebuilt the curve
from the M24 extract alone: all of $k=1..24$, one training set and the matched
four-window test cohort, with no splice. The single-extract curve is
$R^2(k) = 0.0825/0.3574/0.4146/0.4273/0.4296$ at $k = 1/8/16/20/24$, monotone at the
printed points and smooth
through the old extract boundary ($+0.0032$ at $k=16{\to}17$). The level
offset in the earlier two-extract curve was a composition shift between the two training extracts. The marginal week is worth
${\sim}{+}0.0001$ $R^2$ by $k = 22$--$24$, down from $+0.0038$ at $k = 16$ on the
three-window-cohort curve that the registered forms were fit to. Here too
the saturating form beats log-linear
(fit RMSE $0.0103$ vs.\ $0.0162$), but
the single exponential \emph{underfits}: its fitted asymptote ($0.4227$) sits below the
observed $k{=}24$ point ($0.4296$). A pooled curve mixing per-customer saturation
rates has a heavier tail than any single exponential, which is what the
per-decile analysis finds (\S\ref{sec:allocation}). The misfit is what a mixture
would predict, not evidence against the saturating law.
On the test calendar, the pooled weekly information curve saturates at roughly 22--24 weeks, so this pooled weekly-aggregate probe benefits from five to six months
of history on that calendar. That is a measured
ceiling for this instrument, not a claim that weekly aggregates of this
behavior are exhausted. Refit on training-period anchors, the probe shows diminishing returns on both holiday-free and test-calendar anchors. The sharp bend
persists on the test-calendar anchors when scored on all their held-out customers, but it is not shown to persist on the holiday-free ones (\S\ref{app:calendar}) or on any public
dataset of \S\ref{app:public-ladder} where the bend can be estimated. For a team building on this class of data, the first step is to lengthen the window
to at least this saturation point (\S\ref{sec:discussion}).

\subsection{The prediction we failed}
\label{sec:miss}

The pooled curve does not keep climbing past this plateau, and this is where our one fully blind prediction failed. The test was an internally registered
extension to a 32-week
extract (\S\ref{sec:prereg}). Before running it, we committed to the mixture account's
prediction of a $+0.0142$ pooled gain from week 24 to 32, with any gain $<0.007$ counting as a falsification.
The observed gain was $-0.0004$
(95\% CI $[-0.0014, +0.0004]$, customer-clustered bootstrap; Figure~\ref{fig:prereg-arc}b).
The interval excludes the
entire internally registered range $[0.0126, 0.0231]$, and its upper bound sits below the
$0.007$ falsification threshold. The mixture
extrapolation is rejected by its own falsification criterion, and the null of a saturated pooled curve survives: weeks 25--32 add nothing to the pooled
weekly-aggregate ridge instrument, which plateaus at $k\approx 21$--$22$ on the test calendar's M32-extract ladder.
The plateau is not an artifact of our
extraction pipeline. A second extract of the same export, built by a separate pipeline, reproduces the $k{=}1..32$
ladder verdict on its own training and test splits. Three criteria, committed before scoring, all hold:
the saturating fit beats the log-linear one (RMSE $0.011$ vs.\ $0.020$), the
tail is flat ($\Delta(24{\to}32) = -0.0005$, below the $+0.007$ threshold), and
the late marginal is below the committed $20\%$ of the $k{=}16$ marginal. That is a pipeline replication on
the same export, not an independent dataset.

The transformer also shows diminishing returns. Its matched-cohort marginal gain
per week falls over its first three steps,
$+0.0156 \to +0.0083 \to +0.0046$ $R^2$/week across
M8$\to$\penalty300 M12$\to$\penalty300 M16$\to$\penalty300 M24 (seed 42), on the four-window matched cohort
($n{=}372{,}778$). The fourth step, M24$\to$M32, gives $+0.0014$ $R^2$/week on the narrower M32 cohort and across the training-extract mismatch described below. Over the same window range the transformer slows but is still climbing, on training cohorts that grow with window length. Its matched M24 level
($0.6100$, seed 42) sits far above the weekly-aggregate ridge asymptote of this two-feature-per-week probe. That comparison crosses instruments and objectives, but the gap is roughly four times the largest objective effect we
measure, so we take it to be robust to that change. The transformer's observed matched within-cohort gain is
$\Delta(24{\to}32) = +0.011$ (customer-clustered 95\% CI $[0.010, 0.013]$; single seed).
It lands on the saturating prediction ($+0.012$) and excludes the log-linear one
($+0.041$). The gain is scored on the four-window${\cap}$M32 evaluation cohort ($n{=}351{,}174$, $94.2\%$ of the four-window cohort). Commitment (iii) of Table~\ref{tab:prereg}
scores the pooled-probe verdict, and commitment (iv) scores this result (\S\ref{sec:prereg}). One confound qualifies this single-seed contrast. Its two checkpoints were trained on different extracts, the 24-week model on the M24 training extract and the 32-week
model retrained on a re-extracted M32 training set, so the
window is not the only factor that changed. On the ridge sub-ladder, the M32 training extract
and the cohort mask together shifted levels by an offset of about $-0.033$. On a transformer-side check, the mask
alone contributed $+0.0003$. We have not measured the combined offset on the transformer. A transformer-side offset of that size would
put the true within-cohort gain near the log-linear prediction. Commitment (iv) is thus a hit as committed, but an unresolved one. This mismatch does not affect
commitment (iii), whose pooled-probe verdict comes from a single extract. One check avoids the
mismatch but cannot measure it. A within-extract truncation control keeps the
same 32-week checkpoint and training extract, and scores the checkpoint on the same rows with
inputs cut at evaluation to the most recent 24 weeks. That control loses $0.023$ $R^2$
(customer-clustered 95\% CI $[0.021, 0.024]$), so the gain is not solely an artifact of the
training extract. That drop is about twice the $+0.011$ within-cohort gain, and evaluation-time
truncation adds a distribution shift of unknown
bias direction. The control therefore
corroborates the gain without sharpening the estimate. The saturating fit's asymptote is $0.629$ against a fitted $R^2(32)$ of
$0.622$. That leaves about $0.007$ of fitted headroom beyond 32 weeks on seed 42, on par with
our bounds on scale, capacity and architecture.
Neither the asymptote nor the fitted $R^2(32)$ is a measurement. Both come from one fit of the registered three-parameter saturating form to the four seed-42 matched estimates at $8$, $12$, $16$ and $24$ weeks. The headroom is what that form implies from those four
points, and it does not bound what a longer window could buy.

\subsection{Where the history goes: the bend moves with activity}
\label{sec:allocation}

The more active the customers, the sooner the weekly curve bends. We refit the weekly ladder
within activity deciles of the matched cohort, one ridge per (decile, $k$). Each ridge is fit on
training rows binned by input-window activity and evaluated within the matching test decile, so
every curve is calibrated to its own subpopulation. Across the top five deciles the saturation rate is monotone in
activity.
For the top decile the half-saturation point is $k_{1/2} = 3.9$ weeks, rising through
$5.2$ and $10.0$ weeks over the next two, then past the 16-week window ($24.7$ and $27.1$ weeks) for the two below them. The bottom half of the
cohort shows no detectable bend within 16 weeks; its fits degenerate to near-linear,
with curves still climbing at the window edge. These activity-stratified curves show heterogeneity but do not establish a mixture law. The within-decile claim is about shape,
not level (within-decile $R^2$ is much lower than pooled, since activity itself
absorbs most between-customer variance). Within roughly four months of history, how much history a
customer needs is predictable from their activity, and context is worth allocating
inversely to it: the customers who need the most feature weeks are the ones a fixed short
window starves (\S\ref{sec:matched}).
A process-identification account of context
offers one explanation: longer windows reduce uncertainty about which process
generated the sequence, so reaching the minimum attainable error can require a window
longer than that process's memory length~\citep{butera2026why}. We conjecture that a
more active customer's generating process is identified from fewer weeks; the per-decile
ordering above is qualitatively consistent with that.
The 32-week extension bounds this prescription rather than extending it.
The same per-decile refit finds week-24-to-32 gains that are small
but \emph{positive in every decile}, the heaviest included
($+0.012$--$+0.021$; every decile's customer-clustered 95\% CI excludes zero, including the
top three at $+0.016/{+}0.015/{+}0.012$). The lowest three deciles average $+0.018$ against
the mixture's predicted $+0.045$--$+0.059$. The lightest deciles still gain slightly more
than the heaviest, but the magnitudes fall to a third of the mixture's extrapolation.
The steep late climb predicted for the light tail did not appear. Every decile's own refit still gains while the pooled curve is flat. The saturation therefore lies in how one shared model allocates its fit across activity deciles, not in exhausted or corrupted signal within any subgroup. Under a single pooled model, the
within-decile week-24-to-32 change is significantly \emph{negative} for the mid-activity
deciles 4--8 (peaking at $-0.034$ for decile 6) while the two heaviest deciles are
statistically flat. The harm thus forms an inverted U that peaks mid-distribution rather than growing
toward the heavy end, and it vanishes once each decile gets its own model. It is a
different signature from the inverse context-scaling reported for long-horizon forecasting of
stochastic time series~\citep{ahuja2026retrieval}, where added context degrades the forecast itself.
It is also a different signature from the length-heterogeneity account in recent
click-through-rate recommendation work, where raising a model's maximum input sequence
length degrades short-sequence users through attention polarization and length imbalance in
the training data~\citep{zhang2026lain}. Their length-imbalance term is itself a form of
shared-model fit allocation, so this half of their account is a candidate mechanism within ours
rather than a rival to it. The shapes, however, differ. Under a single
pooled model the harm here peaks in the middle activity deciles, while their account places it on the users with
the shortest histories. The two analyses also stratify users differently, since our deciles bin
input-window activity and their account turns on sequence length.
Allocation-by-activity is a within-window
prescription: it holds where curves are still separating, and by roughly six months
of history the marginal week is worth ${\sim}0.003$ $R^2$ or less in every decile on the 32-week per-decile refit
instrument.

Read as a budget rule, the prescription retains about $99\%$ of the pooled-of-decile-refits instrument's pooled
log-space $R^2$ ($0.595$ of $0.601$) at $49.2\%$ of the uniform-$k{=}24$ feature-week budget (about $99.9\%$
at $89.2\%$) when its per-decile refits are scored jointly on the matched cohort. The rule routes each customer into an
activity decile, and each decile gets only as many recent
feature weeks as still pay for themselves in pooled $R^2$; here that allocation runs broadly inversely to activity.
The routing itself uses each customer's
full-window activity total, which the budget does not count. Routing alone (zero
feature weeks) already retains $90.0\%$, and one week per decile retains $91.2\%$, so the
allocation closes the last tenth (the routing-to-full-history gap), not the whole curve.
These values are internal to the pooled-of-decile-refits instrument (uniform-$k{=}24$
$=0.601$), a different instrument from the single pooled ridge elsewhere in the paper
($k{=}24=0.430$); the level gap between the two is the decile-routing gain on the same weeks.
The allocation is found by a greedy search on the same matched test rows that score it, so
both figures are optimistic estimates, and the first clears $99\%$ by only
${\sim}1{\times}10^{-4}$ $R^2$.
The payoff is a smaller feature-week budget with no accuracy gain. On this grid every decile's curve rises monotonically
through $k{=}24$, so the per-decile oracle coincides with uniform-$k{=}24$. Operationally, the budget rule needs
a single long-window aggregate to route on, plus a few weeks of detailed features for heavy
users and the full window for light ones. Behavior-conditioned forecasting instead conditions one shared neural
process on a latent class inferred per load profile~\citep{soleimani2026bcnp}.

\subsection{Every trained family gains from a longer window}
\label{sec:universal}

All eight families climb (Table~\ref{tab:family-deltas}).
The eight are trained families, not distinct
mechanisms. With an inter-event interval of identically zero, the two continuous-time models
reduce to a plain causal attention stack and a standard LSTM cell
(\S\ref{app:baselines}), so in the main runs they duplicate mechanism classes already in the
table rather than adding two new ones. At 1M,
$\Delta$(M16$-$M8) runs from $+0.070$ for the causal attention variant (ContiFormer stand-in) and the transformer under pinball to
$+0.107$ for a single-feature linear model ($a\,x + b$ on the input-window activity
total). Broadly, the
less a family can represent, the more the window is worth to it. The pattern is clearest in the two linear
rows: ridge and the single-feature model gain $+0.099$ and $+0.107$ against \mbox{$+0.070$--$+0.085$}
for the four event-level baseline families, and stay ahead of the sequence models at matched
objective (Table~\ref{tab:losseq}). The boosted trees' $+0.084$ sits inside the event-level band. The ordering is not monotone inside the sequence block. No single feature
is needed for the tabular climb. Dropping any one of the six keeps the
boosted-tree matched gain at
$+0.090$--$+0.098$ (seed 42), while the input-window activity total \emph{alone} delivers
$+0.110$. Here the window gain comes from more weeks of signal rather than from richer features.

\begin{table}[t]
\caption{The window lever in every baseline family: $\Delta$(M16$-$M8) log-space
$R^2$ at 1M, full test split unless marked matched. Transformer deltas are in
Table~\ref{tab:window-grid} ($+0.0701$--$+0.0800$, 3-seed, by objective). Seeds and objectives as
flagged in the text: recurrent and continuous-time rows are seed-42 runs under the Huber
recipe; GBM is a 3-seed fit; the linear rows are deterministic. The two
continuous-time rows (ContiFormer, then Neural Hawkes) are simplified stand-ins rather than the published
models, and ran with an inter-event interval of identically zero
(\S\ref{app:baselines}).}
\label{tab:family-deltas}
\centering\small
\begin{tabular}{>{\raggedright\arraybackslash}p{0.36\linewidth}lc}
\toprule
\textbf{Family} & \textbf{Input} & $\Delta$(M16$-$M8) \\
\midrule
GRU (2-layer) & per-event numeric channels & $+0.076$ \\
LSTM (2-layer) & per-event numeric channels & $+0.077$ \\
Causal attention variant (time bias inactive) & per-event numeric channels & $+0.070$ \\
LSTM variant (decay inactive; no point-process likelihood) & per-event numeric channels & $+0.085$ \\
Gradient-boosted trees~\citep{ke2017lightgbm} & 6 aggregate features & $+0.084$
($+0.089$ matched) \\
Ridge & 6 aggregate features & $+0.099$ \\
Single-feature linear ($a\,x + b$) & input-window activity total & $+0.107$ \\
\bottomrule
\end{tabular}
\end{table}

A longer window narrows the spread between architectures. At M16 the transformer, GRU, LSTM,
ContiFormer, and Neural Hawkes all land within
${\sim}0.005$ of each other on the same objective with their clocks as built (seed 42, full test split). This band includes both continuous-time models, whose time handling was inert here.
A separate test that gives each a working clock detects no gain for ContiFormer and helps Neural Hawkes, by about twice the band ($+0.0116$, 95\% $t$ interval over three seeds $+0.0102$ to $+0.0130$; different runs from the band's, not bridged; scope in \S\ref{app:baselines}).
Even the largest family gap at short windows narrows. Neural Hawkes
starts $0.015$ \emph{below} the transformer and ContiFormer at M8 ($0.4619$ vs.\ ${\sim}0.477$; seed 42) and converges
by M16 ($0.5474$ vs.\ $0.548$--$0.553$).
At 24 weeks the two recurrent families, the
ContiFormer stand-in, and the ridge all land within
${\sim}0.008$ of the transformer's M24 score (GRU $0.5874$, LSTM $0.5857$, ContiFormer
$0.5810$, ridge $0.5796$ vs.\ transformer-Huber $0.5869$; seed 42). Even the ridge, which trailed the transformer by $0.016$ at M16, joins this group. These runs come from different container builds, so we treat them as within-family curves, not cross-family wins (\S\ref{sec:eval}).
In every family we trained, the window gain tracks the conditioning set rather than an
architectural capability, with train-side eligibility unresolved
(Table~\ref{tab:window-grid}).

\begin{table}[t]
\caption{Each 10M-tier transformer entry sits
within $0.006$ of its 1M counterpart, while the ridge \emph{level} dips below its 1M counterpart at every
window. Ridge is the exception among the two families in this table, which are also the two whose
training set grows with the tier (see text). Window ladders, log-space $R^2$ on the full test split (seed 42). The ridge M16-10M
and M24-10M values were recomputed separately (footnote in text).}
\label{tab:scale-10m}
\centering\small
\begin{tabular}{lcccc}
\toprule
\textbf{Model (10M tier)} & \textbf{M8} & \textbf{M12} & \textbf{M16} & \textbf{M24} \\
\midrule
Transformer (Huber) & $0.4757$ & $0.5199$ & $0.5510$ & $0.5860$ \\
Ridge (6-feature) & $0.4095$ & $0.4732$ & $0.5146$ & $0.5633$ \\
\bottomrule
\end{tabular}
\end{table}

In this design's operating regime, customer scale moves log-space $R^2$ only a little and the window moves it a lot. Across the complete
$\{$M8,M12,M16,M24$\}\times\{$1M,5M,10M$\}$ transformer grid, each 10M result
(Table~\ref{tab:scale-10m}) sits within $0.006$ of its 1M
counterpart.
At 3 seeds the M24 results match across scales
(1M $0.5885{\pm}0.0015$,
5M $0.5832{\pm}0.0039$, 10M $0.5885{\pm}0.0033$, mean~${\pm}$~population SD). The
per-seed ranges at 5M and 10M are wide ($0.0094$ and $0.0072$), so we state the M24 flatness
only to within that seed-noise width. The window gain
agrees at 10M and 1M in both forms: full-set
$\Delta$(M16$-$M8)
$+0.0753$ at 10M vs.\ $+0.0754$ at 1M (seed-42), matched $+0.0934{\pm}0.0005$ vs.\
$+0.0939{\pm}0.0022$ (3-seed). Going
from the 1M to the 10M tier moves the window gain by ${\leq}0.002$, so in this regime the scale-by-window interaction stays within seed noise. Each axis varies one factor at a time within the
fixed d128, 2-layer recipe, apart from the tier contrast's realized learning-rate schedule (\S\ref{sec:limitations}). Capacity is probed only at 1M
(\S\ref{sec:ceiling}). On corrected inputs, a separate check crosses the 8- and 16-week windows with model size at 1M:
the base model still gains from 16 weeks over 8, and no interaction with model size is shown (\S\ref{sec:repro}). The M12 10M result
also has 3 seeds
($0.5223{\pm}0.0020$; the cross-scale M12 spread at 3 seeds is ${\leq}0.0018$ against per-seed ranges up to $0.0125$), so
scale-flatness at that window does not rest on single seeds. Across the M8--M16 grid the GRU
and LSTM rows have a maximum within-window spread of
$0.004$ (18 runs), which we treat as fixed-exposure measurements, not scale
response (\S\ref{sec:ceiling}). Ridge is within $0.002$ from 1M to
5M but not to 10M.
The ridge 10M window column (Table~\ref{tab:scale-10m}) climbs
$+0.1050$ from M8 to M16, tracking the 1M ridge gain ($+0.0992$). The linear floor has the
\emph{steepest} window gain at 10M.\footnote{We recomputed the M16-10M and M24-10M ridge values outside the training pipeline, with code that reproduces the adjacent M12-10M entry's pipeline value end-to-end from raw data to $3.9\times10^{-9}$.} The linear-floor
level dip at 10M relative to 1M \emph{narrows} in $R^2$ with the window
($-0.028/{-}0.023/{-}0.022/{-}0.016$ at M8/M12/M16/M24).
The linear floor is also the one exception to the scale bound, though the comparison set is small: of the families
with results at all three tiers, it and the transformer are the only two whose training set grows with
the tier at all. What is similar across tiers for the linear family is its window \emph{gain}, not its level.
The sequence-model window gain is $+0.069$--$+0.085$ at every scale tested, with both endpoints
from seed-42 runs
(on the same full test split, the linear/GBM floor climbs $+0.084$--$+0.105$ across scales). The low end of the sequence-model range comes from the 5M Huber transformer's seed-42 run and the high end from the 1M Neural Hawkes stand-in (the LSTM-variant row of Table~\ref{tab:family-deltas}).
Window length, not the population tier, is the larger axis on this task.

\subsection{Architecture is second-order at matched objective}
\label{sec:lossnull}

Much of what first looks like an architecture effect on this task is the training objective. At fixed window and architecture, the objective reorders the leaderboard, with MSE~$>$
Huber~$>$ pinball at every window. Its ${\sim}0.03$--$0.05$ spread
is smaller than the window effect but decisive for cross-family comparisons.
``Objective'' here names the loss on the aggregate prediction: the sweep moves that term and
leaves the fixed-weight auxiliary per-event term unchanged in every configuration of the sweep
(\S\ref{app:objective}).
Under the Huber objective, the
transformer at M16
($0.5508{\pm}0.0016$) ties the GRU ($0.5522$, seed 42, within one seed-SD of the
transformer's 3-seed mean); under the MSE objective it beats
the same Huber-trained GRU by
$+0.035$. That margin does not survive loss equalization (Table~\ref{tab:losseq}).
When the recurrent baselines are retrained under the same MSE objective at their own
budget, at 3 seeds and every window through M16, all three
sequence architectures climb ${\sim}+0.08$ together from M8 to M16 (seed SDs ${\leq}0.0011$
throughout). We then equalize the training
\emph{budget} as well, retraining both recurrent families at 3 seeds on the transformer's own
budget ($8{,}192$-effective batch, 20 epochs, and the near-full training set in
place of the baselines' fixed cached draw). The retrained baselines cache $4.0$ of the $5.0$M rows once and reuse them every epoch, while the transformer sees every row over a run; this shortfall favors the transformer.
Every family$\times$window result lands within $\pm0.005$ of the transformer's
3-seed mean and slightly above the corresponding own-budget result (seed SDs ${\leq}0.0002$;
shifts from own to equalized budget of $+0.0008$ to $+0.0019$), and the sign flip of the
transformer$-$GRU gap at M16 is within seed noise. The spread across the three families is within $0.005$ at
every window through M16 and ${\sim}0.001$ at M16.

\begin{table}[t]
\caption{Architecture is second-order at matched objective: loss- and budget-equalized
retraining at 1M, MSE objective, log-space $R^2$ on the full test split (3-seed means;
columns are M8/M12/M16). \emph{Own budget} is each recurrent family's original
training budget (batch 128, single process, 10 epochs, fixed cached draw);
\emph{transformer budget} is the transformer's ($8{,}192$-effective batch, 20 epochs,
near-full stream). Seed SDs are ${\leq}0.0011$ (transformer; recurrent, own budget) and ${\leq}0.0002$
(recurrent, transformer budget).}
\label{tab:losseq}
\centering\small
\begin{tabular}{lccc}
\toprule
\textbf{Model (MSE objective)} & \textbf{M8} & \textbf{M12} & \textbf{M16} \\
\midrule
Transformer & $0.5075$ & $0.5557$ & $0.5875$ \\
GRU, own budget & $0.5038$ & $0.5533$ & $0.5863$ \\
LSTM, own budget & $0.5024$ & $0.5520$ & $0.5850$ \\
GRU, transformer budget & $0.5055$ & $0.5547$ & $0.5879$ \\
LSTM, transformer budget & $0.5033$ & $0.5537$ & $0.5869$ \\
\midrule
Transformer $-$ GRU (transformer budget) & $+0.0020$ & $+0.0010$ & $-0.0004$ \\
Transformer $-$ LSTM (transformer budget) & $+0.0042$ & $+0.0020$ & $+0.0006$ \\
\bottomrule
\end{tabular}
\end{table}

A
second matched operating point runs the equalization the other way, retraining the transformer
down to the baselines' batch size and epochs (batch 128, 10 epochs, both sides on the full train split; three seeds at
M16, single seed elsewhere).
At this budget the recurrent nets come out ahead at every window from M8 to M16, though the down-budgeted transformer converges with them at M8 and M12.
All recurrent models in this comparison were retrained at this same reduced budget, so they are not
the rows of Table~\ref{tab:losseq}. At M16 both recurrent nets land
$0.0060$ above the down-budgeted transformer, outside the convergence band in the baselines' favor (GRU $0.59019$, LSTM $0.59013$
vs.\ transformer $0.58417{\pm}0.0016$; 3-seed means on both sides; recurrent
seed SDs $\le 0.0002$). Every M16 transformer seed falls below both
recurrent means.
The two operating points differ in batch size and epochs, so we report them separately rather than pooling them.
At neither budget does any window from M8 to M16 show a transformer advantage
outside the convergence band. The $+0.035$ ``transformer beats GRU'' margin above
is therefore a loss confound, and we use the Huber-trained baseline results only in
objective-matched comparisons.
On this task the objective and the conditioning set carry the result. Architecture
is second-order to both, though not absent. \S\ref{sec:baselines} reports the residual
margin inside the band and how the checkpoint-selection rule bears on it.

\subsection{Most of the objective gap is monotone but not affine}
\label{sec:objcal}

A monotone remapping of each model's predictions closes most of the gap between training objectives. \S\ref{sec:lossnull} leaves the training objective as the largest factor that reorders the
leaderboard, and part of that spread is definitional (\S\ref{sec:eval}). On a fixed
cohort, log-space $R^2$ is an affine transform of log-space MSE, so only the squared-error model is
scored on the statistic its objective optimizes; the Huber and pinball models are scored on a statistic theirs does not optimize. If the objective gap were an affine displacement, the best affine correction
would close it. On both objective pairs at seed 42 (MSE$-$pinball and MSE$-$Huber; Table~\ref{tab:objcal}) it closes $28.6$--$30.7\%$, while a
\emph{monotone} map fit on validation closes $89.9$--$94.1\%$. Most of the gap is monotone but not affine, and
$61$--$63$ percentage points of the closure come from the non-linear part of the map.

\paragraph{The screen.} We ask how much of a gap between two models is placement on the scale.
A monotone map from prediction to prediction is the most flexible correction that a purely
ordering-based score cannot see. We fit one on \emph{validation only}, then re-measure on test.
Such a map cannot reorder customers, so whatever it recovers was placement rather than
knowledge the model lacked. Whatever survives the map is not placement that a monotone map fit
on validation can find. The map is isotonic regression
(increasing, out-of-range clipped) fit on the full validation split
($n_{\mathrm{val}}{=}239{,}886$) and applied to test ($n_{\mathrm{test}}{=}480{,}673$); test
rows never enter the fit, and each model gets its own map. We check that the map preserves
order: it moves a model's Spearman $\rho(\hat y, y)$ by
$5.3{\times}10^{-6}$--$3.4{\times}10^{-5}$, producing ties and never reorderings, and an exact
count finds zero inversions against the raw order for every model. The screen
runs on nine models (three objectives $\times$ seeds $\{42,123,456\}$) at M16 and 1M in one
recipe family. It is cheap, since it needs only the per-row predictions and no retraining, and it is
reusable wherever a claimed gap sits on a metric that mixes ordering with placement.
\paragraph{The screen is not new; the measurement is.} Ranking models by their average score under one scoring function is sound when that function is consistent for the functional being forecast \citep{fissler2022guide}. Isotonic recalibration of each model, the core of our screen, has precedents.
\citet{gneiting2025potential} recalibrate weather forecasts by applying isotonic
distributional regression to each model's deterministic output and comparing the resulting
``potential'' CRPS, which they note is invariant under strictly increasing transformations of
model output. Isotonic recalibration is likewise routine for response prediction in online advertising, where
\citet{deng2021calibrating} bin, apply isotonic regression, then interpolate.
\citet{wuthrich2023isotonic} apply it to a \emph{continuous} regression target
rather than a class probability, recalibrating an insurance pricing model so that
no price cohort systematically cross-finances another.
\citet{gneitingresin2023} decompose point-forecast scores, and \citet{arnold2023decompositions} decompose mean CRPS. We measure the fraction of the test-score gap between two models that closes after fitting a separate isotonic map to each model on validation data. \citet{kull2015decompositions} already treat the
two-term split of a proper score into calibration loss and refinement loss as well known.
They refine that split for classification into adjustment loss, post-adjustment calibration loss,
grouping loss and irreducible loss, with the separation proved for the Brier score and
log-loss. The refinement isolates what a correction weaker than full calibration can remove.
We use the split but none of the theorems. Our score is squared error on a continuous
target in log space. The share we report is what a fitted monotone map removes, and it
falls on the calibration side of the split they inherit rather than in one of the four terms they
add. The part the map cannot remove also has names in this finer decomposition. For a single model, the
surviving part divides in turn into whatever calibration loss the fitted map leaves, a grouping
loss (rows that share a predicted value but not a conditional mean), and a term irreducible
given the inputs. \citet{perezlebel2023grouping} give an estimator for the grouping loss of
classifier confidence scores. Our residual is a \emph{difference} between two models scored on
the same test rows, so the irreducible term is common to both models and cancels. What survives
our map is therefore a grouping-loss difference plus a
difference in the calibration loss each model's validation-fit map leaves on test. These names
locate our residual inside an established decomposition and supply no number for it. The standing guidance is to fix the
prediction target functional a priori, and only then choose a scoring function
consistent with it \citep{fissler2022guide}. That guidance applies to our comparison, which scores
the squared-error, Huber and pinball models on one log-space $R^2$. Log-space $R^2$ is consistent only
for the log-space mean, a functional the Huber and pinball losses do not elicit.
What this paper adds is the size of the map's effect on this axis. We apply the screen across
a nine-model \emph{objective menu} (three training losses $\times$ three seeds on one task and
recipe family) and report the fraction of the between-objective gap the screen removes, without
using it to rank forecast systems. That fraction is the finding, and it
is large enough to change which experiments are worth running.

\paragraph{The spread across the nine models shrinks to a tenth.} Before remapping, the nine models' log-space $R^2$ values span
$0.0514426$. After each model's own validation-fit monotone map, they span $0.0052436$, a
$9.81\times$ compression. On this metric, most of what looks like an accuracy gap between
training objectives is where each model sits on the scale.

\begin{table}[t]
\caption{The closure ladder: linear corrections close under a third of the objective gap, and a validation-fit monotone map closes most of it.
Same test rows for every model (all models are M16-1M, $n{=}480{,}673$), seed 42, log-space $R^2$.
\emph{Best offset} is the additive constant $c{=}\mathrm{mean}(y){-}\mathrm{mean}(\hat y)$;
\emph{best affine} is the maximum over all $a\hat y{+}b$, which in closed form is Pearson
$r^2(\hat y, y)$; \emph{isotonic} is the validation-fit monotone map. The two linear corrections
are fit \emph{on test}, so they are flattered and give an upper bound on what a linear
map can gain for each model. The isotonic correction is fit on validation.}
\label{tab:objcal}
\centering\small
\begin{tabular}{lcccc}
\toprule
\textbf{Correction} & \textbf{MSE$-$pinball gap} & \textbf{closed} & \textbf{MSE$-$Huber gap} & \textbf{closed} \\
\midrule
none (raw) & $0.0454020$ & --- & $0.0354815$ & --- \\
best offset (test-fit) & $0.0312111$ & $31.26\%$ & $0.0256794$ & $27.63\%$ \\
best affine (test-fit) & $0.0314731$ & $30.68\%$ & $0.0253425$ & $28.58\%$ \\
isotonic (validation-fit) & $0.0026592$ & $94.14\%$ & $0.0035827$ & $89.90\%$ \\
\bottomrule
\end{tabular}
\end{table}

\paragraph{The ladder.} Of the three nested correction classes in Table~\ref{tab:objcal}, only the monotone map closes most of the gap. An
intercept alone closes $27.6$--$31.3\%$ of the gap, and a slope adds essentially nothing. The second
linear parameter is worth $-0.58$ points on the pinball pair and $+0.95$ points on the Huber pair, negligible either
way next to the ${\sim}62$-point gain from the monotone map. The ladder therefore refutes, as an account of \emph{this} gap,
the naive reading that ``log-space $R^2$ is affine in log-space MSE, hence the objective gap is an affine
artifact.'' If the gap were an affine
displacement, the best affine correction would close it entirely. At seed 42 the best
affine map closes under a third of it. That affine relation between log-space $R^2$ and
log-space MSE ties two scores of one model on a fixed cohort; it does not make two models'
predictions affine rescalings of each other, which is the step the naive reading needs.
The linear corrections are fit on test while the monotone map is fit on validation; fitting all three
on test shifts the isotonic row by under half a point, in opposite directions
on the two objective pairs, so we treat the non-linear share
as a measured decomposition at seed 42 rather than as a bound.

\paragraph{Most of the gap closes at every seed.} Across all six
matched-seed comparisons, the monotone map closes
$88.90$--$96.76\%$ of the raw gap. The lowest share is MSE$-$pinball at seed 456 and the highest is
MSE$-$Huber at seed 123; the two three-seed-mean comparisons sit at $92.50\%$ (MSE$-$pinball) and $93.51\%$ (MSE$-$Huber).

\paragraph{The models' rank correlations with the target agree to within about $0.002$ in
$\rho$.} The nine models' raw Spearman $\rho(\hat y, y)$ spans $0.0011573$ while their raw
log-space $R^2$ spans $0.0514426$, so the metric spread is $44.45\times$ the rank spread
(clustered 95\% CI $[38.9, 51.3]$, so the ratio is resolved to within
about $15\%$). Even at its clustered upper bound, that rank spread stays under the registered $0.002$ tolerance. The tolerance was set on the nine-model spread of
$\rho(\hat y, y)$, so the agreement concerns each model's correlation with the target. It does
not show that the models' orderings agree, because a common correlation with a third variable
is not enough to establish a common ordering. We compare the orderings directly below,
using as reference the within-objective seed-pair floor, the lowest $\rho$ between the predictions of two seeds of one objective. We registered a falsifier for the calibration
interpretation before measuring it: pinball's raw $\rho$ had to stay at or above $0.855$. It is
$0.859601 / 0.859639 / 0.858680$ (seed 42/123/456), with a smallest margin of $+0.00368$, so the falsifier does not trigger. One
qualification cuts the other way: two objectives disagree on ordering more than two
seeds of one objective do, by a resolvable margin. At matched seeds, the $\rho$ between the pinball and MSE
predictions falls as low as $0.9959808$, below the within-objective seed-pair floor of $0.9979144$.
Paired on the same clustered resample, that difference is $\Delta\rho{=}{-}0.0019336$, with a
clustered 95\% CI of $[-0.0019824, -0.0018861]$ that excludes zero. We claim that gap and
nothing more. The gap $\Delta\rho$ is the excess disagreement over the within-objective seed pair and does not bound
how far apart the two orderings are. That distance is the shortfall of the matched-seed $\rho$ of
$0.9959808$ from $1$, about twice $\Delta\rho$. The cross-versus-within
comparison is measured at the two extremes, one cross pair against one within pair, so it is
established there and is not a uniform statement over every pair of models. As a row-level quantity, $\Delta\rho$ is
not comparable with a seed-level interval. A second qualification cuts the same way. Only one model has a
raw $\rho$ below the registered Huber/MSE band (their post-map $\rho$, $0.859252$--$0.859814$): pinball at seed 456,
$5.7{\times}10^{-4}$ below the band's lower edge. It also has the largest residual after the map
(reported below), so a small part of the objective gap is ordering and not only placement.

\begin{table}[t]
\caption{Each objective puts its point prediction at a different place on the scale. The
columns compare raw test predictions with the target's own mean and median in $\log(1{+}y)$
space, nine models printed as seed 42/123/456. The ordering is strict MSE $<$ Huber $<$ pinball
on \emph{both} columns for $9$ of $9$ models, and the three per-objective ranges do not overlap on
either column. Mean-bias ranges are MSE $[0.0894, 0.1185]$, Huber $[0.2692, 0.2880]$ and pinball
$[0.3184, 0.3648]$. Median-bias ranges are MSE $[-0.2081, -0.1782]$, Huber $[-0.0795, -0.0646]$
and pinball $[-0.0181, +0.0026]$. This table locates each model on the scale; it does not measure how well the model
orders customers.}
\label{tab:objbias}
\centering\small
\begin{tabular}{lcc}
\toprule
\textbf{Objective (seed 42 / 123 / 456)} & \textbf{mean}$(\hat y)-$\textbf{mean}$(y)$ & \textbf{median}$(\hat y)-$\textbf{median}$(y)$ \\
\midrule
MSE     & $+0.0894/{+}0.1185/{+}0.0897$ & $-0.1892/{-}0.1782/{-}0.2081$ \\
Huber   & $+0.2692/{+}0.2828/{+}0.2880$ & $-0.0795/{-}0.0695/{-}0.0646$ \\
pinball & $+0.3184/{+}0.3255/{+}0.3648$ & $-0.0181/{+}0.00004/{+}0.0026$ \\
\bottomrule
\end{tabular}
\end{table}

\paragraph{The mechanism: elicitability.} Elicitability explains why the models sit
at different places on the scale. A scoring function is consistent for the mean functional if
and only if it is a Bregman function \citep{gneiting2011making}. Each objective thus places its
point prediction at the functional \emph{it} elicits, and a metric built on a different
functional penalizes the other objectives for sitting elsewhere. On this target the two functionals
are far apart. In $\log(1{+}y)$ space the median sits
$+0.765462$ above the mean (test; $+0.760714$ on validation), and the global target is exactly
zero for $8.7\%$ of these rows; the target is left-skewed in that space. A median-targeting loss should therefore sit high against the mean and
a mean-targeting loss low against the median. Both predictions hold at every seed (Table~\ref{tab:objbias}). Pinball at
$\tau{=}0.5$ places its marginal prediction median within $0.0181$ of the target median on
all three seeds, while sitting $+0.32$ to $+0.36$ above the target mean. MSE is the reverse, nearest the mean ($+0.09$ to $+0.12$) and furthest below the median.
Huber, an M-functional that interpolates between median and mean as $\delta$ varies, sits strictly
between the other two on every seed. The losses target conditional means, medians or Huber functionals, while the ordering of the fitted models' marginal biases in Table~\ref{tab:objbias} is an empirical result. Both columns of that table are marginal level comparisons, one scalar per model; they
locate each objective's point prediction and do not assess conditional
calibration. A check on a second score comes out the way the mechanism predicts. Scored raw on the
shared seed-42 rows, the Huber model, which sits nearer the median, beats the MSE
model by $0.0258032$ on log-space MAE (the median's own score) while losing by $0.0354815$ on
log-space $R^2$. After each model's validation-fit map, the MAE ordering reverses and the MSE model's MAE
is lower by $0.0032747$. The map pulls every model toward the conditional mean before
that score is taken. Both halves rest on a single MSE seed and neither includes a pinball model, so this corroborates the mechanism but does not test it.
The mechanism is a hypothesis that survived a consistency check; it is not a
demonstration. The ordering check is limited to nine models in one recipe family, and the theory's
elicitability statement concerns population functionals and says nothing about held-out $R^2$ in
finite samples.

\paragraph{A small residual survives the map.} Its interval from resampling customers excludes zero,
and its interval across the three training seeds contains zero. On the shared seed-42 rows the post-map
MSE$-$pinball gap is
$+0.0026592$, with a customer-clustered paired bootstrap 95\% CI of
$[+0.0023468, +0.0029677]$ ($B{=}1000$, cluster $=$ customer, $383{,}208$ clusters over
$480{,}673$ rows) that excludes zero. The per-seed estimates cover all three seeds:
seed 123 is
$+0.0027866~[+0.0024883, +0.0030932]$, seed 456 is $+0.0052436~[+0.0048059, +0.0056768]$, and all three are sign-consistent and exclude
zero on log-space $R^2$. The same three seeds also exclude zero on log-space MAE and on a
$\tau{=}0.9$ pinball score, both of which also favor the MSE model. At seed level
the three matched deltas are $0.0026592 / 0.0027866 / 0.0052436$, mean $+0.0035632$, and
$t(2){=}4.2366$ against $t^\ast{=}4.3027$, so the seed-level 95\% CI is
$[-0.0000556, +0.0071819]$ and \emph{contains} zero. The MSE$-$Huber residual behaves the same way and fails the
same bar:
mean $+0.0023822$, $t(2){=}3.4941$, seed-level CI $[-0.0005512, +0.0053157]$, also containing
zero. We report the surviving MSE$-$pinball effect as $+0.0027$ (seed 42) to $+0.0036$ (three-seed mean), seed-unresolved.

\paragraph{What this does not license.} \emph{(i) It is not
a ranking claim in either direction, and specifically not ``MSE ranks customers better.''}
After the map, Spearman $\rho$ favors Huber by a mean of
$+1.2{\times}10^{-4}$, though the two objectives' per-seed values overlap. All three Huber models beat all three MSE models on
the top target decile ($n{=}48{,}068$). $R^2$ after the best monotone map is an ordering
functional, but a squared-error-weighted one. Our map is fit on validation and closes within half a
point of the one fit on test (Table~\ref{tab:objcal}). The residual therefore says MSE's ordering supports a
monotone conditional-mean predictor better \emph{in the sense this metric scores}, and nothing
more. \emph{(ii) One recipe family.} M16, 1M, the primary dataset, no target winsorization,
$\log(1{+}y)$ space. We claim no transfer across scale, capacity or objective family.

\paragraph{The window gain survives the same screen.} We recalibrate the window models too, to ask whether part of the window gain is placement.
We fit each model's own monotone map on \emph{validation} and re-score on test, the same
procedure used on the objective axis. The map closes $-0.73\%$ to $6.65\%$ of the 8-to-16-week
gain for the Huber models across the three seeds (mean $3.37\%$). Hence $93\%$ to $101\%$ of the window gain survives a monotone
remapping, while on the objective axis the map removes $88.9\%$ to $96.8\%$ of the gap. With the same procedure,
task and recipe apart from the window, the closed fractions differ by more than an order of magnitude.
Unlike the objective pair, the two window models are recalibrated on \emph{different} validation
cohorts. The closure therefore depends on how the two fits are aligned, a choice the seed range
does not cover. Refitting each model's map on only the validation rows both cohorts share moves the
three-seed mean closure to $1.8\%$. Recentering each remapped model on the matched cohort moves that
mean to $7.8\%$. The contrast with the objective axis survives that whole span. Neither alternative alignment
carries an interval, so we treat the closure as a direction rather than a point estimate. For the
placement share we rely instead on a one-sided ceiling over three window steps ($8{\to}16$, $8{\to}12$ and $12{\to}16$ weeks),
which assumes nothing about how the fits are aligned and agrees with this direction.
We give the short-window model the best monotone map that exists \emph{on the very rows it is
scored on} and leave the longer-window model raw. The longer window still wins in all nine
step-and-seed comparisons, by $+0.0093$ to $+0.0667$. Even under that handicap, at most
${\sim}30$--$32\%$ of the 8-to-16-week gain could be placement. On the 8-to-12-week step,
however, the ceiling reaches $44$--$58\%$, and across all nine comparisons it runs as high as $67\%$.
With the same test-fit ceiling at seed 42, the objective axis could be ${\geq}97.6\%$ placement, and
on one objective pair the remapped loser \emph{overtakes} the raw winner outright. That asymmetry,
not either number alone, is what the screen establishes. The screen rules out placement as the
explanation of the window gain. It does \emph{not} rule out seasonality: this screen leaves the calendar
collinearity of \S\ref{sec:limitations} untouched, and the calendar check of \S\ref{app:calendar} separates it in part, on the weekly-aggregate ridge probe.

\paragraph{What else this does not license.} \emph{(iii) Deltas only.} We cite gaps and their closures,
never a per-model post-map level. The one deliberate exception is the one-sided ceiling above,
whose purpose is to show that a model's post-map level still loses to another model's raw level. \emph{(iv) A diagnosis, not a correction we propose applying.} \emph{(v) The bias
table locates each model and does not rank them.} Limits (i) to (v) leave one claim. On this recipe
family most of the objective axis's apparent accuracy spread is placement on the scale. That
placement is monotone but, at seed 42, not affine, and a small residual survives that we can
resolve over customers but not over seeds.

\begin{figure}[tb]
  \centering
  \includegraphics[width=0.72\linewidth]{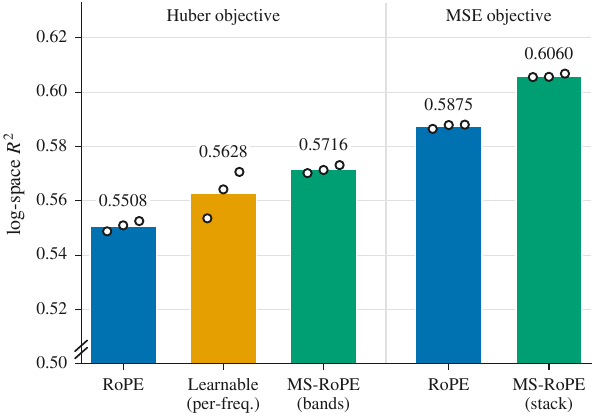}
  \caption{%
    \textbf{In log space, structured physical-time positional encoding beats RoPE as implemented and stacks
    almost additively with the MSE objective.}
    M16-1M configurations, log-space $R^2$ on the full test split (not the matched cohort of
    Table~\ref{tab:window-grid}); bars are 3-seed means and dots are individual seeds
    $\{42,123,456\}$, including the stacked configuration ($0.6060{\pm}0.0006$).
    Under the Huber objective, MS-RoPE's semantic frequency bands beat
    RoPE ($+0.021$) and sit above the unconstrained learnable control (\S\ref{sec:time}) on
    means ($+0.009$, inside the control's seed-to-seed variation).
    Stacked on the MSE objective, the two gains are nearly orthogonal
    ($0.6060$ observed vs.\ $0.6083$ fully-additive, $96\%$ of the additive gain, 3-seed),
    giving the best log-space M16 result on the primary dataset. The same stack at M24 reaches
    $0.6419{\pm}0.0024$ (3-seed).
    The $y$-axis starts at $0.50$ (break marked); bar labels carry the values.
  }
  \label{fig:pe-stack}
\end{figure}

\subsection{Time representation gives the largest model-side gain over the rotary encoding as implemented}
\label{sec:time}

Every model-side factor above, other than how a model represents time, moved $R^2$ by at most $0.007$ in magnitude, on 3-seed means where we have them and seed 42 otherwise
(apart from two single-seed configurations: content at 12 weeks, \S\ref{sec:ceiling}, and the decay-gated recurrent variant at 8 weeks, \S\ref{sec:universal}).
A dedicated M16 ablation finds the same flatness, with one
single-seed configuration just outside that range
(constant-coded category and brand content $-0.004$, linear anchor of \S\ref{app:objective} off $-0.003$, learned absolute positions in place of RoPE
$-0.0073$, a dual mean$+$median head $-0.000$; seed 42), none a gain. Like the category configurations of \S\ref{sec:ceiling}, the constant-coded category and brand variant places no bound on the value of content. Measured against the rotary encoding as implemented, the largest model-side improvement comes from how the
model represents time itself. Better time representation is a level lever, not a mechanism by which the model
exploits a longer window. Holding everything else fixed (M16, 1M, Huber, same code build, single
factor varied), we compare integer-position rotary encoding against two physical-time variants.
Because the base model receives no other elapsed-time input (\S\ref{sec:history}), physical-time
encoding is the only route by which the model is given elapsed time anywhere in the paper, so the contrast
measures what that encoding is worth here rather than the value of a second source of elapsed time.
The unconstrained control (``Learnable (per-freq.)'' in Figure~\ref{fig:pe-stack}) learns one linear scale per rotary frequency, a raw form of the log-space per-frequency scalars of \citet{karypis2026lerope}, in the spirit of the
rescaling literature but trained with the network rather than
derived~\citep{chen2023extending,peng2023yarn} or searched~\citep{ding2024longrope}.
The
structured variant (MS-RoPE) ties those per-frequency scales to $K{=}4$ shared semantic bands
over physical $\Delta t$ (hour, day, week and year) and superposes all four on every frequency pair with learned softmax mixing
(Table~\ref{tab:pe}; specified as equations in \S\ref{app:pe}).

\begin{table}[t]
\caption{Structured physical-time bands beat the integer-position baseline as implemented and sit above the
unconstrained learnable control on 3-seed means. The margin over the control sits within the control's own
seed spread, so that ordering is not seed-resolvable.
Time-representation sweep at M16, 1M, Huber objective, log-space $R^2$ on the full test split, with 3 seeds
per variant on the same code build. Positional encoding is the only factor varied.
All $\pm$ are population SD, and the
$\Delta$ column is the mean
of paired per-seed deltas $\pm$ population SD.}
\label{tab:pe}
\centering\small
\begin{tabular}{lcc}
\toprule
\textbf{Time encoding} & \textbf{$R^2$ (3-seed)} & $\Delta$ vs.\ RoPE \\
\midrule
RoPE (integer positions) & $0.5508{\pm}0.0016$ & --- \\
Learnable per-frequency (unconstrained) & $0.5628{\pm}0.0070$ & $+0.0120{\pm}0.0079$ \\
\textbf{MS-RoPE (4 physical-time bands)} & $\mathbf{0.5716{\pm}0.0012}$ & $+0.0208{\pm}0.0014$ \\
\bottomrule
\end{tabular}
\end{table}

The nearest non-rotary multi-scale prior work is cycle-aware time encoding for irregular
clinical events~\citep{xiao2024xtsformer}; rotary-over-$\Delta t$ work is nearer still
(\S\ref{sec:related}). MS-RoPE differs in applying the rotary encoding over
physical $\Delta t$, with bands \emph{initialized} at hour/day/week/year timescales
rather than learned cycle features.

All three variants share the rotary implementation of \S\ref{app:pe}. MS-RoPE's total gain over integer RoPE is seed-resolved:
$+0.0208{\pm}0.0014$ (Table~\ref{tab:pe}), positive at all three paired seeds,
with $t(2){=}21.3$ against $t^\ast{=}4.3027$ and a seed-level 95\% CI of $[+0.0166, +0.0250]$.
The gain comes in two steps, and neither is seed-resolved on its own. The first bundles two mechanisms inside the one
positional-encoding factor: physical $\Delta t$ replaces integer positions, and the
per-frequency scales become learnable. Together they add $+0.012$ on 3-seed means, positive
at all three seeds but seed-unresolved at $n{=}3$ ($t(2){=}2.16$ against
$t^\ast{=}4.3027$). No variant in this table changes one without the other, so this step does not
isolate either. The second step, to MS-RoPE's mixed semantic bands, adds a further
$+0.009$ over the unconstrained control on 3-seed means. That margin sits within the
control's own seed spread, so the ordering over the control is not seed-resolvable at
$n{=}3$.
The softmax mixing
concentrates on the day and week bands, which take $0.37$ and $0.38$ of the mass (hour $0.06$, year
$0.19$), with cross-seed mixing SD $\leq 0.021$ per band. Across all three seeds, each band's learned timescale stays within a factor of two of its physical
prior, contracting modestly and sign-consistently.

The time-representation and objective gains compose. A single stacked configuration (MSE
objective $+$ MS-RoPE, M16, 1M) reaches
$R^2 = 0.6060 \pm 0.0006$ (3-seed), against $0.5875$ for MSE alone and $0.5716$ for
MS-RoPE alone: $96\%$ of the additive gain, with the two nearly orthogonal
(Figure~\ref{fig:pe-stack}). The encoding gain
persists across the whole window ladder
($+0.015$--$0.022$ over plain RoPE, smallest at M8; seed 42, full test split), and
the window gain itself holds under it ($+0.116$ M8$\to$M24 under MS-RoPE vs.\
$+0.110$ under RoPE, seed 42 for both, full test split).
The encoding adapts its resolution to the window. As the window grows M8$\to$M24, softmax mass
moves from the day band to the week band (day $0.50\to0.26$, week $0.30\to0.51$, strictly
monotone across the seed-42 Huber ladder). On the same ladder, the slowest band's
learned timescale tracks the observable range of $\Delta t$ (time since the first retained event), from $134$ days at M8 to $241$ days at
M24, against a $1$-year prior. That
adaptation is large in the weights but contributes little to the window gain in $R^2$.
The window gains under the two encodings differ by
${\sim}{+}0.006$ of $+0.116$ (seed 42, full test split). The encoding gain and this
window-gain difference are measured in log space and are recipe-specific: under a raw-target recipe
and statistic the gain reverses sign and the window gain is attenuated under MS-RoPE instead
(\S\ref{sec:external}). The stacked configuration gives
the best log-space M16 result on the primary dataset (the same stack at M24 reaches
$0.6419{\pm}0.0024$, 3-seed). Its matched-cohort value is
$0.6052{\pm}0.0007$ (3-seed), below its full-split value at every seed. Its paired matched-cohort gain over
the plain-RoPE Huber model (Table~\ref{tab:window-grid}, matched row) is $+0.0336{\pm}0.0017$, with every seed's clustered CI excluding zero.

\subsection{Objective-matched baseline comparison}
\label{sec:baselines}

With objective, cohort, and denominator held fixed at once, the event-sequence transformer beats the
stronger of our two tabular baselines, and adding MS-RoPE widens the gap
(Table~\ref{tab:objective-matched}). This is our strictest cross-family comparison
(squared-error family, matched cohort, M16, 1M). The stronger baseline is a fixed-hyperparameter
gradient-boosted tree ensemble (400 trees, 63 leaves) on six aggregate
features of the same event fields as the ridge.

\begin{table}[t]
\caption{The transformer beats both of our tabular baselines at matched objective, and adding MS-RoPE widens its lead. M16, 1M, matched cohort ($n{=}407{,}338$),
log-space $R^2$. Every row's aggregate loss is squared error on the log target (transformer rows add the auxiliary per-event term). Sequence models see
raw event streams; GBM and ridge see six aggregate features.
GBM (3-seed SD ${\leq}1.4{\times}10^{-4}$) is fit and scored by separate code that reproduces the full-split ridge results of Table~\ref{tab:matched-cohort} to within ${\sim}10^{-5}$. $^\ddagger$matched-cohort value scored from the per-row
predictions, 3-seed mean $\pm$ population SD (full-test-split $0.6060{\pm}0.0006$,
3-seed); the Transformer (MSE) row is likewise 3-seed; the GBM row is seed 42 (its 3-seed mean matches at display precision); the ridge row is
deterministic. The stacked configuration is the only transformer configuration whose
matched $R^2$ sits marginally below its full-split value at every seed, so the full-split value is not a conservative proxy for it,
and we score it on the matched cohort directly.}
\label{tab:objective-matched}
\centering\small
\begin{tabular}{llc}
\toprule
\textbf{Model} & \textbf{Input} & \textbf{Matched $R^2$\,$\uparrow$} \\
\midrule
\textbf{Transformer (MSE $+$ MS-RoPE)} & raw events & $\mathbf{0.6052{\pm}0.0007}^\ddagger$ \\
Transformer (MSE) & raw events & $0.5945{\pm}0.0004$ \\
GBM & 6 aggregate features & $0.5867$ \\
Ridge & 6 aggregate features & $0.5412$ \\
\bottomrule
\end{tabular}
\end{table}

At matched objective the plain-RoPE transformer's margin over GBM (before any
time-representation gain) is
$+0.008$ and shrinks as the window grows
(the tabular baselines ride the same window lever).
That margin does not establish architectural superiority at a fixed window.
Time representation widens the margin considerably. On
the full test split the stacked configuration beats GBM by
$+0.0250$, against $+0.0065$ for MSE alone, a roughly $4\times$ widening that comes from
representation rather than capacity. On the matched cohort, which this
paper otherwise prefers, the margin grows from $+0.0078$ to $+0.0185$
(3-seed transformer values), a conservative like-for-like multiplier of
${\sim}2.3\times$
(Table~\ref{tab:objective-matched}).
The GBM comparator is fixed-hyperparameter and untuned (Table~\ref{tab:recipe}). A
tuned GBM could shrink this stacked margin. As long as it still trails the plain-RoPE transformer, though, it would raise the multiplier.
The ${\sim}2.3\times$ multiplier thus applies to \emph{this} baseline and is not a bound in either direction over tuned tabular models.
GRU/LSTM at matched objective and the transformer's budget sit within
${\sim}0.001$ of the plain-RoPE transformer at M16, within $0.005$ at every
window through M16 (\S\ref{sec:lossnull}). Their
own-budget runs under the Huber objective also sit within $0.005$ at M16: GRU $0.5729$, LSTM
$0.5715$, vs.\ $0.5731$ for the Huber transformer on the identical cohort and
denominator (seed 42 throughout, so the transformer value is the
seed-42 run rather than the 3-seed matched mean in Table~\ref{tab:window-grid}).
Recurrent results this close at matched objective and cohort do not settle the architecture question.
Under the MSE objective on the matched cohort, with the recurrent models at their own budget, the cross-family comparison has 3 seeds
and resolved CIs: the transformer sits $+0.0036/{+}0.0026/{+}0.0024$ above GRU and
$+0.0050/{+}0.0041/{+}0.0039$ above LSTM at M8/M12/M16 (customer-clustered paired
bootstrap CIs all excluding zero, sign-consistent at all three seeds). The seed-paired $t(2)$ interval at
M8 transformer$-$GRU grazes zero, $[-0.0001, +0.0073]$, so that comparison is unresolved at seed level
even though its row-level interval excludes zero. Five of the six margins are seed-resolved, but all are small. They sit within the $\pm0.005$
convergence band, with M8 transformer$-$LSTM at its edge ($+0.0050$, clustered-CI upper bound
$+0.0055$). The two sides ran on different training builds, which the comparison rule of \S\ref{sec:eval} asks them to share. The reference run of the bit-exactness check (\S\ref{sec:repro}) was trained on the transformer side's build, and the recurrent side's build reproduces it bit for bit, but that run uses the Huber objective, so the check does not establish equivalence for these MSE runs.
These margins, and the cross-family band they sit
inside, inherit the checkpoint-selection asymmetry of \S\ref{sec:limitations}. We chose the
transformer's epoch on validation error but not the recurrent models' epoch, and
matching selection across the families could move the margins and the band in either direction. The \emph{metric-space} channel of that same section is separate
and touches only the transformer, whose epoch was chosen on original-scale
validation error, while we report in log space. On the one configuration we re-evaluated
end to end (Huber at M16, 1M, seed 42), correcting that choice raised the transformer's
log-space $R^2$ by $+0.0022$ on the full test split, which would enlarge these margins.
That is one configuration at a single seed, not the MSE matched-cohort comparison above, so we treat it as a
direction rather than a bound.

Architecture effects on this axis are small but not zero. The
five sequence families lie within $0.0052$ of one another at a matched (Huber) objective at 16 weeks, at seed 42, on the full test split, with the clocks as built.
With 3 seeds under the MSE objective on the matched cohort, the recurrent models at their own budget and the two sides on different training builds, the
transformer is ahead of both recurrent baselines at every window through M16 by a small,
sign-consistent margin. Retraining the recurrent nets at the transformer's budget
shrinks that margin at every window and closes it at M16 (\S\ref{sec:lossnull}).
Against a window lever an order of magnitude bigger, these effects are second-order.

\subsection{External validity on a second population}
\label{sec:external}

Everything above uses one proprietary population. We replicate the window intervention on a public one,
the Alibaba Taobao UserBehavior dataset~\citep{zhu2018learning}, and $R^2$ rises with window length there too. The dataset spans 9 days,
so windows are measured in days rather than weeks (for months-long windows on four public datasets that span more than a year, see \S\ref{app:public-ladder}), and its target
is ${\sim}69\%$ zero, a much sparser and harder regime. The 3/5/7-day
extracts (D3, D5, D7) are input windows end-anchored to a common date, so all three predict
the same target period while observed history and the eligible training set vary. We fit ridge regression on
the $\log(1{+}y)$ target and score it on a matched cohort of
$6{,}985$ users present in all three extracts' test splits.
Its matched log-space $R^2$ rises monotonically, $0.0213 \to 0.0305 \to 0.0374$, and all three paired
window-delta bootstrap CIs exclude zero
($\Delta$(D7$-$D3) $= +0.0162~[+0.0090, +0.0229]$).\footnote{A start-anchored variant of
these extracts, in which the 2-day target window shifts with the extract, gives a
confounded estimate ($\Delta$(D7$-$D3) $= +0.0154$) that matches the
fixed-target one.
} A day-resolution
ridge ladder on the same data favors the functional \emph{form} of \S\ref{sec:saturation}, though not its bend location. The
saturating form beats log-linear, with $4.5\times$ lower fit RMSE on the full test split and $2.9\times$ lower on the matched fit over the $6{,}822$ users in all three start-anchored extracts' test splits,
where the fitted bend moves later. In the full-split fit, the bend sits at or beyond the edge of the observable
range, so a 7-day ladder fixes the form but not the half-saturation point. There is no universal saturation point: the bend already moves with activity within one population, as shown in
\S\ref{sec:allocation}.

\emph{Extract specification.}
The windows are end-anchored to 2017-12-02 00:00 China Standard Time (D3 $=$ [11-29,
12-02), D5 $=$ [11-27, 12-02), D7 $=$ [11-25, 12-02)), every extract predicting the
same two-day target: total purchases over [12-02, 12-04). A user enters an extract
with at least three events in its input window; we exclude a sparse tail of noisy events
after 12-04.
The ridge model uses four standardized aggregate features (log-scaled input-window activity
total, event count, purchase count, purchase rate) on the $\log(1{+}y)$ target,
regularization cross-validated over 25 log-spaced values in $[10^{-3},
10^{3}]$. The transformer runs reuse the primary recipe unchanged
(Table~\ref{tab:recipe}) with an MSE objective on the log
target (Huber in a robustness check), trained at an effective batch of $256$; the dataset schema differs, so the per-event input
mapping is given in \S\ref{app:taobao-inputs}.

The transformer replicates the direction under an internally registered design
on the end-anchored extracts (three seeds per window, matched cohort $n{=}6{,}985$):
$\Delta$(D7$-$D3) $= +0.047~[+0.035, +0.059]$ (customer-clustered paired bootstrap, trained models held fixed).
Table~\ref{tab:taobao-cells} lists the results behind these contrasts: ridge and per-seed
MSE levels by window, and the contrasts under both objectives.
Every seed is positive ($+0.062/{+}0.019/{+}0.059$), each with its own
clustered CI excluding zero. Excluding one run that
collapsed to the constant-zero predictor from its first epoch, the seed-mean window response is
monotone D3$<$D5$<$D7. The D7$-$D3 contrast
involves no D5 run and is identical with or without the exclusion. The seed spread
(SD $0.024$) came
in at over $2\times$ the largest spread the design was powered for, so the test was underpowered at the internally registered,
ridge-derived effect ($+0.0154$). It succeeds only because the observed effect is ${\sim}3\times$ larger (one-sided $t(2)$ test, $p{=}0.04$, across the three seed deltas).
The clustered CIs do not resolve the seed question (\S\ref{sec:limitations}), so the magnitude stays
seed-unresolved. Absolute levels also stay near zero (D3 seed-mean $-0.007$). The result is a window-direction
replication on a difficult ${\sim}2$-day-horizon, $69\%$-zero target and is not evidence of useful absolute
accuracy. As extracted, longer end-anchored windows also bring a larger eligible training set
($610$k${\to}686$k rows), so the contrast measures ``longer windows as extracted'' rather than a pure history effect.
The Taobao evidence covers direction in
both the linear and deep families, and shape in ridge: longer windows, as extracted, help on an independent
population with a different native timescale.

The deep-family direction survives a training-objective swap.
Retraining the same configurations under Huber (one factor changed, with the same
code build, data, extracts and seeds) replicates the contrast on runs that did not collapse:
$\Delta$(D7$-$D3) $= +0.045~[+0.030, +0.060]$ (customer-clustered paired bootstrap),
excluding zero.
That paired contrast rests on a single seed pair in which neither run collapsed. On the
broader set of non-collapsed runs, including replacement seeds, the unpaired contrast of window means
attenuates to $+0.022~[+0.011, +0.033]$, with the direction intact and the clustered CI still excluding zero. However, 4 of 9 Huber runs collapsed to the
constant-zero predictor (vs.\ 1 of 9 under MSE), most often at the shortest window (3 of 5
D3 seeds, including replacements). The Huber results therefore rest on fewer non-collapsed seeds than
the MSE results. Huber levels
are never compared to MSE levels here, and no level is compared across
datasets.

\begin{table}[t]
\caption{Taobao window results on the end-anchored extracts: matched cohort ($n{=}6{,}985$
users in all three extracts' test splits), log-space $R^2$ on the $\log(1{+}y)$ target. Every
window predicts the same two-day target, a purchase count that is $69\%$ zero, so absolute
levels sit near zero (a negative value is below the cohort-mean predictor) and only the
direction across windows is interpreted. Ridge is one deterministic fit per window; each transformer
entry is one training seed. $\Delta$(D7$-$D3) is paired by user, with 95\% intervals from a
user-level paired bootstrap ($B{=}1000$) with the fitted models held fixed. $^\dagger$Training collapsed to the constant-zero predictor from
the first epoch: not a level, and left out of the D5 mean, which is over two seeds.
$^\ddagger$Including replacement seeds for the collapsed runs: 5 of 13 with replacements; 4 of the first nine, as
in the text. The MSE rows are the nine
internally registered runs. The Huber results are used only through their contrasts: the paired one
uses seed 456, the only seed whose D3 and D7 runs both trained without collapse; the unpaired one compares means over non-collapsed
seeds (three seeds at D7, two at D3). Huber levels are not compared with MSE levels, and no
level here is compared with the primary dataset.}
\label{tab:taobao-cells}
\centering\small
\begin{tabular}{lcccc}
\toprule
\textbf{Model} & \textbf{D3} & \textbf{D5} & \textbf{D7} & \textbf{$\Delta$(D7$-$D3) [95\% CI]} \\
\midrule
Ridge (4 aggregate features) & $0.0213$ & $0.0305$ & $0.0374$ & $+0.0162~[+0.0090, +0.0229]$ \\
\midrule
Transformer (MSE) & & & & \\
\quad seed 42  & $-0.0299$ & $0.0323$ & $0.0323$ & $+0.062$ \\
\quad seed 123 & $0.0196$ & collapsed$^\dagger$ & $0.0382$ & $+0.019$ \\
\quad seed 456 & $-0.0116$ & $0.0341$ & $0.0472$ & $+0.059$ \\
\quad seed mean & $-0.0073$ & $0.0332$ & $0.0392$ & $+0.047~[+0.035, +0.059]$ \\
\midrule
Transformer (Huber) & & & & \\
\quad collapsed / trained$^\ddagger$ & 3/5 & 1/4 & 1/4 & \\
\quad paired, seed 456 & & & & $+0.045~[+0.030, +0.060]$ \\
\quad unpaired, non-collapsed means & & & & $+0.022~[+0.011, +0.033]$ \\
\bottomrule
\end{tabular}
\end{table}

Back on the primary population, the window gain also reappears under a change of recipe and metric. A second configuration for
this task was specified independently of this study, with its own objective, target
winsorization, and raw-target statistic. We change \emph{only} its observation window,
from 8 to 16 weeks. Its statistic, which we write \emph{clipped-raw} $R^2$ throughout this
section, is $R^2$ on the raw, un-logged target under that recipe's own winsorization at the
$p_{0.5}$ and $p_{99.5}$ percentiles. The window change moves its
clipped-raw $R^2$ from
$0.648$ to $0.718$ ($+0.070$) at the 1M tier, and the same gain replicates at 10M:
$0.655$ to $0.725$ ($+0.070$), the 8-week level holding at
${\approx}0.65$ at both scales. This statistic is defined in a
different space from log-space $R^2$ and is not comparable in magnitude to any other number in this paper. These are seed-42 point
estimates.
The window change alone moves the full-set statistic by $+0.070$ at both
scales (this recipe has no matched-cohort evaluation, \S\ref{sec:matched-method}), a choice about the extracted data that leaves the model unchanged.

The MS-RoPE gain, by contrast, is recipe-specific. Retrained under that raw-target recipe
(raw-target MSE, winsorized as above) and scored in that same clipped-raw
statistic, MS-RoPE \emph{underperforms} RoPE by $0.054$ (M8) and $0.075$ (M16) in that
statistic (two-seed means), opposite in sign to its $+0.021$ log-space gain under the Huber objective at M16. Per-event \emph{test}
loss
is essentially unchanged ($1296.6$ MS-RoPE vs.\ $1322.1$ RoPE at M8; $1271.9$ vs.\
$1268.1$ at M16; seed 42). The deficit
comes with lower aggregate dispersion, measured as \emph{dispersion ratio}
(prediction SD over target SD; unclipped predictions, winsorized targets):
MS-RoPE's predictions are underdispersed relative to RoPE's at both windows ($0.69$
vs.\ $0.77$ at M8; $0.70$ vs.\ $0.86$ at M16; seed 42), consistent with an
objective--calibration interaction. The window
lever transfers under both encodings at both seeds ($+0.070/{+}0.071$ RoPE vs.\
$+0.052/{+}0.047$ MS-RoPE, M8$\to$M16, in the clipped-raw statistic), attenuated under
MS-RoPE in each. At a second
seed the reversal keeps its sign and ordering; across both seeds it runs $39$--$52\times$ the
RoPE runs' seed spread.
The reversal makes the target-space
discipline concrete: an intervention's sign can depend on the recipe and target space it is
trained and scored in.

Extending the observation window moves log-space $R^2$ on this
task more than every architectural, scale, and capacity intervention we measured,
combined. The window lever's limits are measurements from internally registered tests on the test calendar, not
extrapolations. In log space and against the rotary encoding as implemented, physical-time positional encoding produces the largest model-side improvement among the interventions measured here, and the window gain
holds under it.

\section{Related work}
\label{sec:related}

\paragraph{Matched evaluation and matching estimators.}
We borrow the diagnosis of the matching literature, not its estimator. Matching in
causal inference, whether by propensity scores
\citep{rosenbaum1983propensity}, exact matching or their descendants
\citep{stuart2010matching}, builds treated and control groups with comparable covariates
so that an effect estimate is not a composition artifact. Our ``treatment'' is a longer
input window, a data-extraction choice that changes test-set eligibility, and the object we protect is an
evaluation rather than a treatment effect. We intersect test sets exactly on
(customer, target week) keys. On the resulting three-window cohort, targets are numerically identical across windows,
and the denominator is constant by construction
(\S\ref{sec:matched-method}). Covariate-shift and dataset-shift work
\citep{shimodaira2000covariate,quinonero2008dataset} corrects train/test mismatch,
typically by reweighting training. Our mismatch is between the windows being compared,
and fixing the test population corrects the bias in whichever direction it runs.
Composition masks the 8-to-16-week gain and inflates the 24-to-32-week one
(\S\ref{sec:matched}, \S\ref{sec:prereg}).
Concurrent and prior work documents window protocols that reverse comparisons. On subscription churn data, a
window-sufficiency curve inverts when cohort and target move
with the window but rises when both are held fixed~\citep{han2026how}.
Untuned look-back windows can invert architecture rankings on standard forecasting
benchmarks~\citep{abdelmalak2025channel}. Our matched-cohort protocol
(\S\ref{sec:matched-method}) scores every compared window's model on one fixed test set. The same principle,
measuring fidelity along the time axis rather than on pooled records, changes the ranking of
generators in the evaluation of synthetic sequential tabular data~\citep{kwon2026seq2synth}.

\paragraph{Context length.}
In language modeling, work on long contexts asks how to make a trained model accept longer
inputs and how well models use the length they nominally have; accuracy drops when the
relevant content sits in the middle of the context \citep{liu2024lost}. Positional interpolation
\citep{chen2023extending} and frequency rescaling
\citep{peng2023yarn,ding2024longrope} supply the machinery. We repurpose it twice:
as the unconstrained control of \S\ref{sec:time} and, constrained to physical-time
bands, as the model-side intervention that, against our rotary encoding as implemented, has the largest measured effect on log-space $R^2$. Our question is
whether the extra input is worth conditioning on at all. \citet{liu2024lost} pose the same question for retrieval-augmented question answering. We ask it on a task
where a longer window changes both what information the model sees and who is eligible. Customer-value modeling has asked the history-length question before.
\citet{ballings2012churn} varied customer history from one to sixteen years and found diminishing gains
in churn prediction for each of the three classifiers they tested (logistic regression, classification trees and
bagged classification trees). Thirty years ago, another study priced the worth of a purchase history
in dollars against deliberately shorter information sets \citep{rossi1996purchase}. A
short history returned two and a half times the gain from blanket couponing, and a single
observed purchase occasion one and a half times that gain. That study measures worth as
the monetary value of targeted couponing under a random-coefficient choice model and counts history
in discrete purchase occasions rather than calendar window length. Recent work makes the temporal framing of a churn model its
primary object, scoring risk repeatedly from rolling behavioral windows with the
observation and outcome periods held disjoint \citep{mufti2026rolling}. That framework's
windows are fixed where ours are swept: one 30-day observation window and one
30-day churn-evaluation window, scored by accuracy and ROC-AUC on a binary label rather
than by log-space $R^2$ on a forward real-valued total.
\citet{wang2019deepltv} predict future spending from information available at a
customer's first purchase. \citet{li2022billion} jointly predict customer value over
several future horizons. Our comparison varies the length of the input history.
Recurrent networks have been applied directly to customer-base analysis,
with calibration length varied as a sensitivity check on one dataset and the holdout
lengthened to match \citep{valendin2022rnn}. Buy-'til-you-die models condition on each customer's
own variable-length observation window \citep{fader2005counting}, as recent work still does
\citep{ghosh2026btyd}. These models target a
transaction count and an activity probability, whereas we target a forward real-valued total. Neither they
nor a variational-autoencoder buy-'til-you-die model that forecasts forward revenue \citep{naf2026clvae} varies the input window as a designed factor against a matched cohort. \citet{han2026how} run that control. We compare eight model families on a common 364-day target and test a prespecified saturation curve, matching test customer-weeks where available; training eligibility still varies with window length.
Public benchmarks now expose sequence-length axes for user
modeling~\citep{goel2026horizon}, and a production ranking system reports gains that rise monotonically
as history is scaled to $10^4$ events~\citep{guan2025make}. Those gains lie on an event-\emph{count} axis for
engagement ranking. Our saturation result cannot adjudicate that axis, because it is measured in calendar time on activity forecasting
and our input sequences are capped at 750 events, an order of magnitude below that regime
(\S\ref{sec:repro}).
Closest to our claim, \citet{shin2023scaling} vary sequence length, model size, batch size and training-data volume for general-purpose user representations and report both that longer behavior sequences help and that downstream performance does not depend on model capacity alone, concluding that all four factors must scale up in tandem. \citet{shin2023scaling} also compare single and stacked Transformer encoders using the same GPU resources. They measure history in items and leave the saturation point in calendar weeks open.
\citet{dou2026densing} report saturation in users, calendar window and encoder size on
raw payment-behavior sequences, with little gain from extending the window from 60 to 120 days.
Their estimand is the linear-probe AUC, KS and accuracy of a frozen contrastive encoder on
classification tasks; ours is log-space $R^2$ of a forward real-valued total. On our estimand, the 8-to-16-week gain
holds across eight trained families, on matched cohorts wherever a matched comparison exists, and the pooled weekly-aggregate
probe saturates only at 22--24 weeks on the test calendar. We add an internally registered saturation form for this estimand and
a ranking of the axes by magnitude.

\paragraph{Time representation.}
The positional encoding is where a sequence model commits to what a ``position'' is,
and the literature splits on the argument of the phase. One branch keeps that argument
ordinal: sinusoidal absolute positions \citep{vaswani2017attention} and the
learned position embeddings sequential recommenders substitute for
them~\citep{kang2018sasrec,sun2019bert4rec}, with rotary encoding making the ordinal
offset relative \citep{su2021roformer}. A large body of follow-up work reshapes how an ordinal
position enters attention, through its value, phase or magnitude, through the frequencies themselves, or through an
additive bias, while still deriving the argument from the token sequence rather than physical
time. Examples include decay-augmented rotary
(xPos)~\citep{sun2023length}, positional interpolation and frequency rescaling
\citep{chen2023extending,peng2023yarn,ding2024longrope}, learned per-frequency scales (LeRoPE)~\citep{karypis2026lerope}, functional interpolation of
relative positions (FIRE)~\citep{li2024fire}, content-conditioned positions
(CoPE)~\citep{golovneva2024contextual}, and frequency-domain repair for length
generalization (FoPE)~\citep{hua2025fope}. Diagnostic work asks which
rotary frequency bands a trained model actually uses~\citep{barbero2024round}, and a formal result shows that, absent aliasing, no
fixed-frequency rotation over the token index can represent time that advances non-affinely
in that index~\citep{kim2026stretchtime}. The other
branch replaces the argument with physical time: periodic time features
\citep{kazemi2019time2vec}, Bochner-style functional time kernels for event
sequences~\citep{xu2019functional}, continuous-time latent dynamics
\citep{rubanova2019latent,chen2024contiformer}, attention over learned time embeddings
under irregular sampling~\citep{shukla2021mtand}, and multi-scale warping or cross-scale
attention for irregular clinical
streams~\citep{zhang2023warpformer,xiao2024xtsformer}. Closest to the encoding we vary is
the sub-branch that makes elapsed time the rotary argument itself. It spans point
processes~\citep{gao2024rothp}, medical time-series foundation
models~\citep{li2025mira}, masked autoencoders over continuous
positions~\citep{zivanovic2025romae}, and long-horizon multivariate
forecasting~\citep{ibrahim2026ropev1}. In the forecasting work, scaled physical timestamps
replace the integer index inside the rotary operator to preserve relative temporal invariance
under irregular sampling, with the time scales learned per layer. Concurrent work in the same
sub-branch targets routine periodicity in recommendation streams with a random-Fourier rotation
schedule derived from a periodic attention-modulation function rather than a log-linear
frequency schedule~\citep{chen2026clockrope}.
In soft sensing of irregularly sampled industrial processes, a time-aware rotary Transformer
encodes process variables,
positional information and sampling intervals jointly in an embedding layer, and uses a
non-increasing function of the interval in its rotary attention to set the
temporal similarity between representations~\citep{shi2025tart}. Recommendation research has
converged on the physical-time rotary argument from several directions. One deployed ad-ranking
system rotates on the log-scaled gap to the previous event jointly with ordinal position~\citep{hou2026kunlun}, and another replaces position outright with the Unix timestamp~\citep{pardoe2026cadet}. Other papers treat event time and token index as two angle sources
split across dimensions or heads~\citep{wei2025torope}, decompose each timestamp
into coarse-to-fine temporal granularities~\citep{zhang2026rote}, or map the absolute timestamp and its daily and weekly phases to per-plane angles through a learned network~\citep{cheng2026siren}. Concurrently, one paper buckets elapsed time since the sequence's first event at one fixed
granularity per temporal head, beside ordinal rotary heads~\citep{chen2026rest}, and another rotates on the absolute timestamp through a geometric bank of per-plane frequencies with learned coefficients~\citep{liu2026trope}.
Putting physical inter-event time inside attention predates all of these rotary forms.
\citet{li2020tisasrec} already placed the time interval between every pair of
interactions inside self-attention, as a clipped relation embedding alongside
the absolute position embedding, at one per-user interval scale rather than a band set,
and without rotation. They also compared encodings that use positions only, intervals only,
and both. We therefore claim no novelty for multi-scale physical time as such. It has
prior art in several independent lines. The differences in our work are narrower, and we claim priority for none of them.
Where the encoding of \citet{wei2025torope} carries a single rotation per plane, each of our rotary planes superposes one rotary term per band, with amplitudes softmax-normalized across the band set. Our angle argument is
elapsed time from the first retained event, whereas \citet{zhang2026rote} use absolute cumulative calendar counts.
Our setting is long-horizon aggregate regression, while the recommendation work targets
next-item ranking and click-through-rate prediction.
Multi-scale
periods are a recurring inductive bias on the forecasting side too, whether discovered
adaptively from the spectrum~\citep{wu2023timesnet} or, as here, initialized at calendar units.

Our variant (\S\ref{sec:time}) sits in the physical-time branch. Its rotary bands are initialized at
hour/day/week/year over physical $\Delta t$, with learned timescales and softmax mixing. Table~\ref{tab:pe} supports three observations. First, against our rotary encoding as implemented, \emph{some} physical-time parameterization is
the model-side factor in our grid with the largest measured effect on log-space $R^2$. Making the rotary frequencies physical and learnable moves it by
$+0.012$, and constraining them to calendar bands moves it by $+0.0208{\pm}0.0014$
(\S\ref{sec:time}). By comparison, scale
($10\times$ customers) and capacity each move it by ${\leq}0.007$ on 3-seed means (\S\ref{sec:ceiling}), and architecture at a matched objective by ${\leq}0.0052$ at 16 weeks on the full test split, with the clocks as built.
That architecture figure is a seed-42 spread across
five families, the transformer included, and it does not hold at three seeds (\S\ref{sec:limitations}). Second, our $n{=}3$ seeds cannot order the variants \emph{within} the
physical-time branch. The structured variant's
$+0.009$ over an unconstrained learnable control sits inside that control's own seed
spread. Third, the gain is
measured in log space under the primary recipe, and its sign reverses under the
raw-target recipe and statistic (\S\ref{sec:external}). On this axis we thus contribute a variant that sets the argument of the phase and a measurement of how far
a physical-time parameterization can move log-space $R^2$ against a bounded but unresolved ${\leq}0.007$ band.
We do not offer a per-variant accuracy ranking.

\paragraph{Scaling laws.}
The scaling-law literature fits smooth forms to performance as parameters, data, and
compute grow \citep{hestness2017deep,kaplan2020scaling,hoffmann2022chinchilla}; for
ridge regression, asymptotic risk formulas have been derived under
prediction-only self-distillation~\citep{dang2026sdpred}. Throughout this literature, context length is at most
one capacity factor among many. Recent work extends the fits to behavioral
foundation models over user-event sequences but calibrates parameters, data, batch
size, and sampling at a fixed training context length, counted in events rather
than calendar weeks~\citep{gabrielsson2026scaling}. BehaviorGPT-v4, which adopts that
study's compute-optimal guidance, also trains at a fixed context length in
events~\citep{gabrielsson2026behaviorgptv4}. We
scale the conditioning window instead, holding tiers, anchor weeks, target and recipe fixed. Of the two forms we committed to, the better-fitting one saturates.
The pooled weekly
information curve plateaus at 22--24 weeks on the test calendar and stays flat through week 32, and the 24-week
measurement sits closer to the saturating prediction than to the log-linear one
(\S\ref{sec:saturation}, \S\ref{sec:miss}).
A time-series scaling-law analysis derives regime-dependent relationships among the
optimal look-back, dataset size and model capacity~\citep{shi2024scaling}.
Over our tested decade of scale (1M--10M tiers), the window gain is scale-flat within ${\sim}0.006$ $R^2$ in the four families we put
on a scale ladder (\S\ref{sec:universal}). The bound rests on a per-family fit of $R^2$ on log window, log scale and their product, over M8--M16
(M8--M24 for the transformer). The implied window$\times$scale interaction is at most $0.005$ $R^2$, ${\leq}6\%$ of each family's own window main effect. For the three sequence models on the ladder, it is ${\leq}0.0024$. The two recurrent models train on a fixed-size sample at every tier, so their interaction is measured at fixed exposure.
Our fixed-window contrasts do not directly estimate changes in the optimal look-back.
At far larger scale, \citet{zhai2024hstu} report that quality scales as a power law of
training compute across three orders of magnitude on user-action sequences, up to
GPT-3/LLaMa-2 scale. Their ladder raises sequence length together with residual layers,
embedding dimension and attention heads. They report that, within a reasonable range, the
exact hyperparameters matter less than the total training compute applied; that sequence
length matters significantly more in their setting than in language modeling; and that it is
therefore important to scale sequence length and the other parameters in tandem. That
statement about sequence length is qualitative and comes from a single architecture
on a ladder that co-varies window with capacity by design. Such a ladder identifies a compute law and does not separate window from capacity. The statement agrees with our direction but supplies no magnitude ordering
across axes or across architecture families.

\paragraph{Preregistration in machine learning.}
Preregistration, which records a study's hypotheses, design and analysis plan before the study begins, has been proposed for NLP \citep{vanmiltenburg2021preregistering} and
piloted at a dedicated workshop whose protocols were peer-reviewed before any results existed \citep{bertinetto2021preface}.
The rigor literature catalogs the questionable practices a scored preregistration
guards against~\citep{leech2024questionable}. It also documents weak-baseline
overoptimism~\citep{mcgreivy2024weak}, an \emph{inflationary} evaluation bias. The
composition bias we correct runs the other way on our main window gain, which it understates, while it inflates
the 24-to-32-week one. Openly
reported falsified preregistrations have begun to appear in this
literature~\citep{bronder2026instrument,lu2026preregistered}, including a multi-commitment record that
commits numeric predictions in advance and reports falsified ones~\citep{lu2026preregistered}. Our commitments
differ in subject. They concern an input lever of customer-behavior forecasting (the window effect and
its saturation and activity-mixture extrapolations) rather than an evaluation instrument or a
compression method. We registered our four commitments (\S\ref{sec:prereg}) internally, without peer review:
a survival threshold and three numeric predictions.
Three held (Table~\ref{tab:prereg}). The fully blind one, the mixture extrapolation, predicted a
$+0.014$ pooled gain from week 24 to week 32; the observed gain was $-0.0004$. We report the
miss against its own falsifier and do not refit around it.

\paragraph{Event-sequence models.}
The natural models for irregular event streams are temporal point processes (TPPs) and
continuous-time sequence models: the neural Hawkes process \citep{mei2017neural},
attention-based intensity models \citep{zuo2020transformer}, continuous-time
attention \citep{chen2024contiformer}, and the surrounding neural-TPP literature
\citep{shchur2021neural}. Such streams increasingly serve as pretraining data for foundation
models, e.g.\ clinical event sequences aligned with time-series observations~\citep{li2026llm4ehr}
and a single point-process model pretrained jointly on six event corpora, e-commerce streams
among them; its authors present it as an initial step toward event-stream foundation
models~\citep{rezaei2026surf}.
Neural Hawkes and continuous-time attention enter our grid as baselines on the same
cohorts. Both are simplified stand-ins whose time mechanisms were inert in the main runs
(\S\ref{app:baselines}). Model family proves second-order to the conditioning set. At a matched objective, five sequence families land within
${\sim}0.005$ of one another at 16 weeks in seed-42
runs on the full test split with the clocks as built (\S\ref{sec:universal}), and the band does not hold at three seeds (\S\ref{sec:limitations}). A deflationary reading of architecture gains has independent
support in event-sequence forecasting, where Transformer backbones rarely improve
on recurrent ones~\citep{karpukhin2026hotpp}, and in recommendation. In an audit of
eighteen neural models for top-$n$ recommendation
published at top-level conferences, its authors could reproduce only seven with reasonable
effort~\citep{dacrema2019progress}. Comparatively simple heuristics,
such as nearest-neighbor or graph-based methods, could often outperform six of those seven, and the seventh clearly beat those
heuristics but did not consistently beat a well-tuned non-neural linear ranking
method.
Both studies agree with our band only on the direction of architecture gains, because their estimands differ from ours.
One benchmarks next-event and long-horizon event forecasting and the other audits the reproducibility of
top-$n$ recommenders scored on ranking metrics; neither compares model families on a matched-cohort regression task, as we do here.

\section{Data and methods}
\label{sec:methods}

\subsection{Cohorts, extracts, and splits}
\label{sec:data}

The primary dataset consists of customer event streams from a single consumer platform,
exported as fixed cohort extracts. An extract pairs an input window of
$M \in \{8,12,16,24\}$ weeks with a nominal population tier
(1M/5M/10M). Each row is a (customer, target week) pair. The target week, or anchor, is the week a forecast is made from. The row carries $M$ weeks of history
before the anchor and the target after it. An event carries a type code, five categorical attribute codes and two numeric event fields; on this export the six codes reach the transformer as one constant code (Table~\ref{tab:recipe}), except in the corrected-input runs (\S\ref{sec:repro}). Models trained through the data pipeline keep each row's most recent 750 events, except a pair of models retrained with a $1{,}500$-event cap (\S\ref{sec:repro}), and the tabular baselines see six aggregate features of the window.
We extracted each window of the core grid once and never re-extracted it. The
24-week extracts were built after the 8-to-16-week result was in hand, from
the same export, and the saturation prediction they test was committed before they were evaluated
(\S\ref{sec:prereg}). One additional 32-week extract (1M tier), also derived later from the
same export, decides the internally registered saturation-persistence test
(\S\ref{sec:prereg}). We validated it against the M24 extract: on a sample of $34{,}414$ shared
(customer, target week) keys, the two extracts' targets agree to within $2.3\times10^{-9}$. We draw tiers from
non-overlapping ranges of customer sampling buckets, so tiers share no
customers.

The ``5M'' and ``10M''
tiers carry ${\approx}5{\times}$ and ${\approx}10{\times}$ the 1M tier's unique training
customers, the ratios their labels imply. Tier names label bucket ranges and are not exact
unique-customer counts. Test splits hold
480{,}570--480{,}748 rows per window at 1M,
${\approx}2.4$M rows at 5M, and ${\approx}4.8$M
at 10M.

A longer window changes who the model trains on, because the set of eligible customers grows with $M$. At the 1M tier, the eligible training population is
$3.6\%$, $6.1\%$ and $9.6\%$ larger at M12, M16 and M24 than at M8. Longer-window
models therefore train on
up to 10\% more customers over the same number of training rows, and
the added customers are less recently active. Matched-cohort evaluation holds test composition fixed but leaves this
training-side difference in place. To separate this difference from the history each row carries, we retrained the
transformer at 8 and 16 weeks on only the training rows whose customer and anchor are in both windows' training splits
($4{,}314{,}853$ rows, $14.4\%$ fewer than either window's own split), and on each window's own split, at seeds 42,
123 and 456, and scored all twelve runs on the three-window cohort of \S\ref{sec:matched-method} ($n{=}407{,}338$).
The two fixed-row runs train and select on the same rows and targets; only the event history differs.
$D_{\mathrm{int}}$ is the mean over seeds of $R^2(16)-R^2(8)$ with rows fixed, and $E$ is the mean over seeds of
$D_{\mathrm{int}}$ minus the same seed's gain on the windows' own splits. Intervals are 95\% $t$ intervals over the
three seeds. The reading rule (how these runs are analyzed and reported) was fixed before any of them started. $D_{\mathrm{int}} = +0.006$ ($-0.001$ to
$+0.013$) and $E = -0.003$ ($-0.013$ to $+0.007$). With training rows fixed, the gain is not shown at three seeds, so
this contrast does not separate who trains from history, and the eligibility answer rests on the weekly ridge probe.
The fixed-row sets are $14.4\%$ smaller in both windows, so $E$ mixes who trains with how many train;
$D_{\mathrm{int}}$ compares the two windows at equal size. These runs train and score on the corrected inputs (real
category codes and standardized numeric fields), unlike the paper's main checkpoints, so none is compared with a
main-run transformer score, and every main-run number stands. All training inputs predate a 2024-12-06 recoding of
the store field at its source, reported upstream and not measured on this export. Two of the three 16-week test
windows reach back before that date and no 8-week test window does, so part of the retrained models' 8-to-16-week gain
can come from that coding match rather than longer history. The main-run models saw the store field, like every
category field, as one constant code, so this does not reach their gain. The change in the gain from holding training
rows fixed compares models on the same test inputs, so the shared part drops out, leaving only any difference in how
much the models trained on the two row sets use the store field, and in their validation rows.

The M8, M12 and M16 test splits are not nested. Keyed on (customer, target week), the 1M M8
split holds $73{,}232$ rows the M16 split lacks, and M16 holds $73{,}335$ that M8
lacks. Of the M8 split's customers, ${\approx}11\%$ do not appear in the M16 split at
all. The pattern repeats at 5M and 10M and follows from our split construction.
Each split draws a fixed number of eligible customers per anchor week,
so a longer window admits newly eligible customers who displace an equal number of rows.
This is why the two sets of unshared rows, one in each split, are near-equal in size, and why the row
total varies by at most $0.04\%$ across windows while the eligible pool grows.
Consequently, the three-window intersection is a proper subset of \emph{each} split, the M8 split
included (\S\ref{sec:matched-method}), and every matched-cohort claim conditions on that
enumerated intersection (\S\ref{sec:limitations}).

Splits are by customer \emph{and} by time: each customer belongs to exactly one
split, and the splits' target weeks occupy disjoint calendar ranges.
Target weeks run 2024-01-29 to 2024-09-09 in train (nine anchor weeks four weeks apart), 2024-11-04 to 2024-12-30 in
validation, and 2025-02-24 to 2025-04-21 in test. The test split is three anchor
weeks four weeks apart (2025-02-24, 2025-03-24, 2025-04-21). The ranges are separated by
${\approx}8$-week buffer gaps and are identical in every extract. We verified the customer and time splits
by direct join. Across the four core-grid 1M extracts plus
5M and 10M spot checks, train$\cap$test and train$\cap$validation share
zero customers and zero (customer, target week) rows
in all twelve pairwise checks, with zero duplicate rows in any split.
On M8/M16/M24, validation$\cap$test is also empty. Disjointness by customer is therefore
exact on every checked pair, while disjointness in time is weaker. The ${\approx}8$-week buffers
separate target \emph{weeks}, but the 364-day target periods that follow those weeks still
overlap in calendar time across splits, so we verify disjointness by customer and by
target week only (\S\ref{app:calendar} refits the weekly-aggregate ridge probe with no training label overlapping the test period by more than one day).

Because the test anchors are fixed, window length is tied to calendar coverage. The
input window ends at the target week and extends backward, so at every test anchor a longer window also reaches further back in the calendar (\S\ref{sec:limitations}).
Review events are nearly absent from the export before August 2024, so training inputs hold almost none while validation and test inputs hold them.
An upstream export filter dropped nearly all events of adding an item to a cart or list, and review events, in the same block of evening and night hours every day, and the export dated those from the hours just before that block to the next day. Purchases, subscription orders and the target come from a different stream, which neither affected. Every model, window and split reads this one export, and the gap recurs in the same hours of every day, so all comparisons here share it; it shows in the event counts, event dates and gaps between events that the models read.

\emph{Single export.} Every primary-dataset result in this paper comes from training and evaluation on this one
export. Earlier exports of the
same platform define different cohorts (different eligibility filters and test
sizes), so we compare no number here, even qualitatively, against one from another
export.
The public datasets behind the external-validity checks appear where used: the Taobao dataset and its
day-scale extracts in \S\ref{sec:external}, and four public retail datasets that each span more than a year in \S\ref{app:public-ladder}.

\subsection{Target and evaluation}
\label{sec:eval}

The target is total customer activity over the 364 days following the anchor week, a
real-valued total rather than a count, modeled as $\log(1{+}\max(0,y))$ with no winsorization.
It is exactly zero for $4.8\%$--$36.4\%$ of rows depending on the population slice.
Every $R^2$ we report is computed in
this transformed space (``log-space $R^2$''), on the window's full test split unless a matched
cohort (\S\ref{sec:matched-method}) is stated. The two anchor groups of the calendar check (\S\ref{app:calendar}) are scored instead on held-out training customers at training-period anchors, a validation-style evaluation whose $R^2$ values we do not compare with the test-split numbers. We score $R^2$ in log space because the raw
target is heavy-tailed. After back-transforming with $\exp(\cdot)-1$, a few extreme customers dominate the
squared error and the metric swings from seed to seed. Original-scale
scores can then reorder models even when the fit on most customers is unchanged.
The one $R^2$ not in log space is the winsorized raw-target statistic of
\S\ref{sec:external}, and we name it wherever it appears. Every
reported number states its space, and no magnitude is compared across spaces.
On the
matched cohort the denominator is constant, so $\Delta R^2$ is proportional to
$\Delta$MSE, and the matched log-space MSE column (Table~\ref{tab:matched-cohort}) is
the same result expressed as squared error.

\emph{Objective-matched comparison rule.} At fixed architecture, the choice of training objective
moves log-space $R^2$ by
$0.03$--$0.05$, more than any architecture intervention we measured. Part of that
spread is definitional: on
a fixed cohort, log-space $R^2$ is an affine transform of log-space MSE, so on that
cohort minimizing squared error and maximizing the reported metric are the same
objective. Only the squared-error variant is therefore scored on the statistic that its swept loss (the first term of its training
objective, Table~\ref{tab:recipe}) optimizes; the Huber and pinball variants are scored on a statistic their losses do not optimize. That built-in alignment advantage does not
guarantee that squared error wins on held-out data.
A monotone map fit on validation closes most of the objective gap (\S\ref{sec:objcal}), yet at seed 42 the best affine correction closes
under a third of it, which is not what the affine relation suggests.
We therefore treat the spread as a limit on which comparisons are valid and make no claim
that squared error produces the better model. A
gap between two model families counts as a win only when both sides match on objective family,
target transform, cohort and data export, and evaluation protocol. Where both ran as managed cloud jobs, they must also share a training build (\S\ref{sec:repro}). When any
of these conditions fails, the two sides are shown as within-family curves. The comparison rule disqualifies
the transformer's apparent margin over the recurrent models, which does not survive once the losses are matched (\S\ref{sec:lossnull}).\par
\emph{Feature normalization.} Normalization statistics are computed on training
splits only, so no validation or test information enters them. The M8--M24 extracts use the export's scaler, fit on each window's full
training split, and the 32-week extract uses one we fit on its own training split. The recurrent and continuous-time baselines standardize the two numeric
event fields with these statistics; the transformer receives them unstandardized.

\subsection{Matched-cohort evaluation}
\label{sec:matched-method}

A longer input window changes \emph{who} is in the test set as well as \emph{what} the model
sees, as does any intervention that changes eligibility. Longer
windows admit lapsed customers. That shift in test composition alone raises the standard deviation of the log target (log-SD) on the full test set,
$2.16 \to 2.39 \to 2.56$ (and $2.80$ at M24), inflating the $R^2$
denominator. A raw cross-window delta confounds what the model sees with who is scored.
We instead key every test row on (customer, target week), intersect the compared
windows' test sets, and score every window's model on that fixed set.
At 1M the three-window cohort has $n{=}407{,}338$ rows, $84.7$--$84.8\%$ of each test split (the splits are not nested,
\S\ref{sec:data}), and a four-window variant has $372{,}778$ rows. At 10M the three-window cohort has
$n{=}4{,}069{,}475$ rows ($84.8\%$). In each three-window cohort
the targets agree across the windows' extracts to within
$8{\times}10^{-8}$ at 1M and $1.1{\times}10^{-7}$ at 10M (maximum absolute difference). The denominator
is constant by construction, with target log-SD $2.1597$ at every window of that cohort at 1M and $2.16175$ at 10M. On the
four-window variant the log-SD is $2.1592$, and in $109$ of its $372{,}778$ rows the M24 and M8 targets
differ, by up to $0.31$ in log space. This discrepancy moves the Huber $\Delta$(M24$-$M8) by $6{\times}10^{-7}$: re-scoring
that delta with either version of the target leaves it unchanged at four decimals. Any gain in $R^2$ on a matched cohort is therefore
error reduction on the same customers. In both 1M matched cohorts, the ones we measured, every row has at least one
event in the 8 weeks before its target week, so matched window gains are gains for customers active in those 8 weeks,
and most rows that matching drops have no such event (\S\ref{app:cohort-keeps}).

Every matched transformer and recurrent number must first pass two checks.
\emph{Alignment}: we map each row of the saved
prediction arrays back to its (customer, target week) key by reconstructing the evaluation order, and the targets recomputed under that mapping must
equal the saved target arrays elementwise
(maximum difference $0.0$). \emph{Reproduction}: the full-set $R^2$ computed from those arrays \emph{before} the cohort
mask is applied must reproduce the value the training run reported. The recomputed value agrees with it to the printed five decimals at 1M, with a residual
under $1.5{\times}10^{-7}$, and lies within $5{\times}10^{-8}$ of it at 10M. A matched evaluation that fails either check is discarded without repair.
The offline tabular baselines
retain no per-row predictions, so their matched results carry a reproduction check
against the values the ridge training runs reported (\S\ref{sec:baselines}) but no alignment check.
Wherever a matched evaluation exists, we report the matched cross-window delta beside the full-set one, and every
table states which evaluation set it uses. Where a family or contrast has no matched evaluation, we cite the full-set delta. These unmatched cases are the
continuous-time rows and the single-feature linear row of
Table~\ref{tab:family-deltas}, every contrast at 5M, the 10M ridge and recurrent ladders and M24 point, the MS-RoPE window ladder of \S\ref{sec:time}, and the cross-recipe transfer of
\S\ref{sec:external}. Where both evaluations exist in
log-space $R^2$, matching enlarges every gain from M8, so for those comparisons the full-set estimate is the
conservative one; for the transfer scored on the winsorized raw target we claim no direction.

Following \citet{petersen2024are}, we test the matched window gain against a randomized benchmark. Ours swaps the M8 and M16 models' predictions on all of a customer's rows, independently for each customer with equal odds. The two windows' targets are identical on all $407{,}338$ matched rows, so both models' predictions are scored against the same targets. This permutation test assumes that the two models' predictions are exchangeable within customers under the null~\citep{hemerik2018exact}. The observed log-space $R^2$ gain of $0.0959$ exceeded all $1{,}000$ permuted gains ($p = 1/1001$). On $200$ permuted datasets, the nominal 95\% customer-clustered bootstrap interval ($B{=}500$) excluded zero in $15$ ($7.5\%$). Both checks use the seed-42 model pair.

\subsection{Reproducibility}
\label{sec:repro}

\emph{Determinism.}
On the primary dataset, the transformer, recurrent and linear paths are deterministic end-to-end at fixed seed. A repeated
linear-model run reproduces $R^2$ to
17 significant digits, and six groups of repeated runs across the main grid (independent runs of the same configuration)
are bit-identical; each group counts once in any mean. Seed-level intervals on these deterministic paths therefore carry training-seed variance only. These intervals omit the data-sampling and hyperparameter variance that also moves benchmark results~\citep{bouthillier2021}.

\emph{Build anchoring.} Because the training environment can drift over a long study, we treat every
change to it as untrusted until verified.
Outside the three exceptions noted below, every container build behind a cited result passes a
bit-exactness check. We retrain a fixed reference configuration on the
new build, and no result from that build is citable until the retrained model reproduces the reference $R^2$ bit-for-bit
(in single precision). Eight builds have passed this check, and the two
carrying the loss- and budget-equalized comparisons also passed it on the
recurrent paths (\S\ref{sec:lossnull}).
The tested reference metrics therefore agree exactly across builds. That agreement does not establish equivalence for untested configurations. 
For comparisons against the reference configuration, the only nonzero difference comes from
floating-point reduction order.
That difference measures $1.5{\times}10^{-8}$, below citation precision, and every test score in these
comparisons was computed by a single evaluation process.
The three exceptions to the bit-exactness check are the corrected-schedule build (\S\ref{sec:ceiling}), which also
carries the four baseline families' three-seed runs (\S\ref{sec:limitations}); the build behind the Taobao dataset's
transformer results (\S\ref{sec:external}); and the corrected-input build behind the working-clock baselines
(\S\ref{sec:universal}), the fixed-row runs (\S\ref{sec:data}), the transformer refit of the 24-to-32-week gain
(\S\ref{sec:limitations}), the transformer calendar check (\S\ref{app:calendar}), the transformer runs on the other
public datasets (\S\ref{app:public-ladder}), the learning-rate sweep and the window check across model size (both below). The Taobao results are instead cited after
passing the alignment and reproduction checks of \S\ref{sec:matched-method}, applied to predictions regenerated from
each run's checkpoint after verifying the checkpoint's identity. For these results, reproduction holds to
single-precision rounding. The corrected-input build gives the transformer inputs the reference run never had (real
category codes, and standardized price and quantity), so the reference cannot be reproduced on it bit-for-bit. Each
of its results is cited only after passing the checks its reading rule fixed before the results were read: each run
is the pinned configuration on this build, it trained from its own start or from its own saved state, and each score
comes from that run's own model on the rule's test rows.

\emph{Seeds.} On the primary dataset the seed set is fixed at $\{42,123,456\}$, except the transformer calendar check of \S\ref{app:calendar}, which adds $789$ and $1011$. On the Taobao dataset, we added replacement seeds $789$ and $1042$ to the Huber-objective runs, in place of runs that
collapsed to the constant-zero predictor. The Huber runs' healthy seeds therefore differ by window and
are reported unpaired (\S\ref{sec:external}). The transformer's full-set results at M8, M12,
and M16 are 3-seed at all three population tiers, with per-seed ranges as wide as $0.0125$ (at M12-1M).
The M24 results are also 3-seed at all three tiers. The 32-week extension results are
cited at seed 42 only. The 16-week capacity-probe runs are 3-seed, and the 8-week width and depth doublings are single-seed. On the matched cohort the main
$\Delta$(M16$-$M8) contrast replicates at three seeds at both 1M and 10M, as does the stacked MSE and MS-RoPE configuration.
The MSE and Huber matched transformer entries for M8--M16 in Table~\ref{tab:window-grid} are 3-seed,
whereas the pinball matched row, the matched M24 entries, and the matched
$\Delta$(M16$-$M8) pairings are seed-42 evaluations.
Entries
with $\pm$ are 3-seed means $\pm$ population SD; all others not marked as multi-seed are single-seed point
estimates, flagged as such where cited.
At $n{=}3$ the printed population SD is $\sqrt{2/3}{\approx}0.82\times$ the sample SD and is
not a standard error: a two-sided 95\% $t(2)$ interval on a 3-seed mean has half-width
${\approx}3.0\times$ the printed SD. Every model family shares seed 42, and no configuration is
re-run for a better draw; the Taobao Huber-objective results are reported with and without their replacement seeds.
Where a single-seed result contributes to a band, as in the architecture spread and the scale band of
Table~\ref{tab:scale-10m}, we report that band as an observed interval against the detection floor of \S\ref{sec:limitations}.
Such a band is not a resolved equivalence bound.

\emph{Training recipe.} We fixed the recipe before running the grid, and no reported configuration is tuned
individually. The exceptions are two checks on the capacity probe's larger model: a five-configuration retune
(\S\ref{sec:ceiling}) and a five-point learning-rate sweep (below).
Table~\ref{tab:recipe} collects the settings.
Within a comparison, transformer runs differ only in the stated factor: window, objective, positional encoding, a single ablated input or head component, tier, or seed.
For the capacity probe, the stated factors are width, depth, and a $100{\to}500$-step warmup (\S\ref{sec:ceiling}).
Beyond these factors, the only differences are the separately extracted M32 training data of the $24{\to}32$ pair (\S\ref{sec:miss}), the smaller effective batch of the $1{,}500$-event-cap pair and the realized schedule of the tier contrast (\S\ref{sec:limitations}), the corrected-schedule build (\S\ref{sec:ceiling}), and the corrected-input runs of \S\ref{sec:repro}.
The recipe inherits an existing configuration's
architecture and optimizer settings.
The recipe transfer of \S\ref{sec:external} restores that configuration's objective, winsorization, and
target statistic, along with two further settings that
\S\ref{app:objective} states in full.

The recurrent and continuous-time baselines differ from the transformer as Table~\ref{tab:recipe} shows and in training budget (see the compute paragraph of \S\ref{sec:repro}). The tabular baselines share
three of six features. Where the ridge takes the purchase rate and the mean and SD of the second numeric event field, the GBM takes the purchase count and that field's sum and sum of squares.

\begin{table}[htbp]
\caption{The fixed training recipe. One recipe serves every transformer run except the recipe-transfer runs (\S\ref{app:objective}), the $1{,}500$-event-cap pair (\S\ref{sec:limitations}), the runs on the corrected-schedule build (\S\ref{sec:ceiling}),
the corrected-input runs (\S\ref{sec:repro}), and the public-dataset runs, whose changes are stated with them
(\S\ref{app:public-ladder}).}
\label{tab:recipe}
\centering\footnotesize
\begin{tabular}{p{0.27\linewidth}p{0.64\linewidth}}
\toprule
\multicolumn{2}{l}{\emph{Transformer}} \\
\midrule
Architecture & $d_{\mathrm{model}}{=}128$, 2 layers, 4 heads, feed-forward width
$4d_{\mathrm{model}}{=}512$, dropout $0.1$, attention pooling; full graph and
parameter accounting in \S\ref{app:arch} \\
Optimizer & AdamW, learning rate $5{\times}10^{-4}$, weight decay $0.01$ \\
Schedule & at the 1M tier and the ${\approx}8{,}192$ effective batch, effectively constant $5{\times}10^{-4}$ after a 100-step linear warmup. A cosine
decay to $1\%$ of peak was configured, but its period ($99{,}900$ optimizer steps) exceeded the
number of steps a 1M run actually takes (${\approx}9{,}760$) by ${\approx}10{\times}$, so the schedule
traversed under $10\%$ of its period and ended within $3\%$ of peak.
Every 20-epoch transformer run with the 100-step warmup outside the corrected-schedule build of \S\ref{sec:ceiling} uses the same period,
and effective batch is matched across the runs of each contrast above, so this behavior is common to the family rather than a difference between compared runs.
The down-budgeted runs of \S\ref{sec:lossnull} have a period
half as long ($49{,}900$ steps) and, at batch $128$ on one process, take ${\approx}394{,}000$ steps, so their rate cycles between floor and peak about four times. The realized step count does grow with
the training set, so the fraction of the period traversed differs across population tiers
(\S\ref{sec:limitations}). Gradient-norm clip $1.0$; no mixed precision \\
Batch / epochs & 256 per process ($32$ ranks, ${\approx}8{,}192$ effective; the Taobao runs of \S\ref{sec:external} use one process, $256$ effective), 20 epochs; the down-budgeted runs of \S\ref{sec:lossnull} use batch $128$ on one process for 10 epochs \\
Huber $\delta$ & $1.0$ (in the log-target space) \\
Pinball $\tau$ & $0.5$ (the median; the same $\tau$ serves the pinball-objective runs
and the auxiliary per-event head) \\
Training objective & a fixed-weight sum of the swept loss on the aggregate prediction
and an unswept auxiliary per-event loss; the ``objective'' lever moves the first term
only. Both terms, their weights and the gating are specified in
\S\ref{app:objective} \\
Event featurization & per event: a type code for one of four kinds (adding an item to a cart or list, a purchase, a review,
or a subscription order) and five categorical attribute codes
(embedded at dimensions 4--16; on this export all six reach the model as one constant code, \S\ref{app:notfixed}) plus two numeric event fields,
concatenated with two declared time-feature channels that are identically zero on
this export (\S\ref{app:arch}); fields absent for an event type are filled with
$0.0$ before encoding; most recent 750 events (cap paragraph below) \\
\midrule
\multicolumn{2}{l}{\emph{Recurrent and continuous-time baselines}} \\
\midrule
Architecture & hidden width $128$ in all four (the same config field as
$d_{\mathrm{model}}$); GRU and LSTM 2-layer, dropout $0.1$; ContiFormer 2 layers, 4
heads, feed-forward width $4{\times}$; Neural Hawkes a single decay-gated cell. All four
read the four numeric per-event channels (the two numeric event fields standardized, \S\ref{sec:eval}), not the categorical embeddings, and end
in the same $128{\to}64{\to}1$ head plus the transformer's two-scalar linear anchor on the
input-window activity total, but ungated and trained in the same AdamW group as the rest of the model;
two of those four channels are identically zero
here, which in the main runs leaves both continuous-time models with no inter-event interval. The two
continuous-time models are simplified stand-ins for the published architectures
(\S\ref{app:baselines} lists what they omit) \\
Optimizer / schedule & AdamW, learning rate $3{\times}10^{-4}$ (GRU, LSTM) and $1{\times}10^{-4}$ (ContiFormer,
Neural Hawkes), weight decay $0.01$, cosine annealing, traversed in full: the rate reaches ${\approx}2\%$ of peak by the final epoch, in
contrast to the transformer schedule above (\S\ref{sec:limitations}) \\
Batch / epochs & 128, single process, 10 epochs (own budget) \\
Training objective & a single term on the aggregate prediction (Huber, $\delta{=}1$; squared error in the objective-equalized runs of \S\ref{sec:lossnull}), with no auxiliary per-event term and no gate (\S\ref{app:objective}) \\
\midrule
\multicolumn{2}{l}{\emph{Tabular baselines (six aggregate features)}} \\
\midrule
Features & log-scaled input-window activity total, event count, per-event activity summed in log space, purchase
rate (the share of events with positive activity), and the mean and SD of the second numeric event field, all computed over the input
window \\
Ridge & features standardized; regularization strength cross-validated over 25
log-spaced values in $[10^{-3}, 10^{3}]$, closed form \\
GBM & gradient-boosted trees, 400 trees, learning rate $0.05$, 63 leaves (fixed,
untuned) \\
\bottomrule
\end{tabular}
\end{table}

\emph{Parameter counts.} The GRU and LSTM baselines are $2.4$--$3.1\times$ smaller than
the transformer, and Neural Hawkes, a decay-gated recurrent model, is smaller still (Table~\ref{tab:params}). The cross-family
convergence band therefore does not come from the baselines matching the transformer's parameter count.
Counts are architecture properties, invariant across window, seed, and
objective. Totals include small non-trainable buffers such as rotary frequency
caches (${<}0.1\%$).

\begin{table}[ht]
\caption{Per-family parameter counts, read from the trained checkpoints (M16-1M
runs) as sums over state-dict entries. A module shared across layers, such as the rotary
encoding, contributes once for the model and once per layer that references it. The
MS-RoPE row's $+36$ is therefore $3\times12$ state-dict entries that hold only $8$ added trainable
parameters (\S\ref{app:pe}). Tabular models are listed by their fitted degrees of
freedom.}
\label{tab:params}
\centering\small
\begin{tabular}{lr}
\toprule
\textbf{Family} & \textbf{Parameters} \\
\midrule
Transformer (d128, 2 layers) & $491{,}248$ \\
\quad + MS-RoPE variant & $491{,}284$ \\
GRU (2-layer) & $158{,}851$ \\
LSTM (2-layer) & $209{,}027$ \\
ContiFormer & $405{,}523$ \\
Neural Hawkes & $93{,}443$ \\
Capacity probe (d512, 8 layers) & $26{,}479{,}216$ \\
Ridge (6 features $+$ intercept) & $7$ \\
Single-feature linear & $2$ \\
LightGBM (400 trees $\times$ 63 leaves) & ${\leq}25{,}200$ leaf values \\
\bottomrule
\end{tabular}
\end{table}

\emph{Compute resources.} Except for the offline fits, the two recomputed 10M ridge fits
(\S\ref{sec:universal}) and the evaluation jobs named at the end of this paragraph, every run was a managed cloud job on
eight-GPU (V100) nodes. Main-grid transformer training runs on the primary dataset
used four-node distributed data parallelism (32 ranks), as did
the transformer-budget recurrent retrains of \S\ref{sec:lossnull};
the recurrent and continuous-time baselines at their own budget (GRU, LSTM, ContiFormer, Neural Hawkes)
ran as a single process on one node, as did the down-budgeted transformer runs of
\S\ref{sec:lossnull}, the Taobao runs, evaluation jobs,
and 10M linear
refits. The paired working-clock runs of ContiFormer and Neural Hawkes (\S\ref{app:baselines}) used 8 ranks, and the $1{,}500$-event-cap pair (\S\ref{sec:limitations}) trained at a smaller effective batch.
The evaluation jobs of the other runs on the corrected-input build were single-process jobs on one A10G GPU each,
except four fixed-row and three H\&M jobs, which ran on one L4 GPU each.

Training budgets differ between the transformer and the baselines on
three axes: batch size, epochs, and data exposure. The exposure gap biases \emph{against} the baselines.
Main-grid transformer runs train at a per-rank batch of 256 over 32 ranks
(${\approx}8{,}192$ effective) for 20 epochs. Each epoch streams a balanced subset of the
shards rather than the whole split, covering ${\approx}79\%$ of the training rows at 1M (32 of the split's 40 shards) and
${\approx}96\%$ at 10M. The dropped shards rotate across epochs, so every row is reached over the course of a run.
At their own budget, the baselines train as a single process at batch 128
for 10 epochs, reading only a fixed, RAM-cached sample of at most $384{,}000$
rows at every population tier. That sample, $3{,}000$ cached batches reused every epoch,
covers ${\sim}1/13$ of the 1M-tier train split and a smaller share at
5M/10M.
On a thirteenth of the data the baselines still land
within ${\sim}0.005$ of the transformer \emph{at 16 weeks} on the full test split
($0.015$ at 8 weeks; seed 42), so no transformer advantage outside the cross-family convergence band can hide behind the exposure gap.
At the 1M tier we close the exposure, batch and epoch gaps together by retraining both recurrent families at the transformer's full budget and
near-full data exposure. The result for each recurrent family and window moves by at most $+0.002$
(MSE objective; \S\ref{sec:lossnull}). The Huber-trained baseline rows, calibrated cross-family deltas at 5M/10M, and the baseline
scale comparisons (\S\ref{sec:ceiling}) still inherit the fixed exposure cap
alongside the batch and epoch gaps.

Wall-clock time grows with population tier: on the base d128 recipe, 20-epoch 1M-tier runs
complete in 2.2--2.6 hours at 16--24-week windows, with evaluation jobs of 11--15 minutes; the
8-layer d512 capacity probe, at the same tier and a 16-week window, takes ${\sim}11$--$14$ hours. The 10M-tier
transformer runs take roughly a day,
and the single-node 10M linear refits take ${\sim}20$--$26$ hours; the 16-week refit
did not finish within ${\sim}30$ hours and was recomputed.
The campaign comprises several hundred training and evaluation jobs.

For the transformer, the swept levers also differ in cost. Lengthening the window from 8 to 16 weeks, the largest lever we measured, needs no extra training compute.
Growing the population from 1M to 10M takes $10\times$ the data, and the
capacity probe uses ${\approx}54\times$ the parameters. Only the window comparison is cost-normalized (\S\ref{sec:limitations}).

\emph{Realized learning-rate schedule.} At the 1M tier and the ${\approx}8{,}192$ effective batch, the learning rate is effectively
constant at its peak after the 100-step warmup rather than decaying to the configured floor.
The cosine period is set by a fixed estimate of
$5{,}000$ steps per epoch, and the scheduler advances once
per optimizer step, so the realized schedule
depends on how a run's true step count compares with the estimate. At this tier and batch the
true count is under $500$ steps per epoch, so a 20-epoch run traverses under a tenth of
its period and its per-epoch learning rate falls from $5.00{\times}10^{-4}$ only to
$4.89{\times}10^{-4}$. The compared runs in the 1M window, objective,
capacity, and seed contrasts follow matched realized schedules, apart from the capacity probe's longer warmup. Because the estimate
is a constant, runs at one population tier with a common warmup follow the same realized trajectory when they share per-rank effective batch, epoch budget, and rank count.
The match holds to within ten optimizer steps (the 20-epoch step counts run $9{,}762$ to $9{,}769$). Every run in these contrasts shares all three settings. The capacity probe reaches the matched
effective batch through gradient accumulation, but its $500$-step warmup leaves it ending at
$97.9\%$ of peak against $97.7\%$. Two asymmetries do
remain. First, steps per epoch grow with tier size at fixed effective batch, so the 5M and 10M
runs complete a larger share of the same period than the 1M runs. Against a period of
$99{,}900$ optimizer steps, a 1M run takes ${\approx}9{,}760$ steps and a 10M run
${\approx}112{,}000$, so the 10M runs complete one full period and then overrun it.
Because the cosine is periodic and unclamped, the rate turns back up over that overrun. It still ends the final epoch at
$4.5\%$ of peak, against $97.7\%$ at 1M, so the rebound leaves the direction of the asymmetry unchanged.
To the extent that completing the decay helps a final metric, the scale bound
is conservative for our claim that scale buys little.
A budget-matched pair on one training build tests this. Holding the optimizer-step budget fixed leaves the scale delta essentially where it
was (\S\ref{sec:limitations}). Second, the recurrent
and continuous-time baselines run their own per-epoch cosine, which does reach its
floor. On this axis the cross-family comparison therefore handicaps the transformer, and that handicap pulls the
families together (\S\ref{sec:limitations}).

\emph{A learning-rate sweep of the capacity probe.} The capacity configurations share one
recipe, so the $54\times$ point bounds capacity at that recipe only. A separate set of runs re-tunes the
larger model's peak learning rate over a pre-declared five-point grid at fixed budget: d512 with 8 layers at
$6.25\times10^{-5}$, $1.25\times10^{-4}$, $2.5\times10^{-4}$, $5\times10^{-4}$ and $10^{-3}$, against d128 with 2
layers at its own $5\times10^{-4}$, three seeds each, on the 16-week 1M extract, with a 20-epoch budget,
a cosine schedule sized to reach 1\% of the peak rate at the budget's end, early stopping with patience 8 (every run
stopped between epochs 9 and 16), and the checkpoint picked on validation log-space $R^2$. The
reading rule was fixed before any of these runs was scored. On the three-seed validation mean, the d512 pick is
$5\times10^{-4}$, inside the grid, with best epochs 2, 5 and 7. On the full test split, the paired difference in
test log-space $R^2$, d512 at the picked rate minus d128, is $+0.0018$ (95\% $t$ interval over the three paired
seed differences, $-0.0008$ to $+0.0044$, covering seed-to-seed noise only), inside the $\leq0.007$ band we quote for scale and capacity (\S\ref{sec:limitations}): the capacity
bound survives a 5-point learning-rate sweep. No other d512 rate moves the validation mean by more than twice the
picked configuration's seed spread, so this grid cannot show that learning rate was what held the larger model
back, and the fixed-recipe caveat is narrowed rather than retired. Test log-space $R^2$ of every configuration
(three-seed mean, population SD in parentheses), of which only the picked one enters the difference:
\begin{center}\small
\begin{tabular}{llc}
\toprule
Model & Peak learning rate & Test log-space $R^2$ \\
\midrule
d512, 8 layers & $6.25\times10^{-5}$ & $0.6376$ ($0.0009$) \\
d512, 8 layers & $1.25\times10^{-4}$ & $0.6374$ ($0.0008$) \\
d512, 8 layers & $2.5\times10^{-4}$ & $0.6382$ ($0.0011$) \\
d512, 8 layers & $5\times10^{-4}$, picked & $0.6405$ ($0.0012$) \\
d512, 8 layers & $10^{-3}$ & $0.6379$ ($0.0016$) \\
d128, 2 layers & $5\times10^{-4}$ & $0.6387$ ($0.0011$) \\
\bottomrule
\end{tabular}
\end{center}
These runs train and score on code that gives the model real category codes and standardized quantity and price,
where the paper's other transformer runs on the main data had one constant code and raw values (\S\ref{sec:data},
\S\ref{sec:eval}), and they anneal and
select as the corrected re-run of \S\ref{sec:ceiling} does; at $5\times10^{-4}$, both models are the paper's two base
configurations re-run this way. So they measure the capacity gap with a tuned learning rate on those inputs, and
their levels are not comparable to the paper's other transformer results, beside which they are not set. They use the
rotary encoding as implemented (\S\ref{app:pe}), the d128 model keeps its own recipe, and the joint capacity$\times$data
setting stays untested. According to the dataset's maintainers, the store field was re-coded on 2024-12-06, partway through the test
inputs. Every training input is before that date, and the checkpoints, and the d512 learning rate, were picked on
validation inputs mostly before it. Both widths see the same change, so its shared part drops out of the gap, but it
can move each width's test $R^2$, and the gap can carry any difference in how the two models use the store field.

\emph{The window gain at two model sizes.} A separate set of runs crosses the window with model size: the base model (d128, 2 layers) and the $54\times$ model (d512, 8
layers), each trained at 8 and 16 weeks of history on the 1M extracts at seeds 42, 123 and 456; the 16-week runs are
the learning-rate sweep's runs at $5\times10^{-4}$. Both windows' models are scored on the three-window matched cohort
of \S\ref{sec:matched-method}, the same $407{,}338$ test customer-weeks with the same targets, so the gain is not a
change in who is scored. The two windows are separate extracts of one export, so the models still train on different
rows. Write $G$ for the mean over seeds of matched test log-space $R^2$ at 16 weeks minus 8 weeks, for one model size, and
$I = G_{54\times} - G_{\mathrm{base}}$. The reading rule was fixed before any of these runs was scored. Every
configuration's seed spread is at most $0.0015$ (population SD), well under the rule's $0.0043$ cutoff, and no run's
best epoch is its 20th. On these inputs, 16 weeks of history still beats 8 for the base model:
$G_{\mathrm{base}} = +0.0109$ (95\% $t$ interval over the three per-seed differences, $+0.0071$ to $+0.0148$), the same
sign as the window gain of \S\ref{sec:history}. No interaction with model size is shown, and any interaction is smaller
than the $\pm0.007$ band we quote for scale and capacity (\S\ref{sec:limitations}): $I = -0.0014$ ($-0.0039$ to $+0.0012$). As a descriptive value,
$G_{54\times} = +0.0096$ ($+0.0037$ to $+0.0154$). On the full test splits, which do not hold the same customer-weeks at
the two windows, the gains are smaller (descriptive values): $G_{\mathrm{base}}$ $+0.0014$ ($-0.0008$ to $+0.0037$),
$G_{54\times}$ $+0.0025$ ($-0.0046$ to $+0.0097$), $I$ $+0.0011$ ($-0.0046$ to $+0.0067$). Each split holds about
$73{,}000$ customer-weeks outside the matched cohort. At 16 weeks most of them have no event in the 8 weeks before
their target week (\S\ref{app:cohort-keeps}), and they score lower; at 8 weeks the scores barely change with or without
them. So the full-split gain mostly reflects who is scored, which is why the reading rule uses the matched cohort.
These runs train and score on the same corrected code as the learning-rate sweep, which gives the model real category
codes and standardized quantity and price, where the paper's other transformer runs on the main data had one constant
code and raw values (\S\ref{sec:data}, \S\ref{sec:eval}). So they measure the window gain on those inputs, not on the
paper's, and no other score or gain in the paper is set beside them. Both sizes train at peak learning rate
$5\times10^{-4}$, the base recipe's; for the $54\times$ model that is also the sweep's pick at 16 weeks, nothing was
tuned at 8 weeks, and the base model's rate is not tuned. They use the rotary encoding as implemented
(\S\ref{app:pe}), at the 1M tier, at 8 and 16 weeks only, on the three-window matched cohort. According to the dataset's maintainers, the
store field was re-coded on 2024-12-06 without older events being updated, and these runs read that field. Every
training input is before that date; at two of the three test target weeks the 16-week inputs reach back before it and
the 8-week inputs do not, so part of the 8-to-16-week gain can be a train-to-test match in store coding rather than
history. The interaction compares two such gains on the same test inputs, so it carries any difference in how much the
two sizes use the store match. The paper's other transformer runs on the main data had one constant category code, so
they never read the store field. The intervals cover seed-to-seed noise across three seeds only.

\emph{Input truncation cap.} We truncate event sequences to the most recent 750
events at every input window, for every model trained through the data pipeline except the pair retrained with a $1{,}500$-event cap (\S\ref{sec:limitations}). The offline
gradient-boosted-tree fit keeps the oldest 750 events instead, which moves its matched-cohort $R^2$ by $3.6{\times}10^{-5}$. The share of test customer-weeks that hit the cap grows with the window:
$0.19\%$ (M8), $0.91\%$ (M16), $2.36\%$ (M24), $3.91\%$ (M32). On
the matched cohort, $3.0\%$ of customers already hit the cap at M24 and
got no new input from the longer window. On a smaller, less active sample of $33{,}703$ rows from the M24${\cap}$M32 cohort,
the cap truncates $24.2\%$ of the event mass added by lengthening the window from 24 to 32 weeks.
That truncation is concentrated among the sample's most active customers. A fixed
cap therefore attenuates event-weighted estimates of the longest-window gains
(\S\ref{sec:limitations}). The weekly-aggregate ridge probe behind the saturation verdict is
uncapped, so the bend it measures on the test calendar is not a cap artifact.

\paragraph{Code and data availability.} The 1M/5M/10M event-sequence corpora are
proprietary customer behavioral data and are not releasable; the Taobao corpus and the four
retail datasets of \S\ref{app:public-ladder} (Online Retail II, The Complete Journey, CDNOW and H\&M), each spanning more than a year, are public.
We release no data, models, or individual-level predictions. Every
reported result is a population-level aggregate. Training and evaluation code is internal and is not
released, so a written specification takes its place.
Table~\ref{tab:recipe} and \S\S\ref{sec:eval}--\ref{sec:repro} give the fixed recipe, seed set
and evaluation protocol. The longer specifications behind the
reported numbers are in \S\ref{app:instruments}: the composite training objective, the transformer
graph and its parameter accounting, both physical-time encodings, the weekly-aggregate
ridge probe, the four baseline families, checkpoint selection and retention, the
determinism settings, and the Taobao input mapping. Two of those families are
simplified stand-ins for the published models, and
\S\ref{app:baselines} lists the simplifications. \S\ref{app:notfixed} records where the input pipeline departs from that
specification. The internal registration artifacts for the four commitments are
content-addressed, with SHA-256 digests printed in
\S\ref{sec:prereg}.

\subsection{Internal registration}
\label{sec:prereg}

We registered four commitments internally before their deciding evaluations
ran~\citep{vanmiltenburg2021preregistering}. The one fully blind prediction, the mixture tail of
commitment (iii), failed on its own registered falsifier. The other three held under the
qualifications given below and in Table~\ref{tab:prereg}.

(i)~\emph{Composition
survival}: the window effect would count as robust to test composition only if the matched
$\Delta$(M16$-$M8) remained
$\geq{+}0.05$ in log-space $R^2$. We set this bar before the transformer's matched
re-evaluations ran, and the matched gain cleared it (Table~\ref{tab:prereg}).
(ii)~\emph{Saturation form}: we fit a saturating-exponential and a log-linear form to the
$R^2$ curve of the weekly-aggregate ridge probe over $k{=}1..16$ weeks. Before the 24-week
extract was evaluated, we committed their predictions for matched $R^2(24)$, $0.4345$ for the
saturating form and $0.4952$ for the log-linear. A single M24 measurement scores the two, which sit $0.06$ apart at $k{=}24$.
In level space, the observed $R^2(24){=}0.4296$ lands within $0.005$ of the
saturating prediction. Two caveats apply. First, we fit the forms on the three-window matched cohort, the customer-weeks present in all of the
M8, M12 and M16 test sets, but scored them on the M24 extract's four-window cohort, which keeps
only those also present in the M24 test set. On that extract the $k{=}16$ level sits $0.028$
lower ($0.4426 \to 0.4146$), a shift in who is in the training extract rather than in who is
tested (\S\ref{sec:saturation}). On the \emph{gain} $k{=}16{\to}24$, the
observed $+0.015$ sits between the saturating ($-0.008$) and log-linear ($+0.053$)
predictions, nearer the saturating form but matching neither. Each predicted gain is that form's $R^2(24)$ less the
three-window cohort's $k{=}16$ level ($0.4426$), so the negative saturating value reflects its fit sitting below that
observed point, not a predicted decline. On either reading, the within-extract curve on the test calendar has clearly
flattened by $k{=}22{\to}24$, where its marginal gain is $+0.0001$/week. In the validation-style calendar check of
\S\ref{app:calendar}, refit on holiday-free training-period anchors, the same probe's returns still diminish, but the
sharp bend is not shown to persist, and no public dataset of \S\ref{app:public-ladder} on which we assess the bend shows it.
Second, this commitment had less discriminating power than a $0.06$ separation suggests. The
two forms were nearly tied on the data they were fit to (in-sample RMSE $0.0085$
saturating vs.\ $0.0097$ log-linear on $k{=}1..16$), and the losing log-linear form was
already strained inside that range. The observed per-week gain falls $4.9\times$ between the
$k{=}1$--$8$ and $k{=}8$--$16$ segments, while $a + b\ln k$ can fall at most $3.4\times$
across those same segments, whatever its coefficients. The fitted saturating form was also strained. Its
24-week prediction already sat below the observed 16-week level it was fit to
(Table~\ref{tab:prereg}). Commitment (ii) therefore tests whether the better in-sample form extrapolates. It does not
test which of two competing hypotheses is true, so we give it less weight than (iii).
(iii)~\emph{Mixture tail}: before any 32-week job had run, we committed the
mixture hypothesis's prediction that the pooled weekly curve would gain $+0.0142$ $R^2$ from
week~24 to~32 (interval $[0.0126, 0.0231]$). The interval's lower end is the calibrated
bootstrap's 5th percentile, and its upper end is the raw mixture's 95th percentile. Calibration
scales the mixture's gain by the ratio of observed to mixture-predicted gain on the
$16{\to}24$-week segment. The committed falsifier, a gain of $<0.007$, would reject
the mixture in favor of a flat saturated tail.
(iv)~\emph{Deep-model bend}: before any matched-cohort number existed for the
32-week transformer, we fit the same two forms to its matched-cohort window curve. The registration discloses that the 32-week transformer's composition-confounded $R^2$ on the
full test set had already been computed. For the within-cohort gain from week~24 to~32, the two forms
predicted $+0.012$ (saturating) and $+0.041$ (log-linear), and we committed that a gain below
their midpoint would count as following the saturating prediction. The observed gain, $+0.011$, lands on
the saturating form. It comes from a single seed and carries the training-extract mismatch of
\S\ref{sec:miss}: the 24- and 32-week checkpoints were trained on different extracts.
On the full test set, the gain is $+0.024$, so composition inflates it two-fold.
In the M8$\to$M16 case, composition cut the other way and \emph{masked} the gain. An unmatched
test set can overstate a window gain or hide it, and matched cohorts remove the effect of test-set composition in
whichever direction it biases the gain.

Table~\ref{tab:prereg} scores all four outcomes against their pre-committed thresholds, and
\S\ref{sec:matched}, \S\ref{sec:saturation} and \S\ref{sec:miss} report the outcomes. The data falsified the prediction of commitment (iii).
The registration artifacts for the four commitments are content-addressed. Their SHA-256
digests, shown as their first 16 hexadecimal digits, are
\texttt{504e309ee7896795} (i, 2026-07-11),
\texttt{13cb1b424dcbc911} and \texttt{1992d7c8fbc0d9e4} (ii, two files, \mbox{2026-07-13}),
\texttt{367b1328d37ee12b} (iii, 2026-07-15), and \texttt{b8bf45d1fa5bac72} (iv, 2026-07-17).
The registrations are \emph{internal and self-attested}, kept with their full digests in the
experiment records. None was deposited with an external registry or third-party timestamped.
The printed digests let a record be checked as unchanged if it is later disclosed, though not as
written before its runs; readers cannot check the records now. The disclosure column of
Table~\ref{tab:prereg} states what was already known at each commit time.

Beyond the four content-addressed
commitments scored in Table~\ref{tab:prereg}, we fixed further thresholds before
their deciding runs and score each where it is used. They include the ridge-derived effect of
\S\ref{sec:external} with its power threshold and the anchor-equivalence criterion of \S\ref{sec:intro}.
Our prediction for the population term was partially met. This term is the change in log-space
$R^2$ from training on the 10M tier instead of the 1M tier. Before the cross-evaluation of
\S\ref{sec:limitations}, we recorded that scoring both tiers' models on the same test rows would
move the term from the $-0.005$ implied by scoring each model on its own extract into a band of
$-0.001$ to $-0.002$. The epoch-matched comparison, which trains both tiers for the same number of
epochs, lands inside the band at $-0.0016$. The budget-matched comparison, which holds the
optimizer-step budget fixed, misses at $-0.0026$. The pooled estimate over all six seed pairs,
$-0.00211$, sits $0.00011$ below the lower end of the band. The band itself spans only about $1.4$
standard errors of that pooled six-pair contrast.

\section{Discussion}
\label{sec:discussion}

\paragraph{What to do with the lever.} Extending the observation window produces the largest improvement in log-space $R^2$ of any intervention we measured for this task family, roughly ten times the best architecture intervention. Log-space $R^2$ is the one statistic in which every axis was scored. The ten-times figure excludes time representation, which we count separately (\S\ref{sec:time}). The window's gain of $+0.070$--$0.107$ is ${\sim}3$--$5\times$ the transformer's MS-RoPE gain of $+0.021$ over our rotary encoding as implemented. The top of that range comes from a single-feature linear model with no positional encoding, so the transformer-only ratio sits at the low end. The window gain holds at every population scale we tested
and in every model family down to a single-feature linear model. The gain survives matched-cohort evaluation and, through 16 weeks, grows under it, so it is not an artifact of which customers are scored. With the Huber objective, most of the 8-to-16-week gain also survives a monotone remap of the predictions, so it is not mainly about where predictions sit on the scale. A monotone map fit on validation closes only $-0.73\%$ to $6.65\%$ of the gain, and even the best monotone map on the scored rows could account for at most ${\sim}30$--$32\%$ (\S\ref{sec:objcal}). The gain is bounded on the pooled weekly-aggregate probe, which saturates at 22--24 weeks on the test calendar, a bend whose form we predicted before measuring it. We then stress-tested that form with a 32-week extension. The probe stayed flat over the added weeks, which falsified our own mixture extrapolation. Window length and calendar coverage remain unseparated on the test anchors (\S\ref{sec:limitations}). A check of the probe on training-period anchors separates them in part: most of its 8-to-16-week gain persists outside the holiday season, and its returns still diminish there, but the sharp bend is not shown to persist (\S\ref{app:calendar}), and none of the public datasets of \S\ref{app:public-ladder} on which we assess the bend shows it. Every window contrast we report across extracts is ``window as extracted'': a longer window together with the modestly larger, colder training population it admits. On the public datasets of \S\ref{app:public-ladder}, rows are held fixed across windows instead.

A team building on this class of data should do three things, in order. First, lengthen the window to at least the probe's saturation point, 22--24 weeks on the test calendar; returns diminish on both calendars we checked (\S\ref{app:calendar}). Second, match the training objective to the evaluation statistic. Third, adopt a physical-time positional encoding only after it is \emph{scored in the statistic you will report}. That encoding's log-space gain reverses sign under a recipe that trains and scores on the raw, un-logged target (\S\ref{sec:external}).
Expect more customers, more parameters, and deeper stacks to buy very little when each is changed alone inside the fixed d128, 2-layer recipe. Under the same conditions, expect the model trained on the larger population tier to score slightly below the one trained on the smaller tier once both are evaluated on the same rows. That contrast also swaps which customers each model trains on, so it does not isolate customer count (\S\ref{sec:limitations}).

\paragraph{Allocation, within its window.} A fixed short window starves the customers who need the most feature weeks. In the per-decile decomposition, the top activity decile reaches half its fitted asymptotic gain in about four weeks and the next two heavy deciles in five to ten weeks, while light users keep gaining through at least 16 weeks. Light users, the least active, also carry the fewest events inside a short window. The decomposition thus turns one pooled law into an operational rule that allocates feature weeks inversely to activity. The 32-week test bounds that rule. Past 24 weeks the residual gains are small in every decile, so the rule applies to the first few months only.

\paragraph{What the falsification bought.} The claims that survive are stronger for the failed prediction. The failure matters because our own per-decile decomposition inside the window motivated the mixture model. We committed the model's quantitative extrapolation in advance, and its own falsifier rejected it.
The four internally registered commitments scored in
Table~\ref{tab:prereg} are not equal in what they could have shown. One was
fully blind and failed on its own registered falsifier. The three that held are weaker instruments, each for a different reason. Commitment~(i) tests a threshold and does not compare forms, so it can only fail low. Commitment~(ii) separates two forms that were nearly tied on the data they were fit to, and the losing one was already strained by the curvature inside the fit range. In effect, this commitment tests whether the better in-sample form extrapolates (\S\ref{sec:prereg}). Commitment~(iv) is scored as
committed on a single-seed measurement that carries the training-extract mismatch of
\S\ref{sec:miss}. Graded that way, the set is still harder to attribute to analyst degrees of freedom than any single test, and harder still because the table discloses what was known at each commit time. The registrations, however, are internal and self-attested, with no third-party timestamp (\S\ref{sec:prereg}). The discriminating weight sits on the blind registration.

The saturation result extends past the linear probe, though not
independently of it. Before any matched-cohort number existed for the 32-week transformer, we committed that its matched within-cohort gain from week 24 to 32 would follow the saturating form ($+0.012$) rather than the log-linear one ($+0.041$). At that point we already knew the model's full-set $R^2$, which is confounded by a change in which customers are scored. The observed gain of $+0.011$ lands on the saturating prediction. That gain comes from a single seed (customer-clustered bootstrap 95\% CI $[0.010, 0.013]$). The binding uncertainty is the training-extract mismatch: the two checkpoints were trained on different extracts. An offset of the size measured on the ridge sub-ladder would put the observation near the log-linear prediction (\S\ref{sec:miss}, Table~\ref{tab:prereg}). We have not measured that offset on the transformer. Commitment~(iii)'s pooled probe is fit within a single extract and carries no such mismatch (\S\ref{sec:miss}). Allowing for the transformer's mismatch, the two instruments are consistent with the same saturating law at different points on it, and they do not amount to two independent rejections of the alternative. On the test calendar the probe is flat past 24 weeks, while the deep model still gains $+0.011$ from week 24 to 32 and decelerates onto the fitted curve. Beyond 32 weeks, that curve leaves about $0.007$ of headroom on seed 42, on par with the three bounded axes (scale, capacity and architecture). The headroom figure is a point estimate from the fit rather than a measured bound (\S\ref{sec:miss}).

\section{Limitations}
\label{sec:limitations}

\emph{Window length and calendar coverage are not separated on the test anchors.} The most important missing control
concerns the lever itself. All test target weeks fall in one ${\approx}8$-week band:
three anchor weeks, 2025-02-24, 2025-03-24 and 2025-04-21 (\S\ref{sec:data}). The
input window is anchored at the target week, and the target is a 364-day forward
aggregate. Extending $M$ therefore adds specific calendar months, not generic ``more
history''. The 16-week window first reaches the prior November at the earliest of
the three anchors; the latest anchor reaches it only at 24 weeks. The 24- and 32-week
windows reach into the preceding autumn and summer. One seasonal mechanism is thus a
candidate explanation for both main features of the window curve: the size of the M8$\to$M16 step, and the
pooled plateau at 22--24 weeks on the test calendar, where the last of the three anchors finishes covering the
holiday season. On the test anchors themselves, none of the analyses in this paper distinguishes that seasonal mechanism from a pure history effect. The matched
cohorts, the per-decile decomposition, and the objective, family and scale sweeps share
those same three anchor weeks, and the Taobao dataset spans nine days in total. Of the four public datasets spanning more than a year (\S\ref{app:public-ladder}), Online Retail II's anchors run from May to December, off the test calendar, and the weeks its 16-week window adds hold no day from November to January, so its 8-to-16-week gain does not come from added holiday weeks; The Complete Journey carries no calendar dates.
On CDNOW, whose one anchor is 1997-06-23, the weeks the 16-week window adds (1997-03-03 to 04-27) hold no November-to-January day, though that window also holds the join date for $31.1\%$ of rows. On H\&M, to take the holiday season out of the added weeks, a follow-up refit uses the five anchors from 2019-06-03 to 2019-09-23, whose whole 16-week history holds no November, December or January day; it is reported beside H\&M's pooled gain, with its COVID caveat, in \S\ref{app:public-ladder}.
Away from the test anchors, the weekly-aggregate probe partly distinguishes the holiday-season mechanism from a history effect (\S\ref{app:calendar}). We refit it on training-period anchors, fit on 80\% of the training customers and scored on the other 20\%, so it is a validation-style check of the linear probe, not of the trained models or of the test numbers. Its 8-to-16-week gain on holiday-free anchors is $0.72$ (customer-clustered 95\% CI $[0.69, 0.75]$, covering score-side sampling only) of its gain on training anchors that share the test calendar. The decision rule, fixed before the check computed any $R^2$, classifies that ratio as ``persists outside the holiday season.'' The sharp 22--24-week bend persists on the test-calendar anchors, scored on all their held-out customers, and is not shown to persist on the holiday-free ones, though returns diminish on both. The two anchor groups differ in season as well as holidays (summer against spring), so the check does not attribute the gap between their gains to the holidays alone.
Separating window length from calendar coverage on the test anchors needs a control with anchor variation, which
end-anchoring does not provide. On the Taobao dataset we removed an analogous confound by
re-extracting the data with end-anchored windows, which fixes the target period across windows (\S\ref{sec:external}). The
primary ladder is already end-anchored, with a fixed target week and a fixed 364-day horizon at
every $M$. That is precisely why window length and calendar coverage coincide here. The needed control is a calendar-balanced anchor set
spread across the year, so that the ladder averages over calendar alignments. For the trained models such a set
cannot be recovered from this export after the fact, because each row carries the target for its
own anchor only, and the training anchors the probe check reuses are ones those models trained on. That control needs a new extraction and a re-run of the ladder.
This export \emph{does} admit the anchor-\emph{varying} control of \S\ref{sec:intro}, which holds
the added block at eight weeks while the calendar months it covers vary
across the three test anchors. We ran it at two objectives and three seeds, six
settings in all, and it was \emph{void under its own internally registered
anchor-equivalence criterion in all six}. It neither licenses a calendar
attribution nor clears the confound. Two
diagnostics say why. First, the anchors already differ in $R^2$ before the added block
enters. At M8 their $R^2$ spread is $3.7$--$6.2\times$ the bound that criterion
allows. Second, in every setting the anchors' order by baseline difficulty matches their order
by calendar position. Headroom, the $R^2$ each anchor has left to gain, therefore remains a
possible explanation of that ordering everywhere. The obstruction holds at both objectives, so it is not specific to
one ladder.
We do not identify the window gain's \emph{magnitude} as a calendar-independent history effect (\S\ref{sec:intro}).
The gain's direction is less exposed to this confound, though not equally on both objectives. Within the
anchor-varying control, every anchor's 8-to-16-week step is still positive on the
MSE objective. On Huber, in the same control, the latest anchor's added eight weeks are worth
approximately nothing, positive at two seeds and negative at one (\S\ref{sec:intro}).

\emph{A train-side availability boundary, measured and bounded.} A second calendar artifact sits on the input side of the training split. We measured it on the weekly-aggregate probe, where it does not explain the plateau. In the
24-week 1M extract, event history is only partly available before calendar week 2023-10-30. Weeks starting
on or before 2023-10-23 carry $1.5$--$2.0$ events per row against $4.0$--$7.3$ in later weeks; the step
falls between two adjacent weeks, and within a calendar week the density is the same to
${\approx}1\%$ at any lag. The pattern reflects data availability, not recency.
The lag-$w$ column ($w{=}0$ is the week nearest the anchor, \S\ref{app:probe}) falls on a different calendar week for each training anchor, so the three earliest
training anchors (2024-01-29, 2024-02-26, 2024-03-25; \S\ref{sec:data}) have only partial history
at or beyond lags 13, 17 and 21, respectively. Pooled lag columns are therefore drawn from a
truncated population for $11\%$, $22\%$ and $33\%$ of training rows at lags 13--16, 17--20 and
21--23. Refitting the weekly-aggregate probe on the six unaffected anchors alone raises the 8-, 16- and 24-week
values by $+0.0013$, $+0.0025$ and $+0.0013$, and moves the
16$\to$24-week gain in the direction opposite to what a truncation account of the plateau predicts
($+0.0150 \to +0.0137$). The 8-week value moves by $+0.0013$ even though
no column is truncated there. The refit's shifts thus mostly reflect the smaller training set (nine anchors to six,
$5.04$M rows to $3.36$M), not the truncation.
In the 32-week extract the same boundary falls inside the eight weeks that the 24-to-32-week extension adds, for five of the nine
training anchors: $55.6\%$ of 32-week training rows have at least one added week before it, and $47.2\%$ of the added
row-weeks fall before it. All 32 weeks before every test anchor fall after it. Refitting the weekly-aggregate probe on only
the four training anchors whose 32 weeks all fall on or after the boundary gives a smaller 24-to-32-week gain than
refitting it on a random subset of the same size that keeps each of the nine anchors' share: the difference is $-0.0031$
(log-space $R^2$; 95\% customer bootstrap interval $-0.0038$ to $-0.0022$). Those four anchors are also the four latest,
June to September 2024, so truncation and season cannot be separated here. These numbers are measured on the weekly-aggregate linear probe at the 1M tier. The same refit of the transformer,
with three seeds, finds that dropping the five earlier training anchors does not move the 24-to-32-week gain beyond
the size effect: the mean over seeds of the gain trained on the four anchors minus the gain on the random subset is
$+0.037$ (log-space $R^2$; 95\% $t$ interval $-0.022$ to $+0.095$, 2 degrees of freedom). The transformer refit trains
and scores on corrected inputs (real category codes and standardized price and quantity), so it does not re-score the
paper's 24-to-32-week gain, which stands as reported, and it keeps the same four latest anchors, so it cannot separate
truncation from season either. The transformer reads the store field, which was recoded at its source on 2024-12-06:
training inputs and the weeks the 32-week window adds at test carry the old coding and the later part of each test
window the new one, so part of each 24-to-32-week gain can be a coding match rather than history. The check compares
two such gains on the same test inputs, which cancels the shared part but not any difference in how the two models use
the older weeks.

\emph{Each bound is an interval that leaves its effect unresolved.} The $\leq 0.007$ band we quote for the scale and capacity
axes is one such interval. The architecture band is another: the $\leq 0.0052$ spread across five
families at 16 weeks and seed 42 on the full test split, with the clocks as built and the transformer inside it. We pooled three-seed spreads over the
seven fixed-configuration 16-week 1M settings, the one class of full-set settings we could pool. These seven remain after dropping the
two settings with visibly wide spreads (d512 with 8 layers, and the unconstrained control), and they are homogeneous
(Bartlett $p = 0.15$, Levene $p = 0.14$). Together they give
$\hat\sigma = 0.00128$ on $14$ degrees of freedom. That is a sample SD over seeds, which is a different
quantity from the detection floor. For the same class, the detection floor at $n{=}3$ is $0.0059$: the smallest difference that a paired three-seed comparison detects
with $80\%$ power at a two-sided $5\%$ level. On the point estimate, the
$\leq 0.007$ band is therefore about $1.2$ times that floor. However,
$0.007$ sits inside the floor's own $90\%$ interval ($0.0045$--$0.0086$), so for this class the band is
about right, and we cannot show that it is conservative.
That floor does \emph{not} transfer to the rest of the grid. Across fourteen three-seed configurations the seed
spreads are heterogeneous by a factor of $34$ (Bartlett $\chi^2 = 35.7$, $p = 6.7\times10^{-4}$),
so our grid has no single detection floor and we quote one only for the class it
was measured on. One consequence is that the capacity probe's $-0.006$ is
\emph{unresolved} at $n{=}3$. Its own equivalence margin, the farthest its paired 95\% interval reaches from zero, is $0.016$.
That margin is wider than the floor, so that configuration cannot support a bound in either direction.
Checkpoint selection adds a second effect, comparable in size to all three bounds and separate from
seed variation. For the transformer ladder runs, we chose the epoch behind each number in a metric
space we do not report. Re-selecting in the reported space moves individual results by amounts whose top end
exceeds any of these bands. The shift lands differently on each axis, so we take the axes in turn
below. On scale it moves both sides of the contrast almost equally. On capacity the shift differs between the two
sides by about the width of the bound itself. On the architecture band the test split gives a sign
but no bound. Taken together with the floor above, all three axes
are bounded but unresolved, and none of them is a null.

\emph{The architecture band compares families that did not share a learning-rate schedule.}
The $\leq 0.0052$ seed-42 architecture band spans five families, the transformer
included. The transformer was optimized differently from the other four, and those four did not share a peak learning rate among themselves (Table~\ref{tab:recipe}). The transformer runs trained at an effectively constant
learning rate: a cosine decay was configured, but its period exceeded the
realized optimizer-step count by about ten times, so the schedule ended within $3\%$ of peak
(\S\ref{sec:repro}). The recurrent and continuous-time runs set their cosine period to the
epoch count and step once per epoch, so they annealed in full, to roughly $2\%$ of peak.
Annealing generally helps a final metric, so the asymmetry handicaps the transformer relative to
the other families in the band. That pulls the families together, which is the
direction that would create this band spuriously rather than hide a real one. In the other direction, the
transformer trains for twice as many epochs, so argument alone cannot settle the net sign. We report
the band as measured under the schedules that ran, and do not claim it would survive a matched
one.
Extending the four baseline families to three seeds breaks this bound. Restricted to those four,
the three-seed spread is $0.0059$, wider than the
$\leq 0.0052$ band (both on the full test split), with GRU and ContiFormer as its extreme pair. Two things follow. First, the schedule argument just given does not cover
that pair: both annealed per epoch and both were scored at the last epoch, so the
asymmetry that would pull the transformer toward the baselines is absent exactly where
the bound fails. That pair did not share a peak learning rate, though: GRU trained at $3{\times}10^{-4}$ and ContiFormer at $1{\times}10^{-4}$ (Table~\ref{tab:recipe}). Second, the exceedance is not an
artifact of which models we included. The transformer's three-seed mean falls inside the
four baselines' range, so the spread is the same $0.0059$ whether or not the transformer is
counted; the extreme pair is two baselines either way. Two caveats remain. One is that learning-rate gap, which keeps us from attributing the exceedance to family; the other is that the transformer's
three-seed runs trained on a different build from the four baselines. That build mismatch,
measured at seed 42, is below citation precision on the baseline path, so it does not explain the exceedance
there.
We report the band at seed 42, where it was set; it does not hold at three seeds. Either way it is a measured
interval, not one of the four registered commitments of \S\ref{sec:prereg}.

\emph{The population-scale comparison is not schedule-matched either.} The same cosine period is used
at every population scale, but the number of optimizer steps is not: it grows with the training
set. At 1M, training covers a tenth of the period. At 10M it covers the whole period and
overruns it, and the 10M runs end at a learning rate far below the
1M runs' near-peak rate. The period, the realized step counts, and both end-of-run rates
are in \S\ref{sec:repro}. The 10M runs are thus annealed while the 1M runs are barely annealed
at all, and the $\leq 0.006$ scale bound compares two different optimization regimes rather
than a change in one factor. The larger population holds the schedule advantage, so if completing the decay helps a
final metric, a matched comparison
would if anything move 10M down relative to 1M. That would make the bound conservative for the claim we
make with it, that scale buys little.
The schedule asymmetry is real, but removing it leaves the scale delta where it was. A follow-up set of twelve runs on one build retrains the d128 16-week configurations at 1M and 10M in two blocks
of three seeds, one epoch-matched and one holding the optimizer-step budget fixed. Budget-matched, the 1M$\to$10M delta in log-space $R^2$ on the
held-out test split is $-0.0053$, against
$-0.0045$ epoch-matched on the same build and $-0.0053$ for the earlier confounded pair at the
same window and capacity on the corrected schedule. Against that confounded pair, matching the budget moves the delta by
under a percent, and nominally toward a slightly larger gap rather than a smaller one. Within
the follow-up the two blocks differ by $0.0009$, or
$0.68$ standard errors on seed-paired differences over the same three seeds, and both are
negative at every seed. The follow-up is its own recipe family, and we do not pool it with the ladder
the bound is quoted from, so it tests the confound rather than the bound. Within the follow-up
the largest delta is $0.0053$, inside the $\leq 0.007$ band we quote for scale on three-seed means. Its
deltas are small and consistently negative rather than flat.
Much of the delta is also not an effect of
population size. When we reproduce each run's own checkpoint selection, the validation-to-test drop is
larger on the 10M extract than on the 1M extract by $+0.0039$, reproduced to a population SD of
$2.7{\times}10^{-5}$ across three recipes spanning $15\times$ in relative step budget. Those
three select their checkpoint on the log-space validation series; the ladder the bound is quoted
from selects on original-scale validation error and carries an offset of the same sign and
order, $+0.0029$ at 16 weeks. Pooling all four gives $+0.0037$ at a population SD of
$4.1{\times}10^{-4}$, $15\times$ looser, and we quote the three-recipe pair rather than the
pooled one. Subtracting each recipe's own offset leaves residuals of $-0.0014$, $-0.0006$ and
$-0.0014$; their mean across the three recipes is $-0.0012$, with a $t$ interval of
$[-0.0024, +0.0001]$ that spans zero. We keep the test-space figure as the reported
estimand, because selection ran on the validation split and leaves its scores optimistically biased. The decomposition diagnoses that figure without replacing it, and the diagnosis points to the extracts themselves. The tiers are customer-disjoint extracts, each model is scored on its
own extract's test split, and a difference in how each extract's validation and test splits relate to each other enters
every cross-tier comparison.
With the evaluation set held fixed, so that both models are scored on the same rows, the scale
delta is a third to a half of its size across extracts (as point estimates), and it is negative at all
six seed pairs. We hold the evaluation set fixed in both blocks by scoring the 1M-trained checkpoints on
the 10M extract. On that shared set the 10M-trained minus 1M-trained delta in log-space $R^2$
is $-0.0026$ budget-matched and $-0.0016$ epoch-matched, against $-0.0053$ and $-0.0045$
measured across extracts. Neither block alone resolves the magnitude: each
three-seed interval contains both zero and that block's value across extracts. Pooling the six pairs
does. This pools one manipulation across two recipes, not two levels, and gives
the interval $[-0.0039, -0.0003]$, which excludes both zero and $-0.0053$. That
interval counts the six pairs as six units, and the two recipes reuse the same three training
seeds; taking the seed as the unit leaves three and gives $[-0.0055, +0.0013]$, which
excludes neither. We quote the six-pair interval, which the pooling above licenses; with the seed as the unit, the
magnitude is \emph{unresolved} at $n{=}3$. Neither the center nor the sign moves with that
choice: the design is balanced, so both poolings give $-0.0021$, and the six negative
pairs above are three negative seeds. On the six-pair interval, the population term is negative, at roughly $-0.002$,
and is not the figure measured across extracts. That figure, $-0.0053$, should not
be read as a population effect. In support of this,
the extract mismatch itself contributes $-0.0029$ under one recipe and $-0.0027$ under
the other. The two schedules' scored checkpoints sit $15\times$ apart in optimizer-step
ratio, yet these contributions agree to $1.3{\times}10^{-4}$. The population residuals the two schedules leave behind
do not agree as closely: $-0.0016$ epoch-matched against $-0.0026$ budget-matched, a difference
three seeds do not resolve. A property of the extracts should not move with the schedule, and a
training effect should.
Two of the cross-evaluation's limits are structural rather than a matter of seed count. The contrast is between
training \emph{cohorts}, not training-set sizes: the tiers are disjoint bucket ranges that share
no customers, so cohort identity changes with cohort size. Only a nested subsample of the
larger cohort would isolate size, and that contrast is not in this paper. The second limit is that the interval above
is over training seeds, with the evaluation rows held fixed. For the budget-matched block, a customer-clustered paired
bootstrap over the evaluation rows gives an interval of
$[-0.00268, -0.00253]$ on that block's $-0.0026$. The two
intervals reflect different sources of uncertainty over different units, and the row-level one is far
tighter: its standard error is about nineteen times smaller than the
budget-matched block's seed-level standard error. We do not treat the tighter interval as resolving this contrast.
The sign runs against what a scaling account predicts. On
the 10M cohort's own test split, the model trained on the 1M cohort scores higher than the model
trained on the 10M cohort at every seed, though it never saw a customer in that cohort. It does
so despite a disadvantage on both counts that could matter. The 10M-cohort model is scored on its own training cohort, and its checkpoint was chosen on
that cohort's validation split. The 1M-cohort model was trained and selected on a different
cohort. Whatever a train-test cohort match is worth therefore goes to the model that loses. At three seeds we claim the sign and not a magnitude.
The schedule asymmetry does \emph{not} affect the window lever, the capacity configurations, or the tabular baselines. The window lever is measured within
a scale, and window length does not change the number of training rows. The 8- and 16-week
runs at 1M take the same number of steps and end at the same learning rate to four significant
figures, so the window deltas are schedule-matched at 1M, where we measured the step counts. The capacity configurations within the
transformer family are also schedule-matched, apart from the capacity probe's longer warmup: they share the effective batch size and match the realized step count to within ten optimizer steps. The tabular baselines
have no schedule at all.

\emph{Checkpoint selection does not use the reported metric.} For the transformer ladder runs, we chose the epoch behind each number by minimizing validation error on the
\emph{original} scale. Every primary $R^2$ we report is in log space precisely because a few extreme customers dominate the original scale. The log-space validation series is recorded at every epoch
but was not the selection criterion. Re-selecting on that recorded
log-space series moves the reported statistic by $+0.002$ to $+0.009$ depending on the configuration. We
re-evaluated one of those configurations end to end against the full test split to check that this is not
an artifact of the validation data: the two checkpoints differ there by $+0.0022$, tracking the
validation-side estimate for that configuration. Two consequences follow. First, the effect does not cancel out of the
contrasts, because it is not equal across configurations: the base model and the widest capacity variant differ by
about the width of the capacity bound itself, which makes that bound the most exposed of the
contrasts for which we can measure the offset at all. Second, we
cannot correct the effect without retraining, because the widest variant's log-best epoch was not among
the checkpoints retained. We therefore report the capacity bound as measured under the selection
rule that ran, and interpret it as including selection sensitivity rather than as a bound on
capacity alone.
We also retrained both models in the capacity contrast with log-space selection and the corrected schedule together
(a retrain separate from the five-configuration retune of \S\ref{sec:ceiling}),
so the two changes cannot be separated in that retrain. The contrast
comes out at $+0.0004$ with three seeds per model. The magnitude bound is unchanged, and the sign does
not reproduce. Two other quantities inherit the same
rule, in opposite ways. On the scale contrast the offset is nearly the same on both sides at seed 42, the seed of the scale bound. The 8-week 1M
model and its 10M counterpart move on validation by $+0.0043$ and $+0.0041$, so the $\leq 0.006$ bound
is close to selection-invariant for the one such pair on which we measured the offset.
The cross-family band has no such protection. The recurrent and
continuous-time models select no epoch and keep their final one, and the tabular models
have no epoch to select. The $\leq 0.0052$ seed-42 architecture band and the transformer's margins
over GRU and LSTM (\S\ref{sec:baselines}) therefore compare a validation-selected transformer against
unselected comparators. The one \emph{metric-space} re-selection delta we measured on the full test split is positive, and
the transformer is at the top of that band, so correcting the rule would widen it and enlarge those
margins. The correction would move this axis away from a null, not toward it.
The attainable counterfactual strips selection from the transformer rather than granting it to
the comparators, since the baseline cancels from the
difference of the two transformer scores. We computed this counterfactual on the validation series for all six
plain-RoPE M16-1M runs (Huber and MSE, three seeds each). The selection rule as used is worth at most $+0.00034$ to the transformer, an order of
magnitude below the three-seed detection floor and also below the architecture band. Its value is \emph{negative} on three of the six runs, where the rule cost the transformer up to
$0.0022$. That bound on the rule as used is measured on the validation series, not the test split, so we
still interpret the architecture band and those margins as including
selection sensitivity.

\emph{How much of the window gain comes from zero-target rows.} A substantial part of the pooled window gain sits on the zero-target rows. Those rows are $4.8\%$ of the 8-week evaluation set and
$8.7\%$ of the 16-week one, but carry over half of the sum of squares the log-space $R^2$ is
computed over. An exact split of the 8-to-16-week gain into zero-row and
positive-row components assigns $29$--$30\%$ to the zero rows under MSE and $51$--$53\%$ under
Huber (three seeds each). The share is strongly objective-dependent, so ``the window lever'' is not one
quantity measured two ways. Because the zero fraction nearly doubles between the two
evaluation sets, the pooled gain mixes any zero-gate improvement (\S\ref{app:objective}) with a composition shift
that this decomposition cannot separate.
The matched cohort would separate them, since it holds the zero fraction fixed across windows
(\S\ref{sec:matched}). How much of the zero-row share reflects better
prediction of \emph{whether} a customer transacts at all remains open.

\emph{Every one of these bounds is a log-space bound.} We compute each primary $R^2$ in one
space (\S\ref{sec:eval}), and we derived the bounds on all three axes (scale, capacity and architecture) in that
space only. Our own results warn against treating them as invariant to the scoring space: the log-space gain of one model-side
intervention, the MS-RoPE physical-time positional encoding, reverses
sign when we retrain it under a raw-target recipe and score it in that recipe's winsorized statistic
(\S\ref{sec:external}). That reversal is where our target-space discipline comes from. Under
that discipline, those bounds and the ordering the title asserts are log-space statements. None
of the three axes has had its bound re-derived in the statistic a practitioner would report.

\emph{The lever comparison is not cost-normalized, with one exception that runs in our
favor.} The interventions we compare do not cost the same: 1M$\to$10M is ten times the
transformer's training data, and the capacity probe is ${\approx}54\times$ the parameters
($491{,}248 \to 26{,}479{,}216$, Table~\ref{tab:params}). The window is the exception: for the transformer, the dominant lever in this paper costs no
additional training compute, so it is the one lever for which the cost-normalized
comparison and the raw one coincide. Extending the window from 8 to 16 weeks changes how much
history a sequence carries, not the length of the tensor that carries it. Every per-event
feature is allocated at the fixed $750$-event cap of \S\ref{sec:repro} and zero-padded, so
the model attends over $750$ positions at every window and per-step FLOPs are identical across
$M \in \{8,12,16,24\}$. The number of optimizer steps per epoch at 1M is $488.3 \pm 0.2$ across
twenty-four trained checkpoints, independent of window, width, depth and micro-batch size. An iso-FLOP
comparison between history and capacity is therefore ill-posed, since one of its two sides
does not consume the budget.
We ran no configuration at matched cost across the
remaining axes, so this design cannot rank scale against capacity per unit of compute. The
title's claim that longer windows outweigh the observed effects of the other axes, like the ordering in \S\ref{sec:discussion}, is a statement about effect size at
the recipes tested, not about budget-optimal allocation.

By design, the primary evidence comes from one platform, one target and one export era. Restricting the evidence to a single
era keeps the within-paper deltas free of cross-era differences, and it also limits the scope
of the claims. The target horizon is fixed at 364 days, so how the window gain
depends on it is untested in this export era. Within that platform the 1M/5M/10M tiers are customer-disjoint
cohorts (\S\ref{sec:data}), so findings that agree across tiers replicate across
non-overlapping customer cohorts: internal replication, not an independent
data source. The external replication on Taobao recovers the direction of the gain in linear and deep models and its shape in ridge, over a far shorter horizon,
and its intervention is windows as extracted. Longer end-anchored windows also enlarge
the eligible training set ($610$k${\to}686$k rows), so it does not separate more history
from more training rows. Its internally registered deep-model
comparison is positive at all three seeds but was underpowered for its
registered effect size. Absolute levels on that
public dataset are near zero. On two public multi-year retail datasets, Online Retail II and The Complete Journey, the weekly-aggregate ridge probe returns the verdict ``longer history helps here too'' on both: it gains from 8 to 16 weeks, a control that permutes weeks 8--15 across customers within each anchor gains nothing, and returns diminish, but the bend is not shown (\S\ref{app:public-ladder}).
We also run the same probe on two more public datasets that each span more than a year, CDNOW and H\&M; their results are reported with their intervals and caveats in \S\ref{app:public-ladder}. Among those caveats, CDNOW's longer windows contain most rows' join date; on H\&M, a customer's first line in the file can fall inside the longer windows, and seven of the eight target years run into the COVID period (from 2020-03-11, by 26 to 194 days). These checks fit ridge on weekly aggregates and say nothing about scale or about the size of the effects on the main dataset. Unlike the Taobao replication, they hold rows fixed across windows, so their gains do not come from more training rows. The paper's transformer was also trained and scored out of fold on Online Retail II, The Complete Journey and CDNOW, on the ridge probes' own rows and folds, and on each of them, read on its own, it gains from 16 weeks of history over 8 (\S\ref{app:public-ladder}). None is out-of-time validation: on three of the four, earlier anchors' targets overlap later anchors' input weeks, and on CDNOW every row shares one anchor date, with folds by customer. A weekly-aggregate linear probe measures the saturation and
allocation results; an internally registered matched-transformer test at 32 weeks (single
seed) is consistent with the saturation result. Several grid entries are seed-42 point
estimates, and a few cited entries were computed outside the training pipeline; each is flagged where cited. The matched-cohort framework fixes test
composition but conditions on membership in the enumerated intersection of the windows' test splits,
which is neither the full platform population nor every row a longer window scores. At 1M every row of the three- and four-window matched cohorts has an event in the 8 weeks before its target week, so matched gains there are gains for customers active in those weeks (\S\ref{sec:matched-method}). The claims on the three bounded axes are one-factor-at-a-time results, apart from the tier contrast's realized schedule, inside the fixed
d128, 2-layer recipe. Capacity is probed only at the 1M tier (3-seed at 16 weeks,
single-seed at 8); the reported capacity configurations inherit the small model's
hyperparameters, and a five-configuration retune of the widest variant did not resolve that
configuration in either direction (\S\ref{sec:ceiling}); a five-point learning-rate sweep on corrected inputs
leaves the capacity bound standing, and on the same inputs no interaction between the 8-to-16-week window and model
size is shown (\S\ref{sec:repro}). The joint capacity$\times$data setting of
compute-optimal scaling is untested, so these bounds are claims about this operating
regime, not about jointly scaled models.
Finally, a fixed input cap, which keeps the most recent 750 events at every window, attenuates event-weighted estimates of the longest-window
gains: on
the matched cohort, $3.0\%$ of customers had already hit the cap at M24 and so received no new
input from the longer window. The
saturation verdict does not rest on capped inputs. The ridge instrument behind that
verdict is uncapped, and a direct exclusion test shows the gain is not carried by the cap-hit customers:
removing the cap-hit rows leaves the matched gain unchanged ($+0.0115$,
customer-clustered 95\% CI $[0.0101, 0.0130]$, vs.\ $+0.0112$ on the full matched cohort). That test does not bound the cap's effect; an
internally registered cap-raise intervention bounds it directly. Retraining the
M24 and M32 models with the cap doubled to $1{,}500$ events shifts the matched
$\Delta(24{\to}32)$ by $+0.0020$ (customer-clustered 95\% CI for the shift
$[0.0013, 0.0026]$), inside its $\pm0.004$ prediction band. It leaves residual
cap-hit fractions of $0.48\%$ and $0.99\%$ on the matched cohort at M24 and M32 (single seed; the
cap-1500 pair trains at a smaller effective batch than the recorded 750-event runs it is compared against, so its shift bounds the cap and batch changes
together). We report per-window
cap-hit fractions in \S\ref{sec:repro}.

\paragraph{Broader impacts.} This work studies aggregate behavioral forecasting on
platform event data. Only population-level statistics leave the paper (see the
availability statement, \S\ref{sec:repro}). Our findings measure
whether observation history carries information, and for whom. The
allocation result cuts both ways. Extra observation history is worth most for low-activity
customers. That finding can guide better service for lightly observed users or, equally, more
effective targeting.
Two of our measured findings point toward using \emph{less} observation
history. First, the pooled weekly-aggregate probe is flat past 24 weeks on the test calendar. On this instrument and calendar, five to six
months is a measured ceiling. Out to the 32 weeks we measured, storing weekly history past that
ceiling buys nothing the pooled probe can detect. Within a single extract, the measured
$\Delta(24{\to}32)$ is $-0.0004$ (customer-clustered 95\% CI $[-0.0014, +0.0004]$). The
interval's upper bound sits below the internally registered $+0.007$ discriminator (\S\ref{sec:miss}). Second, history can be allocated unevenly
across customers. Routing feature weeks broadly inversely to activity (more to less active customers, fewer to more active ones)
retains about $99\%$ of the value of a fixed 24-week window at $49.2\%$ of its feature-week budget
(\S\ref{sec:allocation}).
That budget counts only the weekly aggregate feature vectors. To place each customer in an
activity decile, the allocation also uses that customer's activity total over the full window,
which the budget does not count (\S\ref{sec:allocation}). The saving is therefore in how many
weeks must be turned into features for each customer. Every customer is still observed over the
full window. A ceiling on how much history helps also does not by itself justify holding
less of it. The lever is observation history, and the allocation result says the marginal
week is worth most for the customers who are least observed.
Our
contribution is measurement discipline: what the lever is,
where its pooled fit saturates, and what this design can and cannot attribute it to.
We offer no deployment capability.
Deploying longer observation windows is a platform-governance decision, carrying the
same obligations as any other use of customer behavioral data.

\paragraph{Data handling.} We do not release, deposit, or reproduce the proprietary event corpora here.
Nothing individual-level leaves the paper: no raw events, no trained weights, and no
per-customer predictions (\S\ref{sec:repro}). Every
quantity we report is a population aggregate over a cohort, such as an $R^2$, a mean, a
clustered interval, or a fraction of a cohort. No reported number is attributable to
an individual customer. 
For the external check (\S\ref{sec:external}), we use the public Alibaba Taobao
UserBehavior corpus as released and unmodified, for research. For the public checks on datasets spanning more than a year (\S\ref{app:public-ladder}), we use Online Retail II from the UCI Machine Learning Repository~\citep{chen2012onlineretail,chen2012rfm}, dunnhumby's The Complete Journey~\citep{dunnhumby2023completejourney} and the CDNOW master file~\citep{fader2001cdnow}, obtained as released and used for research only. We use the transactions of the H\&M Personalized Fashion Recommendations competition~\citep{hm2022kaggle} under the competition's non-commercial research terms, and report only aggregate results from them. We redistribute no part of any of these public datasets.

\section{Conclusion}
\label{sec:conclusion}

The observation window is the one dominant input-side lever among those we measured for
customer-behavior forecasting on this platform. We rank the levers in log-space $R^2$, with test
cohorts and targets fixed and evaluation matched. The lever holds across eight trained model
families at 1M and across a $10\times$ span of population scale in every family we ran at all
three tiers. Controlling test composition makes its 8-to-16-week gain larger. Window length is
confounded with calendar coverage (\S\ref{sec:limitations}). The window gain's magnitude is
measured \emph{pooled} over three test anchors that share one eight-week band, and the
anchor-varying transformer control we ran returned void under its own criterion
(\S\ref{sec:limitations}). A check of the weekly-aggregate probe on training-period anchors partly separates window length from calendar coverage, since most of its 8-to-16-week gain persists outside the holiday season (\S\ref{app:calendar}). Separating them fully for the trained models needs a
calendar-balanced re-extraction.

On the pooled weekly-aggregate probe, a saturation law also bounds the lever. We
predicted its form and measured its bend at five to six months on the test calendar, and our prediction of further
gain past that point failed on our own internally registered falsifier (\S\ref{sec:prereg}).
The transformer still gains from week 24 to 32, at one seed and subject to the training-extract mismatch
of \S\ref{sec:miss}.

We bound scale, capacity and architecture at $\leq 0.007$ on the sequence families, with the
architecture bound taken at 16 weeks and with the clocks as built. Each axis was varied one factor at a time, with the recipe
fixed within each family apart from the tier contrast's realized schedule. The three-seed
detection floor we measured on the one class we could pool sits at the scale of those bounds
rather than under them (\S\ref{sec:limitations}). We therefore claim the order-of-magnitude
contrast between the window and the three bounded axes. We do not claim equivalence among
settings on any one of them. In log space and against our rotary encoding as implemented, time representation gives the largest model-side improvement of any intervention we
tested. It clears the scale, capacity and architecture
bounds and composes with the window. Matched cohorts, bit-exact build anchors, and internally
registered predictions with committed falsifiers are reusable measurement machinery. The
matched cohorts and committed falsifiers are what let such a prediction fail decisively
(\S\ref{sec:prereg}). A team building on this class of data should start by lengthening the
window to at least the probe's saturation point on the test calendar (\S\ref{sec:discussion}). On tasks like this one, run a cheap ridge check on longer history before scaling the model.

\section*{Use of AI assistants}

\textbf{Used.} An AI coding assistant wrote and edited research code: model variants, the data
and evaluation harnesses, the scripts that derive the public datasets' window extracts from their raw data, and
the analysis scripts behind the tables and figures. It drafted and edited prose. It surfaced
candidate related work, which we read and cited ourselves. It produced intermediate analyses and
summaries of run outputs, and it assisted in interpreting results: it proposed readings of
run outputs and of the fitted saturation law, which we checked against the recorded artifacts
and either adopted, narrowed, or discarded. Where a reading could not be sourced to
those artifacts, we narrowed or dropped it. Because the extract and harness code reformats data, we count this as
assistance with cleaning and reformatting datasets. One of the paper's organizing ideas also came
out of that assisted analysis rather than from us unaided: that the pooled saturation law is consistent with a
mixture, so context is worth most where activity is lowest. We count this as help developing
a conceptual framework. We re-derived it from the recorded artifacts before adopting it, and the
claim and its stated scope are ours. The predicted value against which commitment (iii) is scored (the mixture model's
pooled gain from week 24 to 32) was computed in that assisted per-decile analysis.
It also drafted the toy model in \S\ref{app:toy}, its short derivations, and the
public-data predictions in \S\ref{app:public-ladder} that were written from it before those
runs. We count this as help developing a theoretical model, formulating mathematical claims and
writing proofs; we checked each derivation ourselves.
For the checks added after the four registered commitments, it also drafted the experiment designs, the
registered decision rules and thresholds, and some of the hypotheses those checks test; we reviewed them, adjusted them where needed, and fixed
each before its run.

\textbf{Not used.} We did not use it to generate or augment training or evaluation data, for
translation, or for qualitative or thematic analysis.
No number reported here was
produced by a language model: each was computed by code from runs recorded in our own artifacts and
was verified against them before it was cited.

The paper states no formal theorems; the derivations of \S\ref{app:toy} are covered under Used. The two
fitted functional forms, the saturating and log-linear laws, were specified by us in advance of
the fit.

Where the assistant produced analysis, the readings drawn from it are ours and were checked
against the recorded artifacts before use. We reviewed and adopted the experimental design and the internally
registered criteria; all claims and their stated scopes are ours, as is responsibility for this paper.

\bibliographystyle{plainnat}
\bibliography{references}

\clearpage
\appendix

\section{Calendar check: window length versus holiday season}
\label{app:calendar}

Every test anchor falls between 2025-02-24 and 2025-04-21 (\S\ref{sec:data}), so lengthening the window from 8
weeks to 16--24 weeks also brings November and December into the input, the confound of \S\ref{sec:intro}. To
separate the two for the weekly-aggregate ridge probe of \S\ref{app:probe}, we rerun it (same features, same
ridge with leave-one-out penalty choice, same target) on anchors from the training split of the M24 extract and score it on
held-out training customers. This is a validation-style check: fit and score both come from the training period, so
its $R^2$ levels are not comparable to the test-split levels of \S\ref{sec:saturation}, and we interpret only the gains and slope
ratios within it. The anchors fall every 28 days, and we use two groups of them:
\begin{itemize}[leftmargin=1.4em,itemsep=1pt,topsep=2pt]
  \item \emph{Holiday-free}: 2024-06-17, 2024-07-15, 2024-08-12 and 2024-09-09. No window of up to 24 weeks holds a
        day of November or December.
  \item \emph{Test-calendar}: 2024-02-26, 2024-03-25 and 2024-04-22, one per test anchor at the same week of the
        year, so holiday weeks enter the window as they do on the test split (to within two edge days).
\end{itemize}
The other two training anchors, 2024-01-29 and 2024-05-20, fit neither group's definition. The holiday-free group needs
every window up to 24 weeks clear of November and December, because the 22--24-week bend is also estimated in that
group; the 24-week window of 2024-05-20 starts 2023-12-04. The test-calendar group holds one anchor at the same week of
the year as each test anchor, and 2024-01-29 matches none.
Customers are split 80/20 by a fixed hash of their id, and a customer stays on one side at every anchor. Scaling and
ridge are fit within each group on the 80\% and scored on the 20\% ($447{,}322$ holiday-free and $335{,}885$
test-calendar score rows; score-side target variance $7.64$ in both). Intervals are 95\% percentile intervals from a
customer-clustered paired bootstrap over score-side customers, jointly across both groups ($B{=}1000$, order
statistics 25 and 976). They cover scoring noise only, not fit noise or variation across anchors. The groups, the estimands and the decision rule were
fixed before this check computed any $R^2$. Before either group was evaluated, the probe had to reproduce the $R^2(k)$ curve of
\S\ref{sec:saturation} at $k{=}8, 16, 22, 24$ on the matched four-window test cohort ($n{=}372{,}778$), and it does so exactly.

\emph{The window gain persists outside the holiday season.} The probe's 8-to-16-week gain, $R^2(16)-R^2(8)$, is
$0.0352~[0.0342, 0.0361]$ on test-calendar anchors and $0.0253~[0.0246, 0.0259]$ on holiday-free anchors. The
holiday-free gain is $0.72~[0.69, 0.75]$ times the test-calendar gain. The pre-stated rule concludes ``persists'' when the
holiday-free gain's interval excludes zero and that ratio's interval lies at or above $0.5$, and ``mostly seasonal''
when the ratio's upper limit is below $0.5$. So on this probe most of the gain survives without the holidays, and
holiday coverage is not its main source. The remainder may still be calendar-related.
A shuffled-weeks control adds no $R^2$ in either group: permuting weeks 8--15 among customers at the same anchor
changes $R^2$ by $-0.000017~[-0.000027, -0.000007]$ on the holiday-free anchors and $-0.000013~[-0.000026, 0.000001]$
on the test-calendar anchors. With anchor indicators added to every fit, the conclusion is unchanged: the gains are
$0.0252~[0.0245, 0.0258]$ holiday-free and $0.0354~[0.0344, 0.0363]$ test-calendar, their ratio is
$0.71~[0.69, 0.74]$, so the gain persists outside the holiday season, and that refit's own shuffled control also adds
no $R^2$.

\emph{The sharp bend persists on the test calendar.} We call the bend present when the 95\% upper limit of the ratio of the per-week slope over
$k{=}22$--$24$ to the slope over $k{=}8$--$16$ is under a tenth. On test-calendar anchors, scored on all their held-out customers, it is present, with a
ratio of $0.071~[0.056, 0.086]$. On holiday-free anchors the ratio is $0.180~[0.160, 0.201]$, so the bend is not shown to persist.
Returns still diminish in both groups: the last weeks, $k{=}22$--$24$, add less per week than $k{=}8$--$16$
($R^2(8), R^2(16), R^2(24) = 0.435, 0.461, 0.467$ holiday-free, against $0.430, 0.465, 0.476$ test-calendar), and a
saturating curve fits better than a log-linear one in both groups (RMSE $0.0050$
vs.\ $0.0163$ holiday-free, $0.0072$ vs.\ $0.0126$ test-calendar; descriptive only, no test). With anchor indicators added to every fit, the bend persists on test-calendar anchors, scored on all their held-out customers, and is not shown to persist on holiday-free ones. We therefore scope the sharp
22--24-week bend to the test calendar.
The public datasets of \S\ref{app:public-ladder} do not show it either: none of the three whose bend is evaluated
meets that criterion, and CDNOW's bend is not evaluated. The saturation results of \S\ref{sec:saturation} are measured on the test
anchors, the calendar where the bend persists, so their numbers stand; this check limits only how far the bend's location
generalizes.

\emph{Scope.} This is the weekly-aggregate ridge probe on training-period anchors with an 80/20 customer split, not
the transformer and not the test-split numbers of \S\ref{sec:saturation}. The two groups differ in season as well as
holidays (summer vs.\ spring), so the comparison shows the gain does not need holiday coverage, but cannot attribute
the gap between the groups to the holidays alone. Two of the three test-calendar anchors, 2024-02-26 and 2024-03-25, also have only partial event history at or past lags 17 and 21, respectively (the train-side availability boundary of \S\ref{sec:limitations}). That range covers both weeks the $k{=}22$--$24$ step adds, so part of that group's sharper bend may come from missing history rather than the calendar. The 8-to-16-week gains, and hence the ratio $0.72$, use lags 8--15, which lie on the recent side of that boundary at every anchor here. The check is separate from the anchor-varying transformer control of
\S\ref{sec:intro}, whose void verdict stands.

\emph{The transformer on the same split.} We repeated the check with the paper's d128, 2-layer transformer on the
corrected inputs (real category codes and standardized numeric fields). It was trained once per window, at 8 and 16
weeks, on the same 80\% of customers at all nine training anchors, at seeds 42, 123, 456, 789 and 1011, and scored on
the held-out 20\% in the two groups, on the rows present at both windows ($385{,}015$ holiday-free and $284{,}047$
test-calendar). The groups and the decision rule were the ridge check's, fixed for the transformer before any run
started; intervals are 95\% $t$ intervals over seeds ($4$ degrees of freedom), and the ratio's is Fieller's.

The 8-to-16-week gain is $+0.009~[+0.007, +0.011]$ on test-calendar anchors and $+0.011$ $[+0.008, +0.013]$ on
holiday-free ones, and the holiday-free gain is $1.14~[1.08, 1.20]$ times the test-calendar gain, so by the same rule
the transformer's gain also persists outside the holiday season.

These intervals cover seed noise on one held-out set,
while the ridge probe's above cover scoring noise, and the two models differ in fit: the transformer trains once on
all nine anchors and is scored per group, where the ridge fits within each group. So the two are not compared. The
transformer's read is validation-style, on held-out training customers, not the test-split numbers of
\S\ref{sec:saturation}, and no test-split transformer score is set beside it. Its input scaler was fit on the full
training population, held-out customers included (under $1\%$ of it), through the mean and standard deviation of
quantity and price only. Every matched row has an input event in both windows. As with the ridge check, the groups
differ in season as well as holidays.

\emph{What the longer window carries.} A second check on the same probe asks whether the 8-to-16-week gain needs to
know which of the older weeks the activity fell in, or only the older weeks' totals. Weeks count back from the anchor
starting at 0 (\S\ref{app:probe}), so the 8-week fit holds weeks 0--7 and the 16-week fit adds weeks 8--15. The
probe's features are, per week, the log of one plus the clipped-at-zero activity sum and the event count. We fit it on
every training row and score it on the matched four-window test cohort ($n{=}372{,}778$ rows), along a nested chain of
inputs: the 8-week features; plus the number of events in weeks 8--15; plus their summed weekly log activity (the sum
over those weeks of $\log(1+a_w)$, with $a_w$ the week's clipped activity sum); then the full 16-week features, whose
only extra freedom over the step before is a separate activity weight and count weight for each of weeks 8--15. By
construction, the three increments sum to the gain. The probe's gain on this cohort is $0.0573~[0.0558, 0.0586]$. The event
count adds $0.0051~[0.0045, 0.0057]$, the summed log activity after it $0.0522~[0.0505, 0.0536]$, and the per-week
weights $-0.00002~[-0.00007, 0.00002]$, a share of $-0.04\%~[-0.12\%, 0.04\%]$ of the gain. The rule, fixed before
any of the new inputs was fit, takes ``more events'' to mean the older weeks' order-free totals, their event count
and summed weekly log activity together, not the count alone. It draws no conclusion if the gain's interval is not above
zero or the per-week part's interval lies wholly below $-0.0001$ (neither happens); otherwise it concludes ``mostly more
events'' when the share's upper limit is below a tenth, ``not only more events'' when its lower limit is at or above
a quarter, and ``both'' in between. The result is ``mostly more events'': beyond the older weeks' event count and summed
weekly log activity, which of weeks 8--15 the activity fell in carries under a tenth of the gain. Repeating the decomposition within the
holiday-free group above (fit on its 80\%, scored on its 20\%; its anchors' weeks 8--15 hold no November--January
week) gives a gain of $0.0253~[0.0246, 0.0259]$, as above. The event count adds $0.0014~[0.0010, 0.0017]$, the summed
log activity $0.0239~[0.0232, 0.0245]$, and the per-week weights $0.00002~[0.000005, 0.00004]$, a share of
$0.08\%~[0.02\%, 0.15\%]$, so the conclusion is the same. The per-week part's interval lies above zero there, unlike on the test
cohort, but stays far under a tenth of the gain.

\emph{Scope.} This is again a linear ridge on weekly aggregates, not the transformer, so it covers only timing that a
linear weight per week can use. Summed log activity also grows with how many of the older weeks had activity and, relative
to weeks 0--7, with a change in rate, so the result does not show that the amount of older activity is what matters.
Those totals also still come from the longer window, so it does not show that window length is irrelevant. The split
between count and activity depends on the order in which they are added, so we do not interpret it. The per-week weights are indexed by
distance from the anchor; holiday weeks, where an anchor's window has them, sit at a different distance for each
anchor; and the totals count holiday activity wherever it falls. This check is therefore not a holiday test; the two groups above
address the holidays, within the limits stated there. Intervals come from a customer-clustered bootstrap with fits held
fixed ($B{=}1000$, order statistics 25 and 976), so they cover test sampling only. Training refit noise is not negligible
when the training anchors sit at calendar positions other than those of the test anchors, as here: in the decision rule's simulations it
was $0.2$--$0.65$ times the test-sampling SD, which would put a nominal 95\% interval's coverage at about 90--95\%.

\emph{Training labels that end before the test period.} The probe of \S\ref{app:probe} fits on all nine anchor dates
of the M24 extract's train split, and seven of them have 364-day targets that run into the calendar period of the test
targets. To check that the gain does not depend on training labels that share calendar days with test labels, we refit
it on only the two train-split dates, 2024-01-29 and 2024-02-26, whose targets end by the first test target day
(the overlap is no more than one day), and score it on the same matched four-window test cohort ($n{=}372{,}778$). Each fit
includes an indicator for each training date, which removes that date's level, so the gain comes from differences
between customers at the same date. A first version without the date indicator failed its negative control; the
decision rule for this version was fixed after that failure and before any value of its gain existed. The 8-to-16-week gain
holds when no training label shares a day with the test period: $R^2(16)-R^2(8) = 0.0509~[0.0493, 0.0523]$.
Descriptive only: $R^2$ goes from $0.3515$ at $k{=}8$ to $0.4024$ at $k{=}16$; permuting weeks 8--15 across customers
within each date gains nothing ($-0.0001~[-0.00012, -0.00008]$); and the nine-date fit of \S\ref{app:probe} gives a
gain of $0.0573$ on the same cohort. The two fits differ in rows, calendar mix and features, so the two gains are not
compared. \emph{Scope.} This check uses ridge on weekly aggregates at the 1M tier, on this paper's extracts; it supports only the direction of the gain, and its intervals
cover test sampling only.

\section{Public datasets spanning more than a year}
\label{app:public-ladder}

The Taobao check of \S\ref{sec:external} spans 9 days, so it cannot speak to windows measured in months. We therefore
run the same weekly-aggregate ridge probe (\S\ref{app:probe}) on four public datasets that each span more than a year.
The first is Online Retail II~\citep{chen2012onlineretail,chen2012rfm}, a public transaction log from one UK online
retailer with many wholesale buyers, covering 2009-12-01 to 2011-12-09. The other three, The Complete Journey, CDNOW
and H\&M, are reported later in this appendix. These four are all the public datasets we downloaded for this
check; the other candidates were set aside before any download, one on data-use grounds and three as unfit for the
probe (dummy data, no total-spend target, not retail), so none was chosen for its result.
\begin{itemize}[leftmargin=1.4em,itemsep=1pt,topsep=2pt]
  \item \emph{Data.} Lines without a customer id and exact duplicate lines are dropped. Deduplication also removes the
        overlap between the file's two sheets.
  \item \emph{Anchors.} Eight anchors fall every 28 days from 2010-05-24 to 2010-12-06, so each has 24 weeks of
        history and a 364-day target inside the span.
  \item \emph{Rows.} A row is a customer at an anchor with at least one non-canceled purchase in the 8 weeks before
        it: $13{,}665$ rows from $3{,}870$ customers. Every compared window (8 weeks or longer) sees those weeks, so
        the selection cannot favor longer windows, and every $k$ scores the same rows.
  \item \emph{Features.} Per week, the log of one plus net spend, floored at zero (cancellations count as
        negatives), and the number of distinct non-canceled invoices.
  \item \emph{Target.} Net spend over the 364 days from the anchor, as $\log(1+\max(0,\cdot))$.
  \item \emph{Fit and score.} Ridge per $k$, with 5-fold cross-fitting by customer, scored as pooled out-of-fold
        $R^2$.
  \item \emph{Intervals and decision rule.} Intervals come from a customer-clustered bootstrap with fits held fixed
        ($B{=}999$, order statistics 25 and 975). The decision rule, including a negative control, was fixed before
        any number existed. Under it, the 8-to-16-week gain counts as helping if its interval lies wholly above zero, the
        permuted control's interval must not lie wholly above zero, and the bend counts as shown only if the
        $k{=}8$--$16$ slope is positive and the late-to-mid slope ratio's 95\% upper limit is under a tenth, as in
        \S\ref{app:calendar}.
\end{itemize}

\emph{Longer history helps here too.} $R^2$ is $0.134$ at $k{=}8$, $0.189$ at $k{=}16$ and $0.202$ at $k{=}24$. The
8-to-16-week gain is $0.055~[0.046, 0.062]$, and the 8-to-24-week gain is $0.068~[0.057, 0.079]$. The negative control
permutes weeks 8--15 across customers within each anchor, and it gains nothing ($-0.0013~[-0.0023, -0.0003]$), so the
gain comes from each customer's own purchases in those weeks, not from having more features. Returns diminish, but
the bend is not shown: the late-to-mid slope ratio is $0.21~[0.13, 0.32]$. The ridge penalty sits at the top of its
grid in 3 of 5 folds from $k{=}18$ and 4 of 5 at $k{=}22$--$24$ (and in none at $k{=}8$ or $16$), so the 24-week gain
may be slightly understated and the bend ratio may shift slightly either way. The 16-week gain is not affected.

\emph{Scope.} This check uses one small public retailer and ridge on weekly aggregates. It shows that a months-scale history
effect appears on a public multi-year dataset as well as on ours. This check says nothing about scale, the transformer, or the
size of the main dataset's effects. Its anchors (May--December 2010) do not share the main test split's
February--April calendar. The weeks the 16-week window adds hold no day from November to January (across the anchors they run from 2010-02-01 to 2010-10-10), so the 8-to-16-week gain does not come from added holiday weeks. The 8-week window itself holds November days for the last two anchors (and December 1 to 5 for the last), days that both windows see. The weeks the 24-week window adds hold December or January days for the first two anchors.
Earlier anchors' target periods overlap later anchors' input weeks in calendar time (different rows, with folds by
customer), so it is not out-of-time validation.

\paragraph{A second multi-year dataset: The Complete Journey.} We applied the same probe and decision criteria to
dunnhumby's The Complete Journey~\citep{dunnhumby2023completejourney}, about two years of grocery purchases by
$2{,}500$ frequent-shopper households of one retailer. Its days are numbered, with no calendar dates. Households join
the panel over its first months, so before a household joins, a zero week means it was not in the panel yet, not that
it did not shop. To keep those zeros out of the longer windows, the three anchors (days 292, 320 and 348, each with a
364-day target inside the file) all have 24-week histories that start after $99.6\%$ of households had joined (9 rows
still have the household's first basket inside their history). A row is a household with a basket in the 8 weeks
before an anchor: $6{,}524$ rows from $2{,}311$ households. Per week the features are the log of one plus spend and
the number of baskets; the target is the log of one plus spend over the 364 days from the anchor. Intervals use a
household-clustered bootstrap with fits held fixed ($B{=}999$). The decision rule was fixed before any Complete
Journey $R^2$ was computed, and the shared code reproduces the Online Retail II run above to within $10^{-10}$.

\emph{Longer history helps here too.} $R^2$ is $0.4822$ at $k{=}8$, $0.5137$ at $k{=}16$ and $0.5176$ at $k{=}24$. The
8-to-16-week gain is $0.0315~[0.0245, 0.0384]$, and the 8-to-24-week gain is $0.0354~[0.0270, 0.0439]$. The same
permuted control gains nothing ($-0.0017~[-0.0030, -0.0004]$). Returns diminish, but by the rule the bend is not
shown: the late-to-mid slope ratio is $0.005~[-0.108, 0.114]$, and its upper limit is not under a tenth. Unlike Online
Retail II, the ridge penalty stays below the top of its grid at every $k$. Households whose last basket comes more
than 28 days before their target period ends ($9.2\%$ of rows; here a household that stops buying looks the same as one that leaves the panel) account for $26.6\%$ of the target's variance and carry $27.7\%$ of the 8-to-16-week drop in squared error. Refit
without them, the gain is $0.0312$. These are descriptive point values with no interval, so we draw no conclusion from the
small gap between the shares, and we do not interpret the refit because leaving those rows out conditions on the target period.

\emph{A prediction across the two datasets.} Before this run, the decision rule recorded the prediction that these households would gain
less than Online Retail II's customers. In a toy model where each customer buys at their own Poisson rate (\S\ref{app:toy}), a $T$-week
count estimates that rate with reliability $x/(x+1)$, where $x = T\,\mathbb{E}[\lambda]\,\mathrm{CV}^2(\lambda)$, and,
once $x > 1/\sqrt{2}$, the gain from doubling $T$ shrinks as $x$ grows. At $T{=}8$ weeks, the window being doubled, both
datasets are above that threshold ($x$ about $12.9$ here and $2.8$ on Online Retail II, measured from two disjoint 4-week
counts), with a median of 7 baskets per row in the 8 weeks before the anchor here against 1 invoice there. The gain is
smaller here: on absolute $R^2$, the scale the rule uses, $0.0315$ here minus $0.0549$ there is
$-0.0234~[-0.0344, -0.0124]$, so the prediction holds. As a share of the $R^2$ left to gain at $k{=}8$,
$G_{16}/(1-R^2(8))$ with $G_{16}$ the 8-to-16-week gain, which the rule reports as descriptive only, the two are close:
$0.061~[0.047, 0.074]$ here and $0.063$ there (from its recorded values, no interval). Online Retail II's gain was
known when the prediction was written, so the comparison orders two datasets with one side already seen, and the two differ in
much besides purchase frequency (what the target is made of, how lapse shows up, tenure mix, season, and how
customers came to be in the data), so we do not attribute the gap to frequency.

\emph{Scope.} This check uses one grocer's frequent-shopper panel and ridge on weekly aggregates. With Online Retail II it
shows a months-scale history gain on two public multi-year datasets. This check says nothing about scale, the transformer, or
the size of the main dataset's effects, and with no calendar dates it says nothing about season or holidays. Earlier
anchors' target periods overlap later anchors' input weeks (different rows, with folds by household), so it is not
out-of-time validation.

\paragraph{A third public dataset: CDNOW.} We applied the same probe and decision criteria to the CDNOW master
file~\citep{fader2001cdnow}: every purchase through June 1998 by the $23{,}570$ people who first bought from the online
music store CDNOW in the first quarter of 1997, 18 months in all. The file has no order id, so an occasion is a distinct
purchase day, and every purchase record is kept. Under the shared anchor rule, only one anchor, 1997-06-23, has a 24-week history and a
364-day target inside the file, and a row is a customer with a purchase in the 8 weeks before it: $3{,}445$ rows
($14.6\%$ of the cohort), one per customer. Unlike on Complete Journey, no such anchor can keep customers' join dates out of the longer
windows. Every first purchase falls between 1997-01-01 and 03-25, so the longer windows hold most rows' first-ever
purchase (inside 16 weeks for $31.1\%$ of rows, inside 24 for $95.7\%$; the other $4.3\%$ first bought in the 5 days
before the 24-week history starts), and the weeks before it are zero because the person was not a customer yet. Per
week the features are the log of one plus spend and the number of purchase days; the target is the log of one plus
spend over the 364 days from the anchor. The file is the cohort's whole purchase history, so there is no panel to leave
and the panel-exit check above does not apply. Intervals use a customer bootstrap with fits held fixed ($B{=}999$). The decision rule was fixed
before any CDNOW fit ran, and the run uses the Complete Journey code with only CDNOW data loaders added.

\emph{Longer history helps here too.} $R^2$ is $0.1188$ at $k{=}8$ and $0.1505$ at $k{=}16$, and the 8-to-16-week gain
is $0.0317~[0.0198, 0.0432]$. The same permuted control gains nothing ($-0.0038~[-0.0070, -0.0009]$). Because the
16-week window also holds the join date for $31.1\%$ of rows, this gain means more history including when the customer
joined, not more history alone, and whether the join date makes it larger or smaller than an unconfounded measure is not known.
The rule, fixed before the run, therefore also evaluates the gain on the other $2{,}375$ rows, whose first purchase came by
\mbox{1997-03-02} and whose 8- and 16-week windows fall wholly after joining. There the 8-to-16-week gain is
$0.0462~[0.0311, 0.0614]$ and the control again gains nothing ($-0.0104~[-0.0159, -0.0048]$): longer history helps here
too, with the join date outside those weeks. The two row sets overlap and use different folds and seeds, so we do not
compare their gains. We do not report the 8-to-24-week gain, whose window holds the join date for $95.7\%$ of rows, and,
under the rule, the bend is not evaluated here: weeks 22 and 23 come before the first purchase for $83.3\%$ of rows, so the late
slope sits near zero by construction. As on Complete Journey, the ridge penalty stays below the top of its grid at
every $k$, on both row sets.

\emph{A prediction across datasets.} Before this run, the decision rule recorded the prediction that CDNOW's customers would gain more than
Complete Journey's households. In the toy model (\S\ref{app:toy}), the gain from doubling $T$ is largest at $x = 1/\sqrt{2}$; here
$x$ is about $0.85$, measured the same way, against $12.9$ on Complete Journey (a median of 1 purchase day per row in
the 8 weeks before the anchor here, against 7 baskets there), so the toy ranks CDNOW higher. The rule also noted before
the run that the toy model ranks reliability rather than $R^2$, and that the prediction's main uncertainty lay in the unknown
share of each dataset's target variance that a perfectly known rate would explain, rather than in $x$. On absolute $R^2$, the scale the
rule uses, CDNOW's gain minus Complete Journey's is $0.0003~[-0.0129, 0.0136]$, so the prediction is unresolved: the
interval allows either ordering. Against Online Retail II the toy made no prediction; the difference there is
$-0.0232~[-0.0373, -0.0094]$, shown as descriptive. Online Retail II's and Complete Journey's gains, and Complete Journey's prediction
result, were known when this prediction was written; CDNOW's gain was not. The datasets also differ in much besides
purchase frequency (who the customers are, the join date in CDNOW's longer windows, what the target is made of, what
counts as an occasion, season, and era), and any of these could push the ordering either way.

\emph{Scope.} This check uses one online music store's new customers from one quarter of 1997, a single acquisition cohort
rather than a store's customer base. It is scored on the $14.6\%$ still buying at one forecast date, each 90 to 173
days past their first purchase, with ridge on weekly aggregates. The weeks the 16-week window adds (1997-03-03 to
04-27) hold no day from November to January. The intervals cover scoring noise at that one date, so nothing in them
shows what another anchor or target year would give. With Online Retail II and Complete Journey it shows a
months-scale history gain on three public datasets that each span more than a year, on CDNOW with the join date inside
the longer windows. This check says nothing about scale, the transformer, or the size of the main dataset's effects. With one
anchor, no target period overlaps another anchor's input weeks, but every row shares one date and folds are by
customer, so it is not out-of-time validation either. The data are used as released, with no modification or
redistribution, and only aggregate results are reported.

\paragraph{A fourth public dataset: H\&M.} We applied the same probe and decision criteria to the transactions of the H\&M
Personalized Fashion Recommendations competition~\citep{hm2022kaggle}: $31{,}788{,}324$ purchase lines by
$1{,}362{,}281$ customers of one fashion retailer from \mbox{2018-09-20} to \mbox{2020-09-22}. The file has no order id or quantity,
so an occasion is a distinct purchase day, and every purchase record is kept (a repeated record is a second item). Prices come in an
unstated rescaled unit, so spend is price divided by the file's median line price, a constant fixed before any fit. The
anchors follow Online Retail II's rule: eight Mondays 28 days apart, 2019-03-11 to 2019-09-23. A row is a customer with
a purchase in the 8 weeks before an anchor: $3{,}164{,}024$ rows from $829{,}771$ customers, a selection every compared
window sees. Per week the features are the log of one plus spend and the number of purchase days; the target is the
log of one plus spend over the 364 days from the anchor. Folds are by customer, so a customer sits in one fold at every
anchor, and intervals use a customer-clustered bootstrap with fits held fixed ($B{=}999$). The decision rule was fixed before any
H\&M fit ran.

\emph{Longer history helps here too.} $R^2$ is $0.1868$ at $k{=}8$ and $0.2892$ at $k{=}16$, and the 8-to-16-week gain
is $0.1024~[0.1017, 0.1030]$ (descriptive, as a share of the $R^2$ left to gain at $k{=}8$:
$G_{16}/(1-R^2(8)) = 0.1259~[0.1250, 0.1267]$). The same permuted control gains nothing
($+0.000002~[-0.000006, +0.000010]$). Returns diminish, but the bend is not shown: the per-week slope over
$k{=}22$--$24$ is still about a third of the slope over $k{=}8$--$16$ (the late-to-mid slope ratio is
$0.343~[0.337, 0.349]$; a bend needs its upper limit under a tenth). The ridge penalty stays below the top of its grid
at every $k$. Two caveats concern the longer windows. First, a customer's first line in the file can fall
inside them, and the file does not say whether that is a new customer or one returning after a gap. Second, seven of the eight
target years run into the COVID period (from 2020-03-11, by 26 to 194 days); a level shift common to an anchor's rows
is present at every $k$, but anchors are pooled and the longer windows see more of each anchor's season, so part of the
pooled gain could come from telling anchors apart. To take the holiday season out of the added weeks, a follow-up
refit uses only the five anchors from 2019-06-03 to 2019-09-23, whose whole 16-week history holds no November,
December or January day, with each anchor's level taken out of the fit by anchor indicators: the gain shows without
holiday weeks in the added history, $0.1054~[0.1045, 0.1062]$ on $2{,}020{,}268$ rows (descriptive: $R^2$ $0.1846$ to
$0.2900$; permuted control $-0.000006~[-0.000012, -0.000001]$). A first version without the anchor indicators failed
its negative control, so this one adds them; the decision rule for this version was fixed after the pooled run and that
failure, and before any value of its gain was written. Those five target years end 2020-05-31 to 2020-09-20 and hold 82 to 194 days of
the COVID period, so this subset trades the holiday question for a heavier COVID share, and the indicators remove each
anchor's level but not differences across customers in how COVID changed buying. We do not compare this gain with the
pooled one: the rows, fits, COVID share and features all differ.

A second follow-up takes the holiday season out of the whole 24-week history and asks whether returns still diminish
and whether the bend appears. It uses the three
anchors from 2019-07-29 to 2019-09-23, whose 24-week histories start 2019-02-11 to 2019-04-08 and lie wholly before COVID ($1{,}197{,}585$ rows from
$582{,}862$ customers), with the same anchor indicators in every fit, and its decision rule was fixed before any fit on
these rows. Returns diminish without holiday weeks in the 24-week history, and the bend is not shown without holiday
weeks: the late-to-mid slope ratio is $0.242~[0.233, 0.251]$, against the same bar of a tenth (descriptive: $R^2$ is
$0.178$, $0.288$, $0.325$ and $0.332$ at $k{=}8$, $16$, $22$ and $24$, and neither permuted control adds $R^2$: the
mid-weeks control's interval spans zero, the late-weeks control's lies below it). These three target years end 2020-07-26 to 2020-09-20 and hold 138 to 194 days of the COVID period; the
indicators absorb each anchor's level but not differences between customers in how COVID changed their buying. As
with the five-anchor refit, we do not compare this ratio with the pooled one.

\emph{A prediction across datasets.} The decision rule also recorded a prediction before the run, from the toy model (\S\ref{app:toy}): past
$x = 1/\sqrt{2}$, more frequent buying means a smaller gain from doubling the window. H\&M's $x$ is about $1.25$,
measured the same way, against $12.9$ on Complete Journey (a median of 1 purchase day per row in the 8 weeks before the
anchor, against 7 baskets there), so both sit past that point and the toy ranks H\&M higher. The prediction holds: H\&M's gain minus
Complete Journey's is $0.0709~[0.0639, 0.0778]$. Against Online Retail II the toy made no prediction at these $x$
values; the difference there is $0.0475~[0.0401, 0.0561]$, shown as descriptive. The rule that decides from $x$ which ordering, if any, the toy predicts (the case rule) was written before the
file arrived, and the case was set from purchase-day counts before any fit. Online Retail II's and Complete Journey's
gains, and Complete Journey's prediction result, were known when the case rule was written, and public descriptions of
the file's size and span were recalled while drafting; an earlier draft predicted the same ordering on a looser rule,
and this case rule replaced it before the file arrived. The datasets also differ in much else
(what is bought: fashion items, grocery baskets, invoices from an online gift retailer whose customers are often
wholesalers, CDs; what an occasion is; COVID in seven of H\&M's eight target years; how customers came to be in each
file; calendar; and spend in a rescaled unit), so, although the ordering is the one the toy predicted, we do not
attribute it to purchase frequency.

\emph{Scope.} This check uses one fashion retailer's customers over two years and ridge on weekly aggregates. The intervals
cover scoring noise only. Earlier anchors' targets overlap later anchors' input weeks in calendar terms (different
rows; folds are by customer), so it is not out-of-time validation. With Online Retail II, Complete Journey and CDNOW it
shows a months-scale history gain on four public datasets that each span more than a year, and this check says nothing about
scale, the transformer, or the size of the main dataset's effects. The data are used under the competition's
non-commercial research terms, with no redistribution, and only aggregate results are reported.

\paragraph{The paper's transformer on three of these datasets.} We also trained the paper's transformer on Online
Retail II, The Complete Journey and CDNOW, on the same rows and customer folds as the ridge probes above, with five
seeds per window, and scored it out of fold. Its reading rule was fixed before any run started. On each dataset, read
on its own, the transformer gains from 16 weeks of history over 8 too (Table~\ref{tab:public-tx}).

\begin{table}[!t]
\caption{The paper's transformer on three public datasets. On each dataset, read on its own, the transformer gains
from 16 weeks of history over 8 too. Gain = out-of-fold $R^2$ with a 16-week history minus with an 8-week one, in log
space, over all rows, each scored by the model whose test fold holds it; the mean over 5 seeds. Seed interval: $t$ on
the 5 per-seed gains (4 df). Scoring interval: a customer bootstrap with the models held fixed, 999 draws, the same
draws for both windows and every seed. The intervals cover seed noise and the noise of scoring these rows, not the
choice of folds. Datasets: Online Retail II, a UK online gift retailer with many wholesale buyers, 2009 to 2011; The
Complete Journey, a grocer's $2{,}500$-household panel over about two years, with no calendar dates; CDNOW, one
quarter's new customers of an online music store, with every purchase they made over 18 months. Rows are the ridge
probes': one per customer and anchor with at least one purchase in the 56 days before the anchor, cross-fit by
customer over the ridge probes' five folds. Earlier anchors' targets overlap later anchors' inputs in calendar time,
so this is not out-of-time validation. Target: spend over the 364 days from the anchor, as $\log(1+\max(0,\cdot))$.
Model: the paper's d128, 2-layer transformer and loss, with batch and warmup scaled to these sizes and the learning
rate left at $5{\times}10^{-4}$, untuned (it was set at a batch of $8{,}192$). It sees each line's quantity, price and
spend and its place in the sequence, oldest first, with no date and no item or category; the paper's recipe reads no
calendar or timing feature on the main data either. So the longer window adds older lines, in order, without their
dates. On Online Retail II, cancellations enter as lines with negative quantity. On CDNOW every customer joined in the
first quarter of 1997, and for $31\%$ of rows the 16-week history reaches back to the first purchase, which to a model
with no dates looks like a shorter sequence, as a light buyer's does. On 3 of the 15 folds the learning-rate schedule
was sized for $0.7\%$ to $5.6\%$ more steps per epoch than ran, so it ends at about $1.0\%$ to $1.75\%$ of its peak
rather than at $1\%$, and one fold's warmup is 76 steps, not 72. Each dataset is used under its own terms (CC BY 4.0
for Online Retail II; The Complete Journey and CDNOW
used as released, with no modification or redistribution), and only aggregate results are reported.}
\label{tab:public-tx}
\centering\footnotesize
\setlength{\tabcolsep}{2pt}
\begin{tabular}{lrccccc}
\toprule
\textbf{Dataset} & \begin{tabular}[b]{@{}c@{}}\bfseries Rows /\\\bfseries customers\end{tabular} & \textbf{Gain} & \begin{tabular}[b]{@{}c@{}}\bfseries 95\% seed\\\bfseries interval\end{tabular} & \begin{tabular}[b]{@{}c@{}}\bfseries Scoring\\\bfseries interval\end{tabular} & \begin{tabular}[b]{@{}c@{}}\bfseries $R^2$ at $k{=}8$\\\bfseries / $k{=}16$\\\bfseries (descriptive)\end{tabular} & \begin{tabular}[b]{@{}c@{}}\bfseries Ridge probe\\\bfseries gain (different\\\bfseries model, same\\\bfseries rows and folds)\end{tabular} \\
\midrule
Online Retail II & $13{,}665$ / $3{,}870$ & $0.080$ & $[0.072, 0.088]$ & $[0.071, 0.088]$ & $0.123$ / $0.203$ & \begin{tabular}[t]{@{}c@{}}$0.055$\\$[0.046, 0.062]$\end{tabular} \\
The Complete Journey & $6{,}524$ / $2{,}311$ & $0.027$ & $[0.020, 0.034]$ & $[0.012, 0.040]$ & $0.515$ / $0.541$ & \begin{tabular}[t]{@{}c@{}}$0.031$\\$[0.024, 0.038]$\end{tabular} \\
CDNOW & $3{,}445$ / $3{,}445$ & $0.030$ & $[0.020, 0.040]$ & $[0.016, 0.045]$ & $0.065$ / $0.096$ & \begin{tabular}[t]{@{}c@{}}$0.032$\\$[0.020, 0.043]$\end{tabular} \\
\bottomrule
\end{tabular}
\end{table}

\paragraph{The paper's transformer on H\&M.} We also trained the paper's transformer on H\&M, on the ridge probe's rows
above, split by customer into training, validation and test customers, so a customer sits in one split at every
anchor. It ran with three seeds per window and was scored on the test customers. Its reading rule was fixed before any
run started. Read on its own, the transformer gains from 16 weeks of history over 8 on H\&M too: test log-space $R^2$
with a 16-week history minus with an 8-week one, the mean over three seeds of the paired differences, is $+0.129$ (95\%
$t$ interval over the three per-seed differences, $+0.125$ to $+0.133$, covering seed-to-seed noise only). The
three-seed mean test $R^2$ is $0.175$ at 8 weeks and $0.304$ at 16 weeks (descriptive). The ridge probe's H\&M results
above come from a different model, split and fit, and are not compared with these. The model is the paper's d128,
2-layer transformer and recipe at a learning rate of $5{\times}10^{-4}$, untuned for H\&M. It sees each line's price and
its place in the sequence, oldest first, with no date and no item or category: the file has no category, and the
paper's recipe reads no calendar or timing feature on the main data either. So the longer window adds older purchases,
in order, without their dates. The ridge probe's caveats apply: seven of the eight target years run into the COVID
period, a customer's first line in the file can fall inside the longer window, and how customers came to be in the
file is not documented. Earlier anchors' targets overlap later anchors' inputs in calendar time, so this is not
out-of-time validation. The data are used under the competition's non-commercial research terms, with no
redistribution, and only aggregate results are reported.

Re-scored on the five anchors whose whole 16-week history holds no November, December or January day (2019-06-03 to
2019-09-23; $202{,}457$ of the $317{,}178$ test rows), the same six runs gain $+0.129$ ($+0.124$ to $+0.134$;
log-space $R^2$ about those rows' own mean, three seeds), so the transformer's gain from 16 weeks over 8 shows on H\&M
without holiday weeks in the history. Its reading rule was fixed before any H\&M run was scored. These rows are $64\%$
of the pooled gain's, so the check is not independent of it, and their targets hold 82 to 194 days on or after
2020-03-11, against 0 to 194 across all eight anchors: the subset trades the holiday question for a heavier COVID
share, so it is not compared with the pooled gain.

\section{A simple model of the window gain}
\label{app:toy}

\S\ref{app:public-ladder} orders the public datasets with a toy model of purchase counts. Here we state it, with
what it implies and where it fails on our data. Its reading of our main dataset below came after we saw the results
there, while the predictions in \S\ref{app:public-ladder} used it before their runs. It is not a theorem about
the transformer, which reads much more than weekly counts (\S\ref{sec:miss}).

\emph{The model.} Customer $i$ buys at a steady Poisson rate $\lambda_i$ per week, drawn from a fixed population with
mean $\E[\lambda]$, variance $\mathrm{Var}(\lambda)$ and squared coefficient of variation
$\mathrm{CV}^2(\lambda) = \mathrm{Var}(\lambda)/\E[\lambda]^2$. This is the gamma-Poisson core of classic purchase-count
models~\citep{fader2005counting}, without the dropout process they add. Next year's activity $Y_i$ depends on the window
only through the rate, linearly: $\E[Y_i \mid \lambda_i, \text{window}] = a + b\lambda_i$. A $T$-week count $N_i$ is
then $\mathrm{Poisson}(\lambda_i T)$ given $\lambda_i$, with $\mathrm{Var}(N_i/T) = \mathrm{Var}(\lambda) + \E[\lambda]/T$,
so it estimates $\lambda_i$ with reliability (the squared correlation of $N_i/T$ with $\lambda_i$)
\[
  \frac{\mathrm{Var}(\lambda)}{\mathrm{Var}(\lambda) + \E[\lambda]/T} = \frac{x}{x+1},
  \qquad x = T\,\E[\lambda]\,\mathrm{CV}^2(\lambda).
\]
Because $Y_i - a - b\lambda_i$ is uncorrelated with $N_i$, the best linear predictor of $Y$ from $N$ reaches
$R^2 = \rho\,x/(x+1)$, where $\rho = b^2\,\mathrm{Var}(\lambda)/\mathrm{Var}(Y)$ is the share of target variance that a
perfectly known rate would explain. This holds for any rate population with finite variance. If the population is
gamma, the linear predictor is also the posterior mean of $\lambda_i$, and since the total count is sufficient for
$\lambda_i$ under Poisson weeks, no predictor of any form from the weekly counts does better. Four things follow.

\emph{(a) Diminishing returns.} $x/(x+1)$ rises with $T$ and is concave in it, so each added week helps less than the one
before and the curve flattens toward $\rho$, as the weekly probe's curve does (Figure~\ref{fig:prereg-arc}a).

\emph{(b) Where doubling helps most.} Doubling $T$ doubles $x$, so the gain from doubling the window is
\[
  \rho\,\frac{2x}{2x+1} - \rho\,\frac{x}{x+1} = \rho\,\frac{x}{(x+1)(2x+1)}.
\]
The derivative of $x/(2x^2+3x+1)$ has numerator $1-2x^2$, so at a fixed $\rho$ the gain is largest at $x = 1/\sqrt{2}$,
and of two populations past that point, the one with the larger $x$ (at a similar $\mathrm{CV}^2$, more frequent buying)
gains less. The three cross-dataset predictions of \S\ref{app:public-ladder} were written from this before their
runs: two came out as predicted and one is unresolved. The datasets differ in $\rho$ and in much besides $x$, so we read
those orderings as consistent with the model, not as tests of it.

\emph{(c) Which week activity fell in adds nothing beyond totals.} Given $\lambda_i$, the weeks are independent and
identically distributed, so they are exchangeable. Take any per-week features that are also independent and identically
distributed across weeks given the customer, as weekly counts are here (and weekly log activity is too, if each
customer's spend per purchase is also steady). The normal equations of the best linear predictor of $Y$ from every
week's features do not change when the weeks are permuted, so their solution, when unique, puts the same weight on every
week, feature by feature. Separate weights per week then add nothing over the per-feature totals. \S\ref{app:calendar}
measures that addition on the test cohort: the per-week weights carry $-0.04\%~[-0.12\%, 0.04\%]$ of the
8-to-16-week gain, as the model predicts (on holiday-free anchors, $0.08\%~[0.02\%, 0.15\%]$: above zero, but small).

\emph{(d) Where the model fails.} With one such feature, the best linear predictor's $R^2$ at $T$ weeks has the form
$r\,sT/(1+sT)$ for some $r, s > 0$ ($r = \rho$ and $s = \E[\lambda]\,\mathrm{CV}^2(\lambda)$ for counts); with several,
it is a sum of such terms, each with its own $r$ and $s$. For each term, the per-week gain over weeks 22 to 24, divided by the per-week gain over weeks 8
to 16, is
\[
  \frac{(1+8s)(1+16s)}{(1+22s)(1+24s)} > \frac{8\cdot 16}{22\cdot 24},
\]
since $(1+8s)/(1+22s)$ and $(1+16s)/(1+24s)$ each fall with $s$ toward $8/22$ and $16/24$, and a sum of terms that
each clear this bound clears it too. So under any steady-rate population, whatever $r$ and $s$ are, weeks 22 to 24 add
at least $(8\cdot 16)/(22\cdot 24)$ as much per week as weeks 8 to 16. On holiday-free anchors the probe's ratio
is $0.180~[0.160, 0.201]$, an interval over scoring noise only (\S\ref{app:calendar}), below that floor. Older
weeks are worth less here than steady buying implies, as they would be if rates or spend per purchase drift, buying
follows the season (here weeks 22 to 24 back fall between January and April), or customers lapse; we have not tested
which.

\emph{What the model leaves out.} Our target is log activity over a year, and the probe reads log activity as well as
counts, so the formulas give direction and shape, not levels, and $\rho$ is unknown for each dataset. Our rows are also
selected in ways the model's population is not: each extract's eligibility widens with its window
(\S\ref{sec:matched-method}), and every row of the 1M matched cohorts has an event in the 8 weeks before its target week
(\S\ref{app:cohort-keeps}). The model has no calendar, so the sharp bend on the test calendar, and any seasonal
pattern on holiday-free anchors, are outside it. And real events carry per-event values and timing as well as a
count, which the transformer reads and this model does not.

\section{Who the matched cohort keeps}
\label{app:cohort-keeps}

Every row of the 1M matched cohorts of \S\ref{sec:matched-method} (the three-window cohort, $n{=}407{,}338$, and the
four-window variant, $n{=}372{,}778$) has at least one event in the 8 weeks before its target week. We checked this
directly: no row of either cohort lacks such an event. The matched
cohorts therefore hold only recently active customers, and a matched gain is a gain for customers active in the 8
weeks before the target week. Of the $73{,}335$ M16 test rows that the three-window cohort drops, $66.3\%$ have no
event in those 8 weeks. Of the $107{,}874$ M24 test rows that the four-window variant drops, $71.0\%$ have none and
$36.4\%$ have a zero target (vs.\ $4.8\%$ of the rows it keeps). These are counts over the test split's raw events,
with no model involved. For the whole test split, the relevant estimates are the full-split deltas.

\section{Instrument specifications}
\label{app:instruments}

This appendix gives
the settings behind the trained models and probes this paper draws a
number from. \S\ref{app:notfixed} records where the input pipeline departs
from these settings.

\subsection{The objective actually minimized}
\label{app:objective}

Every transformer run minimizes a two-term loss. Write $L_{\mathrm{agg}}$ for the
loss on the aggregate prediction (the quantity every reported $R^2$ scores)
and $L_{\mathrm{evt}}$ for an auxiliary per-event loss:
\begin{equation}
L \;=\; \gamma\, s_{\mathrm{evt}}\, L_{\mathrm{evt}}
\;+\; (1-\gamma)\, s_{\mathrm{agg}}\, L_{\mathrm{agg}},
\qquad
\gamma = 0.3,\quad s_{\mathrm{evt}} = 1.0,\quad s_{\mathrm{agg}} = 1.5,
\label{eq:objective}
\end{equation}
so the realized weights are $0.30$ on the per-event term and $1.05$ on the
aggregate term. They are constant for the whole run.

The swept ``objective'' factor (\S\ref{sec:results}) sets only $L_{\mathrm{agg}}$,
choosing among Huber with $\delta = 1$, squared error, and pinball at $\tau = 0.5$, all in the
log target space of \S\ref{sec:eval}. Under the primary recipe
(Table~\ref{tab:recipe}), $L_{\mathrm{evt}}$ is pinball at $\tau = 0.5$ in every
run, the Huber and MSE runs included, so the objective sweep changes one of the two
terms and holds the other fixed.
The per-event term uses a causal shift, scoring the prediction at position $i$
against event $i{+}1$'s activity value, and applies only at positions whose target is
strictly positive. Its target is per-event activity in the same log space as the
aggregate target.

Two classification heads exist but carry no loss weight: the per-event
purchase classifier enters $L_{\mathrm{evt}}$ at weight $0$, and the aggregate
classifier enters $L_{\mathrm{agg}}$ at weight $0$. The aggregate classifier is
still on the prediction path. Its logit forms a gate
$g = \sigma(\ell / T)$ with $T = 0.5$, hard-thresholded at $0.5$ with a
straight-through estimator. The gate multiplies both the aggregate head's output and the
linear anchor below, so the aggregate classifier is trained through $L_{\mathrm{agg}}$ rather than
through a classification term of its own.

The aggregate prediction is a residual on a linear anchor:
$\hat{y} = g\,(a\,x + b) + g\,f(z)$, where $x$ is the input-window activity total in
the same log space, $a$ and $b$ are two scalars initialized at $1$ and $0$, and
$f(z)$ is the direct aggregate head (\S\ref{app:arch}) applied to $z$, the pooled representation with $x$ appended. Those two scalars train in their own optimizer group
at $10\times$ the base learning rate with no weight decay, and the rest of the model
is frozen for the first $200$ steps so the anchor calibrates first.

The recurrent and continuous-time baselines minimize a single term on the
aggregate prediction, with no auxiliary per-event term and no gate.
That term is Huber with $\delta = 1$ by default and squared error in the
objective-equalized runs of \S\ref{sec:lossnull}.
The baselines therefore match transformer runs with the same objective on $L_{\mathrm{agg}}$ but
lack $L_{\mathrm{evt}}$. The objective-matched comparison rule of
\S\ref{sec:eval} matches the aggregate loss family, the term that the reported
metric scores.

The recipe-transfer runs of \S\ref{sec:external} differ from this configuration in more than the metric.
Both of their loss terms are squared error, their target is raw and winsorized at the
percentiles \S\ref{sec:external} states, and the linear anchor is off. The
input-window total neither anchors the prediction nor enters the aggregate heads. The
gate and the weights of \eqref{eq:objective} are unchanged.

\subsection{Transformer architecture and parameter accounting}
\label{app:arch}

Table~\ref{tab:recipe} gives the shape ($d_{\mathrm{model}} = 128$, 2 layers, 4
heads, dropout $0.1$, attention pooling). The rest of the architecture, in order:

\begin{itemize}[leftmargin=1.4em,itemsep=1pt,topsep=2pt]
\item Six categorical event fields embed at widths $4$, $8$, $16$, $4$, $16$, $6$
  (index $0$ is reserved for padding and maps to the zero vector).
\item The two numeric event fields enter unstandardized and project $2 \to 64$
  through one linear map.
\item The time-feature block contributes a constant. Its two channels are defined as the
  event's day of week divided by $6$ and week index within the input window divided by
  $M-1$, projected $2 \to 16$. Neither field exists in the export's event records, and
  nothing in the data pipeline computes either, so both channels are identically zero on
  every event (\S\ref{app:baselines}). The
  base transformer's own event-time information therefore comes entirely from rotary positional
  encoding over event order.
\item Those three blocks concatenate to $134$ channels and project to
  $d_{\mathrm{model}}$ through a linear map, layer norm, GELU and dropout.
\item Two pre-norm encoder layers, GELU, feed-forward width
  $4\,d_{\mathrm{model}} = 512$, causal mask, final layer norm. Rotary encoding is
  applied inside attention (\S\ref{app:pe}); nothing positional is added to the
  embeddings.
\item Attention pooling scores position $i$ as $\ell_i = w^{\top} h_i + w_0$, a single linear map,
  masks padding to $-\infty$, applies a softmax over positions and
  returns $\sum_i \mathrm{softmax}(\ell)_i\, h_i$. All-padding rows pool to zero.
\item The two per-event heads are two-layer MLPs ($d = d_{\mathrm{model}}$), $d \to d/2 \to 1$ with GELU and
  dropout. The three aggregate heads all take one extra input, the input-window total $x$ of
  \S\ref{app:objective}: the classifier is $d{+}1 \to d/2 \to 1$, and the conditional
  regressor (for the aggregate target given that it is positive) and the direct head are both $d{+}1 \to d \to d/2 \to 1$.
\item Sequences are the most recent $750$ events, right-padded.
\end{itemize}

Two heads hold parameters that no loss term reaches: the per-event purchase
classifier (weight $0$) and the conditional regressor, which the forward pass
computes but nothing scores once the direct aggregate head is enabled. Both are
counted in Table~\ref{tab:params}, because that count is a sum over state-dict
entries.

That accounting reconciles with the printed total of $491{,}248$. Every component above is fixed by the
recipe, and their state-dict entries sum to $489{,}848$. That sum counts the $64$-entry rotary frequency table of \S\ref{app:pe} once per prefix, $3 \times 64 = 192$ entries in all, and leaves $1{,}400$ for the six
embedding tables. On the primary dataset every run sizes those tables from the schema's default encoder, whose six vocabulary sizes $V_k$
are $6/13/43/5/33/6$. Against the widths $d_k$ above, $\sum_k V_k d_k = 1{,}400$, exactly the residual. Everything else in Table~\ref{tab:params} for this family
is reconstructable from the list above.

\subsection{Time representation: the controls and MS-RoPE}
\label{app:pe}

All three encodings compared in \S\ref{sec:time} act on queries and keys inside attention and add
nothing to the embeddings. The integer-position control (RoPE) uses the
standard geometric frequencies
$\omega_j = b^{-j/(d/2)}$, $j = 0,\dots,d/2-1$, with $b = 10^{4}$.

MS-RoPE replaces integer position with physical elapsed time and superposes $K$
learnable time scales on that one set of frequencies. Let $\Delta t_i$ be the elapsed seconds
between event $i$ and the first retained event in the sequence. With per-band scales
$\tau_k$ and mixing weights $\alpha = \softmax(u)$,
\begin{align}
\theta_{k,j}(\Delta t) &= (\Delta t / \tau_k)\,\omega_j \bmod 2\pi,
\label{eq:msrope-angle}\\
c_j(\Delta t) &= \sum_{k=1}^{K} \alpha_k \cos\theta_{k,j}(\Delta t),
\qquad
s_j(\Delta t) = \sum_{k=1}^{K} \alpha_k \sin\theta_{k,j}(\Delta t),
\label{eq:msrope-mix}
\end{align}
and the map applied to each query and key vector $v$ is
$v \mapsto v \odot c + \mathrm{rotate\_half}(v) \odot s$. As implemented, the map is not in general a rotation in any of the three encodings.
At $d = d_{\mathrm{model}} = 128$ and $n_{\mathrm{heads}} = 4$, each head applies the map with $c_0, \dots, c_{31}$ and
$s_0, \dots, s_{31}$. Within a head, $\mathrm{rotate\_half}$ pairs coordinate $m$ with $m + 16$, but those two coordinates
carry different frequency indices, $j = m$ and $j = m + 16$. Three properties of
\eqref{eq:msrope-mix} separate this encoding from
the two controls of \S\ref{sec:time}.
First, every frequency pair $j$ carries all $K$
bands, so the bands do not partition the frequencies. Second, the mixing acts on the unit vectors, not on the angles.
A convex combination of $K$ unit vectors has modulus $\le 1$, so the mixed pair $(c_j, s_j)$ can leave the
unit circle that both controls' pairs sit on, and destructive interference
between bands acts as a scale-aware attenuation. Third, $\tau$ and $u$ are learned:
$\log \tau$ is a free parameter clamped to $[-5, 25]$, with $\tau$ initialized at
$\{1\,\mathrm{h}, 1\,\mathrm{d}, 7\,\mathrm{d}, 365.25\,\mathrm{d}\}$ in seconds for
$K = 4$, and $u$ is initialized at zero, so mixing starts uniform at $1/K$.
The unconstrained control of \S\ref{sec:time} is the same physical-time map
with one free multiplier per frequency pair instead of $K$ shared bands.
The angle of pair $j$ is $\lambda_j\,(\Delta t/\tau_j)\,\omega_j$, where $\lambda_j$ is its multiplier and $\tau_j$ is log-spaced from $1\,\mathrm{min}$ to $365.25\,\mathrm{d}$ across the $d/2$ pairs.
The multipliers are initialized at $1$, so the control starts on that fixed log-uniform $\tau$ spectrum.
Each head applies only the pairs $j < 32$, so the spectrum actually used spans $1\,\mathrm{min}$ to about $10.9\,\mathrm{h}$.
Free
per-frequency parameters of this kind are the parameterization of a learnable Fourier
feature map. The non-parametric variant of \citet{xu2019functional} uses them over a scalar time span,
optimizing its transformed frequency samples directly as free parameters.
\citet{li2021fourier} use them over multi-dimensional spatial positions, with an MLP modulation
this control does not use. Both add the features
to the representation or concatenate them with it, and neither acts on queries and keys,
so the control is not a special case of either. What it borrows is the free
per-frequency parameterization, which \citet{karypis2026lerope} apply to the rotary frequencies of token position and the control applies to one physical-time axis.
MS-RoPE \eqref{eq:msrope-mix} is not a restriction of that parameterization. It ties
the multipliers to $K$ shared band scales and then mixes $K$ rotations on the unit
circle, which by the modulus argument above changes the functional form.

One module instance is shared by every layer and every head: each layer holds a
reference to it and uses the leading $d/n_{\mathrm{heads}}$ entries
of the mixed vector, the same convention as both controls. This sharing also
explains how the $+36$ parameters of Table~\ref{tab:params} decompose.
Per state-dict prefix MS-RoPE adds $12$ entries over RoPE: $K$ band scales,
$K$ mixing logits, and a $K$-entry snapshot of the initial scales that is written to
the checkpoint for post-hoc drift diagnostics. A two-layer model registers the
shared module under three prefixes (the model's own and one per layer), giving $3 \times 12 = 36$. Only $2K = 8$ parameters behind
those entries are trainable, so a count of distinct trainable parameters gives
$+8$. The difference is bookkeeping and adds no capacity.

\subsection{The weekly-aggregate ridge probe}
\label{app:probe}

This is the instrument behind the saturation law (\S\ref{sec:saturation}), the
allocation result (\S\ref{sec:allocation}) and two of the four internally registered commitments, (ii) and (iii) of Table~\ref{tab:prereg}.
Each week $w$ contributes two features: the log of one plus the clipped-at-zero sum of that
week's activity, and the count of that week's events. Weeks are indexed backwards
from the anchor (the week a forecast is made from) by
$w = \lfloor (\text{target week date} - \text{event date in days} - 1)/7 \rfloor$, so
$w = 0$ is the week nearest the anchor. A week with no events contributes zeros. The
$k$-week fit uses exactly the weeks $w < k$, so it has $2k$ features. A $k$-week
feature set therefore contains every set with fewer weeks, which is what
\S\ref{sec:saturation} means by \emph{nested}. Each $k$ is refit from scratch. Features are
standardized, and the ridge estimator's built-in leave-one-out cross-validation picks the
penalty from $25$ log-spaced values in $[10^{-3}, 10^{3}]$. Ridge is then
fit in closed form on the extract's full train split, so no random seed is involved. The score is
$1 - \mathrm{SSE}/\mathrm{SST}$ on the matched-cohort rows of the test split, with
$\mathrm{SST}$ taken about that cohort's own mean, in the log target space of
\S\ref{sec:eval}. Per-decile curves refit the same instrument within each decile.
\S\ref{app:calendar} reruns it on training-period anchors. There, the sharp 22--24-week bend persists on test-calendar
anchors, scored on all their held-out customers, and is not shown to persist on holiday-free ones, while returns diminish in both groups.
\S\ref{app:public-ladder} runs it on four public datasets that each span more than a year.

\subsection{Recurrent and continuous-time baselines}
\label{app:baselines}

All four baselines use the same extracts as the transformer and apply the train-split scaler of \S\ref{sec:eval}.
In the main runs, their only inputs are the per-event numeric channels, four per event: the two numeric event fields of \S\ref{app:arch} and the
two normalized time features defined there. The
two time channels are identically zero on this export, so only two of the four carry signal. None of
the baselines sees the categorical embeddings.
Hidden width is $128$ in all four, set by the same config field as the transformer's
$d_{\mathrm{model}}$. GRU and LSTM are 2-layer with dropout $0.1$ between layers.
Every baseline ends in the same $128 \to 64 \to 1$ GELU head plus the two-scalar linear
anchor of \S\ref{app:objective}. In the baselines the anchor is ungated and trains in each model's single AdamW parameter group,
with no anchor-specific learning rate, no weight-decay exemption, and nothing frozen at the start. GRU and LSTM pool by taking the last layer's final
hidden state over the packed sequence; ContiFormer and Neural Hawkes mean-pool over
valid positions.

The two continuous-time baselines are simplified versions of their published
designs.
Our ContiFormer is a two-layer causal attention stack with an additive per-head
time bias $\exp(-|a_h|\,|\Delta t_{ij}|/3600) + b_h$ on the attention logits. It has
no ODE solver and so no solver tolerances. Our Neural Hawkes is a single LSTM
cell whose previous hidden state is decayed by
$\exp(-\sigma(W[x_t; h_{t-1}])\,\Delta t)$, with $\Delta t$ the inter-event interval in hours (in days in the paired working-clock runs below). It has no intensity function, no likelihood term and no
Monte-Carlo integration, so it is a decay-gated recurrent regressor rather than a
point process. Both are trained on the same aggregate regression loss as the other
baselines. The architecture band of \S\ref{sec:universal} therefore
compares this transformer against these implementations, not against the published
models.

In the main runs, both models also ran with $\Delta t \equiv 0$, a stronger simplification than either of those. In those runs, the only time input
either one reads is a two-column per-event tensor built from a day-of-week field and a
week-index field. Neither field exists in this export, so both columns are zero on every event.
We checked $15{,}041$ events across $200$ rows
on the 16-week 1M extract and found both columns identically zero. The export does carry a per-event timestamp, but
in the main runs neither model consults it.
ContiFormer's bias becomes $\exp(0) + b_h = 1 + b_h$, a per-head constant added to
every logit in an attention row, which cancels in the softmax; $a_h$ and $b_h$ receive
identically zero gradient, and what remains is a plain causal attention stack.
Neural Hawkes' decay factor becomes $\exp(0) = 1$, so the hidden state is not decayed
between events and the decay gate's parameters likewise receive no gradient, leaving a
standard LSTM cell. As a result, the architecture band does not test continuous-time
mechanisms on this data.
Instead, the band shows that two architectures with inert
time handling still fall within the ${\leq}0.0052$ seed-42 spread of
\S\ref{sec:universal} on the full test split.
We tested this for ContiFormer and Neural Hawkes in a separate set of runs that give each the export's own per-event timestamp, with its clock path as built: each sequence is sorted by time, ContiFormer's time bias above becomes $\exp(-|a_h|\,|\Delta t_{ij}|) + b_h$ with $|\Delta t_{ij}|$ in days, and Neural Hawkes' hidden state is decayed between events by $\exp(-\sigma(W[x_t; h_{t-1}])\,\Delta t)$ with $\Delta t$ in days, where the decay gate's bias starts at $-4$. The day-of-week and week-index columns stay zero. The runs use M16 at the 1M tier, 8 ranks and three seeds, and each seed is run with the clock on and off over the same rows. Each epoch trains on a capped subset of each rank's rows that differs from the main runs' subset, so no score from these runs is set beside the main runs' scores. The reading rule was fixed before any of these runs started. For ContiFormer the paired change in test $R^2$ is $+0.0012$ (95\% $t$ interval over the three paired seed differences, $-0.0009$ to $+0.0033$, covering seed-to-seed noise only), so no gain from a working clock is detected. For Neural Hawkes it is $+0.0116$ (95\% $t$ interval over the three paired seed differences, $+0.0102$ to $+0.0130$, again covering seed-to-seed noise only), so a working clock helps Neural Hawkes. That gain is about twice the ${\leq}0.0052$ spread above. The spread comes from the main runs and the gain from these paired 8-rank runs, and neither is bridged into the other. The spread describes the five families with their clocks as built, and where Neural Hawkes would land among the five with a working clock is not measured.
At hidden width $128$, all four baselines reproduce the counts in Table~\ref{tab:params} exactly.
Their own budget is batch $128$, $10$ epochs, single process, and the
$384{,}000$-row cached draw (\S\ref{sec:repro}).

\subsection{Checkpoint selection and retention}
\label{app:ckpt}

The transformer ladder runs select their epoch on one scale and are scored on another.
This metric-space mismatch is the source of the re-selection offsets quantified in
\S\ref{sec:limitations}. Every reported $R^2$ is computed in log space, except in the
winsorized raw-target runs of \S\ref{sec:external}. For the ladder runs, each reported number
comes from the epoch with the lowest validation mean squared error, computed on the original
scale after the inverse log transform. Not every
transformer run selects on that metric. The later corrected-schedule capacity and scale retrains, the twelve
retraining runs and the cross-evaluation select on the recorded log-space validation series, and
\S\ref{sec:limitations} assesses their offsets under that criterion. The
recurrent and continuous-time baselines select no epoch and keep their final one, and
the tabular baselines have no epoch to select, so \S\ref{sec:limitations} interprets the
cross-family band accordingly.
On the ladder runs, early stopping monitors the same quantity, with a patience of
$8$ epochs and a minimum improvement of $10^{-4}$. The
twelve retraining runs and the cross-evaluation ran with it effectively off (patience $999$). The corrected-schedule capacity and scale retrains kept patience $8$ on the log-space series.
Both capacity configurations had to be retrained (\S\ref{sec:limitations}) because a later build prunes
checkpoints. The build behind the two 1M ladder runs we audited
prunes nothing, so both kept all twenty epochs. The later build behind the widest
capacity configuration and a 10M run keeps only the selected checkpoint and the most recent epoch's,
so the widest configuration's best epoch on the log-space validation series could not be re-scored on test.

\subsection{Determinism configuration}
\label{app:determinism}

On the anchored builds, no run is cited until its build reproduces a fixed reference run
bit-for-bit (\S\ref{sec:repro}); the settings below make that possible.
Transformer runs on those builds execute in strict mode: cuDNN deterministic kernels, cuDNN autotuning
off, PyTorch's deterministic-algorithm enforcement on rather than warn-only, a fixed
cuBLAS workspace configuration set before the first device operation, and NCCL
pinned to the ring algorithm and simple protocol so the all-reduce order is fixed.
The recurrent and continuous-time runners set none of these: they seed Python, NumPy
and PyTorch directly and, at their own budget, run single-process. Only their multi-rank runs carry any of these: the 32-rank transformer-budget recurrent retrains of \S\ref{sec:lossnull} and the paired 8-rank working-clock runs of \S\ref{app:baselines} inherit the NCCL algorithm and protocol settings.
Mixed precision is off. Each dataloader carries its own seeded generator, and every
worker reseeds from a mix of the base seed, its epoch and its rank, so a respawned
streaming worker is reproducible rather than merely distinct.

\subsection{Taobao input mapping}
\label{app:taobao-inputs}

The model recipe is unchanged on Taobao (\S\ref{sec:external}), but the inputs
must change. The public corpus carries user, item, category, behavior type and
timestamp, and none of the numeric fields the primary schema expects.
Behavior type fills the event-type code as
$\{\text{view}, \text{purchase}, \text{cart}, \text{favorite}\} \to \{0,1,2,3\}$.
The default encoder of \S\ref{app:arch}, however, keys its event-type vocabulary on the
primary schema's labels, so all four codes map to the same reserved code and the
event-type input is constant on Taobao.
Item and category identifiers fill the primary schema's item and category slots, hashed modulo
$50{,}000$ and $5{,}000$. The remaining categorical slots are unused.
The first numeric event field is $1$ on a purchase event and $0$ otherwise, and the
second is $0$ on every event, since the public corpus carries nothing to fill it.
Per-event activity, the input-window total and the
target are all purchase counts.
The architecture, the optimizer and the two-term
composition of \eqref{eq:objective} are as in Table~\ref{tab:recipe}, and the
aggregate loss is the one \S\ref{sec:external} names for each configuration. The seeds are those of
\S\ref{sec:repro}, with the replacement
seeds that \S\ref{sec:external} reports.

\subsection{Where the input pipeline departs from the specification}
\label{app:notfixed}

On the primary dataset the transformer's six categorical inputs are constant across events.
  Its input pipeline encodes those fields with empty vocabularies rather than the default
  encoder of \S\ref{app:arch}, so every value maps to one reserved code.

\end{document}